\documentclass[review]{elsarticle}

\usepackage{lineno,hyperref}
\modulolinenumbers[5]
\usepackage{graphicx}
\graphicspath{{./figures/}}
\usepackage{subcaption}
\newcommand{\mbold}[1]{\mbox{\boldmath ${#1}$}}
\usepackage{xr}
\usepackage{changepage}
\begin{document}

\begin{frontmatter}

\title{A Framework for an Assessment of the Kernel-target Alignment in Tree Ensemble Kernel Learning}

\author[mymainaddress]{Dai Feng}
%\address{}
\address[mymainaddress]{Data and Statistical Sciences, AbbVie Inc., North Chicago, IL, United States of America}
\cortext[mycorrespondingauthor]{Corresponding author}
\ead{dai.feng@abbvie.com}

\author[mysecondaryaddress]{Richard Baumgartner\corref{mycorrespondingauthor}}
%\address{\orgdiv{Merck \& Co., Inc.}, \orgname{Kenilworth, NJ}, \country{United States of America}}
\address[mysecondaryaddress]{Biometrics Research, Merck \& Co., Inc., Kenilworth, NJ, United States of America}
\ead{richard_baumgartner@merck.com}
%\fntext[myfootnote]{Corresponding author}
%\ead{dai.feng@abbvie.com}


%\author[mysecondaryaddress]{}

\begin{abstract}
Kernels ensuing from tree ensembles such as random forest (RF) or gradient boosted trees (GBT),
when used for kernel learning, have been shown to be competitive to their respective tree ensembles
(particularly in higher dimensional scenarios). On the other hand, it has been also shown that
performance of the kernel algorithms depends on the degree of the kernel-target alignment. However,
the kernel-target alignment for kernel learning based on the tree ensembles has not been investigated
and filling this gap is the main goal of our work.

Using the eigenanalysis of the kernel matrix, we demonstrate that for continuous targets good
performance of the tree-based kernel learning is associated with strong kernel-target alignment. Moreover,
we show that well performing tree ensemble based kernels are characterized by strong target aligned components
that are expressed through scalar products between the eigenvectors of the kernel matrix and the
target. This suggests that when tree ensemble based kernel learning is successful, relevant information for the
supervised problem is concentrated near lower dimensional manifold spanned by the target aligned
components. Persistence of the strong target aligned components in tree ensemble based kernels is
further supported by sensitivity analysis via landmark learning. In addition to a comprehensive
simulation study, we also provide experimental results from several real life data sets that are in line with the
simulations.

\end{abstract}

\begin{keyword}
\texttt{tree-based ensembles \sep kernel learning \sep kernel-target alignment \sep eigenanalysis \sep relevant dimensionality}
\end{keyword}

\end{frontmatter}

%\linenumbers

\section{Introduction}
\label{sec:intro}
Tree-based ensembles such as random forest (RF) and gradient boosted trees (GBT) are a mainstay in machine learning and in particular they are considered as dominant algorithms for tabular data \cite{feng2018}. Tree-based ensembles can be also interpreted as kernel generators and this interpretation has been expounded theoretically to investigate their asymptotic properties (\cite{scornet2016} and \cite{Chen2018}). On the other hand, the RF and GBT kernels plugged into kernel learning have been shown to perform well, even outperforming their respective ensembles in comprehensive simulations and real life data sets \cite{davies2014},\cite{feng2021}.

The kernel-target alignment has been first proposed in \cite{cristianini2001},\cite{cristianini2006} for classification and later extended in \cite{braun2009} for regression as a means of characterization of the relevant information in supervised learning. It has been shown that the kernel-target alignment can be quantified using eigenvectors of the kernel matrix and the target variable. The kernel-target alignment enables assessment of a match of a given kernel to the learning problem represented by a particular data set. In Refs. \cite{cristianini2001},\cite{cristianini2006}, the analysis of kernel-target alignment was applied to classification, in \cite{braun2009} it was used for regression and in \cite{montavon2011} it was applied in the analysis of hidden layers of deep neural networks.

Landmark (prototype) or (dis)similarity based learning \cite{pekalska2001}, \cite{bien2011}, can be used as an alternative way to develop prediction models in nonlinear feature spaces via the kernel matrix \cite{kar2011},\cite{balcan2008}. In similarity/dissimilarity learning the kernel entries are explicitly interpreted as pairwise similarities/dissimilarities between the points (samples).  Recasting the problem this way, prediction models can be built accordingly. Tree ensemble generated kernels can therefore be readily used within this paradigm.

Kernel-target alignment has never been evaluated for the tree-based ensembles, however eigenanalysis of the RF kernel matrix was suggested as potentially useful in \cite{breiman2000}. Focus of our investigation is the kernel-target alignment of the tree ensemble based kernels and its relationship with the performance of related kernel learning. Furthermore, we propose a sensitivity analysis through landmark learning. The remainder of the manuscript is organized as follows: Section 2 introduces the theoretical framework of the tree ensemble learning, formalizes the notion of the kernel-target alignment and landmark learning, Section 3 details a simulation study that systematically evaluates kernel-target alignment and performance of the ensemble based kernels, Section 4 summarizes the experiments on real life data sets and Section 5 provides discussion, conclusions and future research directions.

\section{Methods}
\label{sec:method}

\subsection{Terminology}
\label{subsection1:method}
Following Breiman \cite{breiman2000} and Refs. \cite{ishwaran2019}, \cite{Chen2018} and \cite{scornet2016}, we consider a supervised learning problem, where training set of $n$ pairs: $D_n=\{(\mbold{X_1},Y_1), (\mbold{X_2},Y_2),\ldots,({\mbold{X_1},Y_n})\}$ is provided. $\mbold{X_i} \in R^p$. In our case $Y_i$ is a continuous target, for which the $Y_i \in R$. Let the target vector be then $\mbold{Y}=[Y_i,Y_2,\ldots,Y_n]^T$.

\subsection{Tree Ensemble Based Kernel Learning}
\label{subsection2:method}
Kernel methods in the machine learning literature are a class of methods that are formulated in terms of a similarity (Gram) matrix $\mbold{K}$. The similarity matrix $K_{i,j}=k(\mbold{X_i},\mbold{X_j})$ represents the similarity between two points $\mbold{X_i}$ and $\mbold{X_j}$.
Kernel methods have been well developed and there is a large body of references covering their different aspects \cite{herbich2001},\cite{schoelkopf2001},\cite{friedmanHastieTibshirani2009}. 
In our work we used a common kernel algorithm, namely kernel ridge regression (KRR). KRR is a kernelized version of the traditional linear ridge regression with the L2-norm penalty. Given the kernel matrix $\mbold{K}$ estimated from the training set, first the coefficients $\mbold{\alpha}$ of the (linear) KRR predictor in the non-linear feature space induced by the kernel $k(.,.)$ are 
obtained:
\begin{eqnarray}
\mbold{\alpha}&=&(\mbold{K}+\lambda \mbold{I_n})^{-1}\mbold{Y}
\end{eqnarray}
where $\lambda$ is the regularization parameter.

The KRR predictor $h_{\textrm{KRR}}(\mbold{X})$ is given as:  

\begin{eqnarray}
h_{KRR}(\mbold{X})&=&\sum_{i=1}^{n} \alpha_i
\mbold{k(X_i,X)}=\mbold{Y^T}(\mbold{K}+\lambda\mbold{I_n})^{-1}\mbold{K_i}
\label{Eq:KRRPredictor}
\end{eqnarray}
where $\mbold{K_i}=(k(X_1,X), \ldots, k(X_n,X))$.

Tree-based ensembles are aggregated from a set of regression trees $h_m(\mbold{X},D_n), m=1,2,\ldots,M\}$, with $h_m(\mbold{X},D_n)$ representing a single tree. Each single tree partitions the feature space into disjoint regions.
Moreover, for a single tree, each region in the feature partition is given by a unique decision path from the tree root to its terminal node.  As a byproduct a tree-based kernel is naturally generated via the regression trees and their respective feature partitions \cite{breiman2000},\cite{Chen2018}. Thus, a kernel that corresponds to a tree-based ensemble is obtained as a probability that  $\mbold{X_i}$ and $\mbold{X_j}$ are in the same terminal node $R_k$($h_m$) \cite{Chen2018}

\begin{eqnarray}
k_{ensemble}(\mbold{X_i},\mbold{X_j})=\frac{1}{M}
\sum_{m=1}^M \sum_{k=1}^T I(\mbold{X_i},\mbold{X_j} \in R_k(h_m)).
\label{EqKernelxgb}
\end{eqnarray}

The ensemble based kernels can be obtained by various feature space partitioning mechanisms \cite{fan2020}. We used the RF and extreme gradient boosting (XGB) as an implementation of the GBT in our work and their algorithmic details are provided for completness in the Appendix. We will use GBT and XGB in the remainder of the manuscript, interchangeably.

\subsection{Kernel-target Alignment}
\label{subsection3:method}
In the Ref. \cite{cristianini2001}, the sample ordering according to the leading eigenvector of the kernel matrix was shown to correspond to the class delineation and was considered as a proxy of the kernel target alignment. This idea was further developed in \cite{braun2009} via eigenanalysis of the kernel matrix.

We define the spectral (here it is also a singular value) decomposition of the kernel matrix $\mbold{K}$ as: 
\begin{eqnarray}
\mbold{K}&=&U\Lambda U^T
\label{Eq:Kalign}
\end{eqnarray}
where $U$ is an orthonormal matrix with the columns $u_i$ (eigenvectors of $\mbold{K}$) and $\Lambda$ is an $n$ by $n$ diagonal matrix with eigenvalues $\lambda_i$-s on the diagonal. We assume that $\lambda_i$-s are ordered according to their magnitude.

The kernel-target alignment components can be obtained as an absolute value of the scalar product $|u_i^T Y|$.
The fundamental result of \cite{braun2009} is concerned with the rate of the decay of kernel-target alignment components. In particular, it was proved that under mild conditions, the kernel-target aligned components decay at the same rate as the eigenvalues of the kernel matrix. As a consequence of this decay, the relevant information pertaining to the supervised problem is concentrated in the leading eigenvectors of the kernel matrix that are strongly aligned with the target \cite{braun2009},\cite{montavon2011}, if such strongly aligned components exist. 

In our investigation, we used normalized kernel-target alignment components given by the absolute value of the Pearson correlation coefficients between $u_i$-s and $\mbold{Y}$, with $\mbold{K}$ being the kernel matrix obtained from the tree-based ensembles.

\subsection{Landmark Learning in Nonlinear Kernel Feature Spaces}
\label{subsection4:method}
In the landmark learning, the empirical similarity map (data driven embedding)  \cite{balcan2008},\cite{kar2011} is  generated by selecting a subset of $n_L$ data points (landmarks) also referred to as a reference set \cite{pekalska2001}. A point in the feature space is represented by the similarities to the landmarks. This approach is akin to a dimensionality reduction of the original kernel problem to a lower $n_L$ dimensional problem \cite{balcan2008}.  A linear model is subsequently developed on the landmark features and a landmark predictor is obtained as:
\begin{eqnarray}
h_{Landmarks}=(\mbold{L}^{T}\mbold{L})^{-1}\mbold{L}^T Y
\label{Eq:landmarks}
\end{eqnarray}
where $L$ is an $n \times n_L$ matrix and $n_L$ is number of landmarks. $L_{i,j}=k(i,j)$ is a similarity of the point $i$ to the landmark $j$.
Consider now the singular value decomposition (SVD) of $L$:

\begin{eqnarray}
\mbold{L}&=&U_L\Sigma_L V_L^T
\label{Eq:Lalign}
\end{eqnarray}
%\hl{Do we need to use another symbol instead of "U", which was used in equation (4)?}

where in this case the matrix $U_L$ is an orthonormal matrix with the columns $u_{L_i}$ (left singular vectors of $\mbold{L}$) and $\Sigma_L$ is an $n_L$ by $n_L$ diagonal matrix with singular values $\sigma_{L_i}$ on the diagonal. We assume that $\sigma_{L_i}$-s are ordered according to their magnitude. Accordingly, the $\lambda_{L_i}=\sigma_{L_i}^2$-s % 
and $u_{L_i}$-s are the eigenvalues and the eigenvectors of the landmark kernel matrix $\mbold{K_L}=L L^T$, respectively.
Following the previous considerations for $\mbold{K}$ in the section \ref{subsection3:method}, the kernel-target alignment components in landmark learning can be obtained as an absolute value of the scalar product $|u_{L_i}^T Y|$. Again, we use here the absolute value of the Pearson correlation coefficient between $u_{L_i}$-s and $Y$, where $u_{L_i}$-s are obtained from the Equation \ref{Eq:Lalign}.

We used landmark learning in a sensitivity analysis, where increasing number of landmarks $n_L$ was randomly selected from the training set.

The code for the simulation and real life data analysis was developed in the R programming language \cite{Rcran}. The ranger \cite{ranger} implementation of RF and xgboost \cite{xgboost} of XGB was used, respectively.
All algorithms were applied using their default parameters.
The regularization parameter $\lambda$ for kernel ridge regression was chosen at minimum value, such that the matrix, $\mbold{K}+\lambda \mbold{I_n}$ was invertible. 

\section{Simulation}
\label{sec:sim}
Simulation scenarios for kernel target alignment evaluation of RF/GBT kernels were set up according to previously reported simulation benchmarks. They included
Friedman \cite{friedman}, Meier 1, Meier 2 \cite{meier},  van der Laan \cite{van2007super} and Checkerboard \cite{zhu2015}.

\subsection{Simulation Setup}
\label{sec:sim:sub1}
For each simulation scenario, the predictors were simulated from Uniform (Friedman, Meier 1, Meier 2, van der Laan) or Normal distributions (Checkerboard), respectively. 

Continuous targets were generated as $Y_i=f(\mbold{X_i})+\epsilon_i$. For the definitions of $f(\mbold{X_i})$ for each simulation case see below.

The five functional relationships $f(\mbold{X_i})$ between the predictors and target for different simulation settings are specified as follows.

1. Friedman. The setup for Friedman was as described in \cite{friedman}.
\begin{eqnarray}
X_{ij} &\sim& Uniform(0, 1), i=1,\ldots,n; j=1,\ldots,p \nonumber\\
\epsilon_i &\sim& N(0,1)\nonumber\\
f(\mbold{X_i}) &=& 10\sin{(\pi X_{i1}X_{i2})} + 20(X_{i3}-0.5)^2+10X_{i4} + 5X_{i5} + \epsilon_i\nonumber
\end{eqnarray}

2. Checkerboard. In addition to Friedman, we simulated data from a Checkerboard-like model with strong correlation as in Scenario 3 of \cite{zhu2015}.

\begin{eqnarray}
\mbold{X_i} &\sim& N(0, \Sigma_{p\times p}), i=1, \ldots, n \nonumber\\
\epsilon_i &\sim& N(0,1)\nonumber\\
f(\mbold{X_i})&=&2X_{i5}X_{i10}+2X_{i15}X_{i20} + \epsilon_i\nonumber
 \end{eqnarray}
The $(j,k)$ component of $\Sigma$ is equal to $0.9^{|j-k|}$.

3. van der Laan. The setup was studied in van der Laan et al. \cite{van2007super}.
\begin{eqnarray}
X_{ij} &\sim& Uniform(0, 1), i=1,\ldots,n; j=1,\ldots,p \nonumber\\
\epsilon_i &\sim& N(0,0.5)\nonumber\\
f(\mbold{X_i}) &=& \tilde{X}_{i1}\tilde{X}_{i2}+\tilde{X}_{i3}^2+\tilde{X}_{i8}\tilde{X}_{i10}-
\tilde{X}_{i6}^2+ \epsilon_i\nonumber\\
\tilde{X}_i &= &2(X_i-0.5)
\end{eqnarray}

4. Meier 1. This setup was investigated in  Meier et al. \cite{meier}.
\begin{eqnarray}
X_{ij} &\sim& Uniform(0, 1), i=1,\ldots,n; j=1,\ldots,p \nonumber\\
\epsilon_i &\sim& N(0,0.5)\nonumber\\
f(\mbold{X_i}) &=& -\sin(2\tilde{X}_{i1})+\tilde{X}_{i2}^2+\tilde{X}_{i3}-\exp(\tilde{X}_{i4})+ \epsilon_i\nonumber\\
\tilde{X}_i &= &2(X_i-0.5)
\end{eqnarray}

5. Meier 2. This setup was investigated in  Meier et al. \cite{meier} as well.
\begin{eqnarray}
X_{ij} &\sim& Uniform(0, 1), i=1,\ldots,n; j=1,\ldots,p \nonumber\\
\epsilon_i &\sim& N(0,0.5)\nonumber\\
f(\mbold{X_i}) &=& -\tilde{X}_{i1}+(2\tilde{X}_{i2}-1)^2+\frac{\sin(2\pi\tilde{X}_{i3})}
{2-\sin(2\pi\tilde{X}_{i4})}+2\cos(2\pi\tilde{X}_{i4})+4\cos^2(2\pi\tilde{X}_{i4})+ \epsilon_i\nonumber\\
\tilde{X}_i &= &2(X_i-0.5)
\end{eqnarray}

For each functional relationship $f(\mbold{X_i})$ (Friedman, Checkerboard, Meier 1, Meier 2, and van der Laan), we simulated data from four scenarios with different samples sizes n = 800 and n = 1600 and number
of features p = 20 and p = 40. Within each scenario, we simulated 200 data sets, and for each data set we randomly chose
75\% of samples as training data and remaining 25\% as test data. We repeated the analysis 200 times to evaluate the kernel-target alignment of RF and XGB kernels on the training sets and its relationship with the respective kernel algorithm performance on the independent test sets.

\subsection{Simulation Results}

\begin{figure}
     \centering
     \begin{minipage}[b]{0.45\textwidth}
         \centering
         \includegraphics[width=\textwidth]{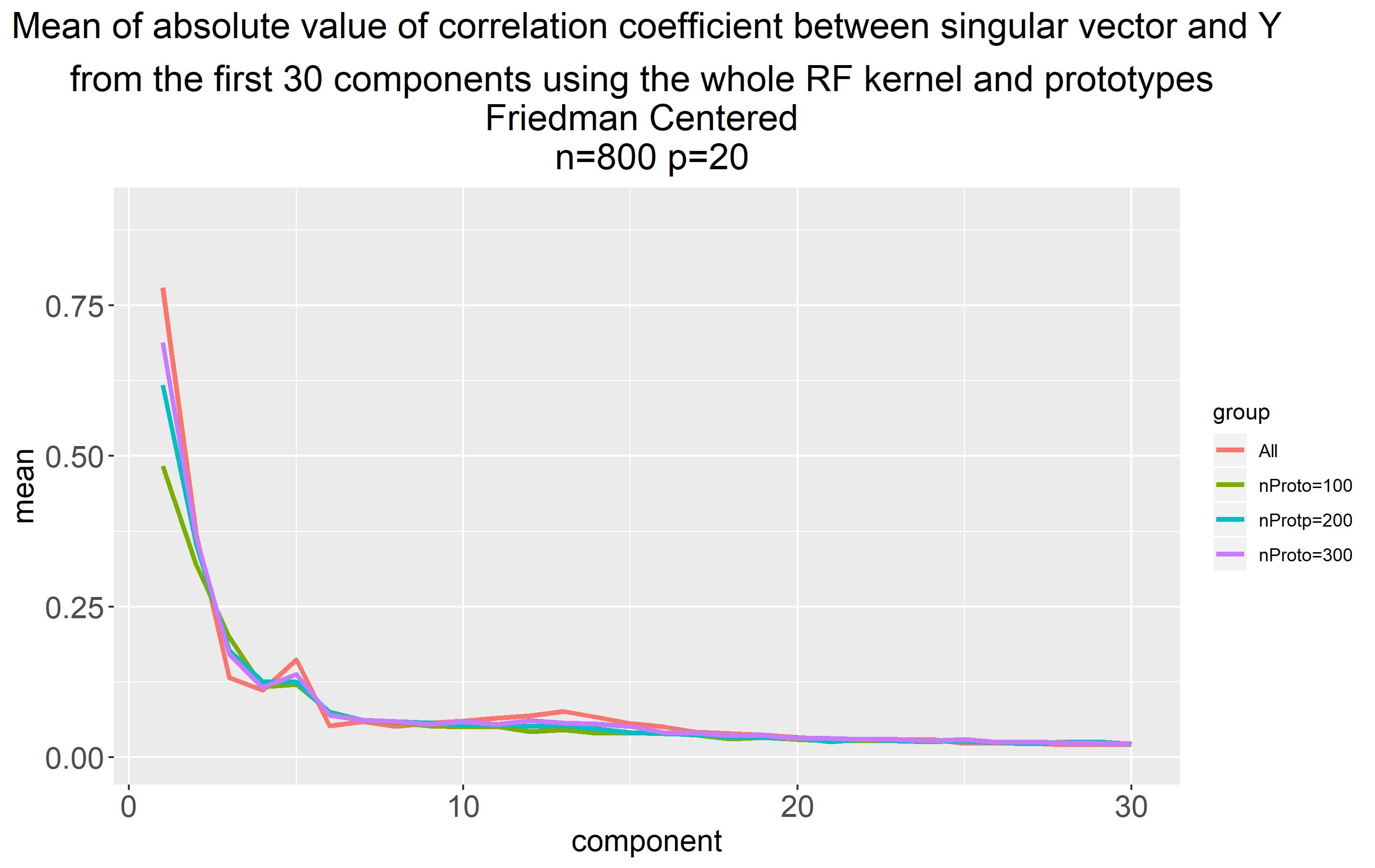}
         \subcaption{}
     \end{minipage}
     \hfill
     \begin{minipage}[b]{0.45\textwidth}
         \centering
         \includegraphics[width=\textwidth]{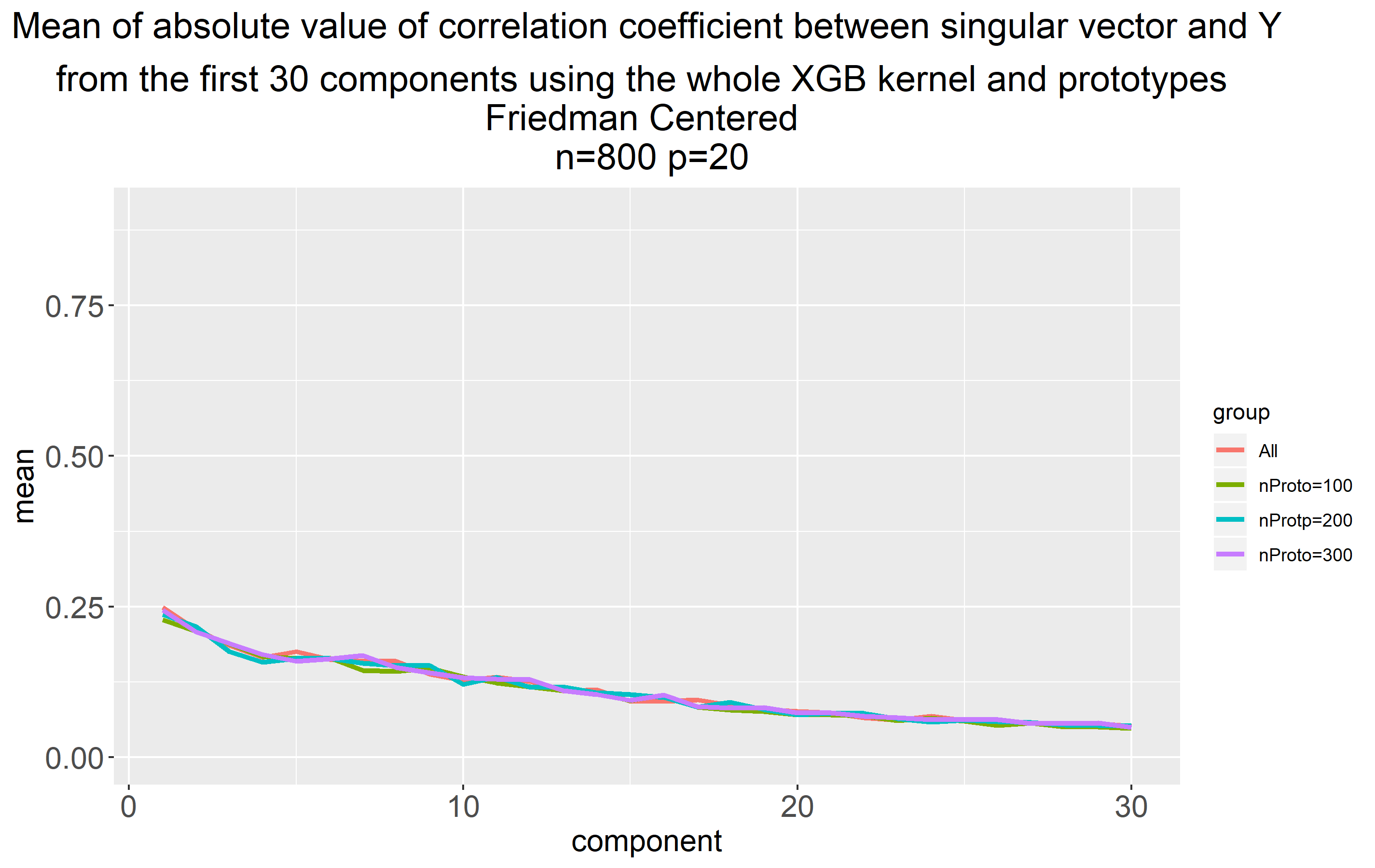}
             \subcaption{}
     \end{minipage}
     \\
     \begin{minipage}[b]{0.45\textwidth}
         \centering
         \includegraphics[width=\textwidth]{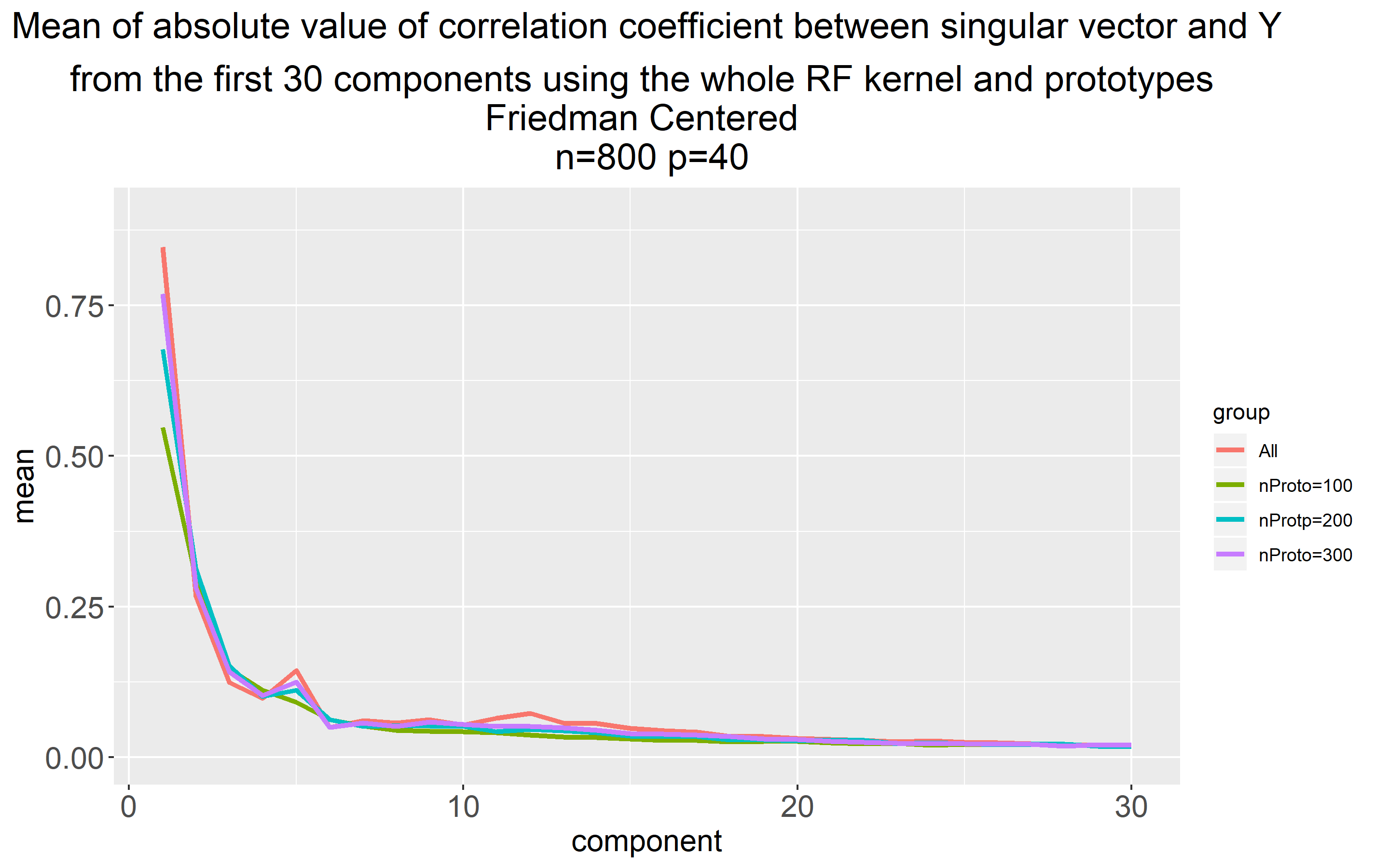}
           \subcaption{}
     \end{minipage}
     \hfill
     \begin{minipage}[b]{0.45\textwidth}
         \centering
         \includegraphics[width=\textwidth]{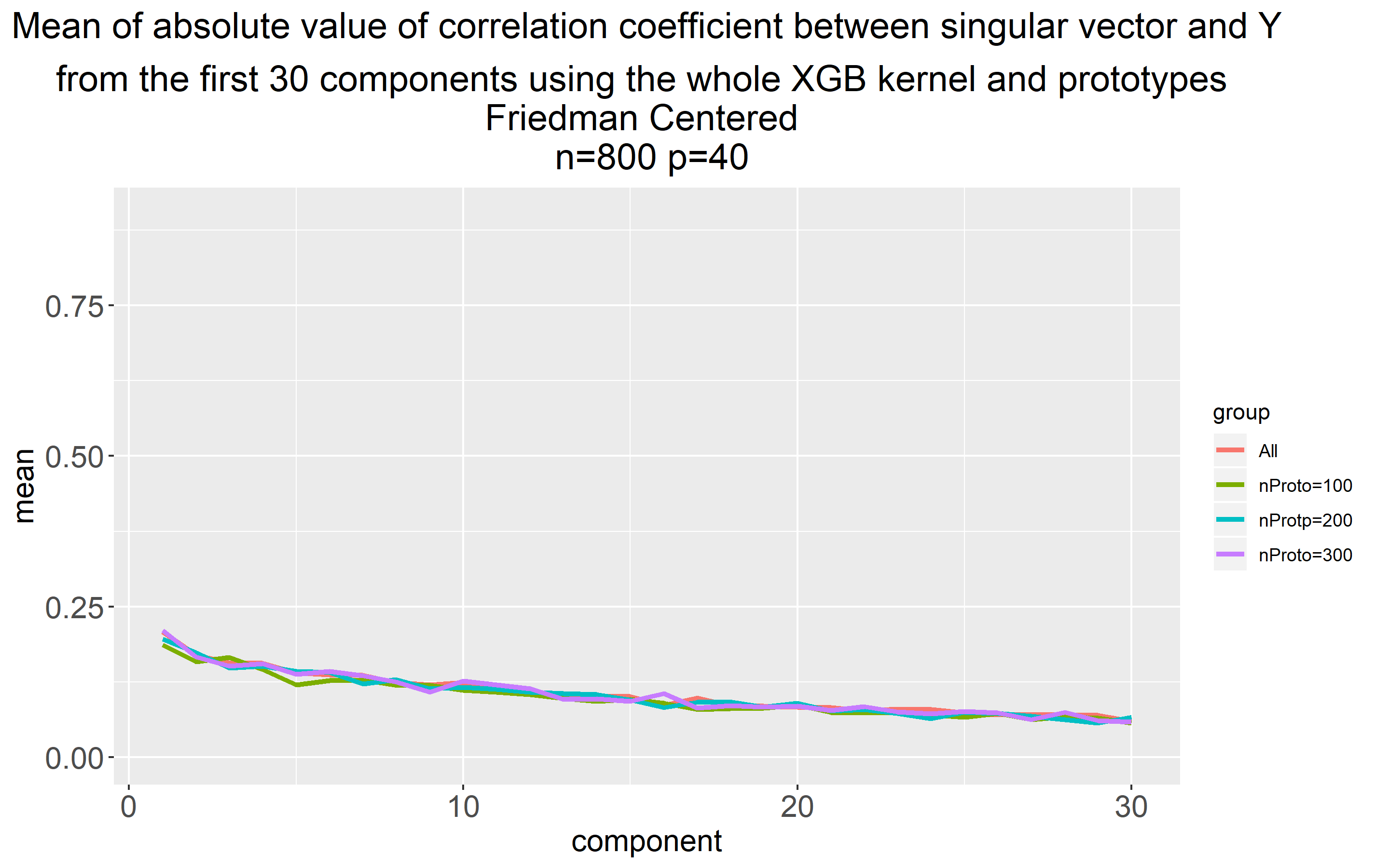}
          \subcaption{}
      \end{minipage}
           \begin{minipage}[b]{0.45\textwidth}
         \centering
         \includegraphics[width=\textwidth]{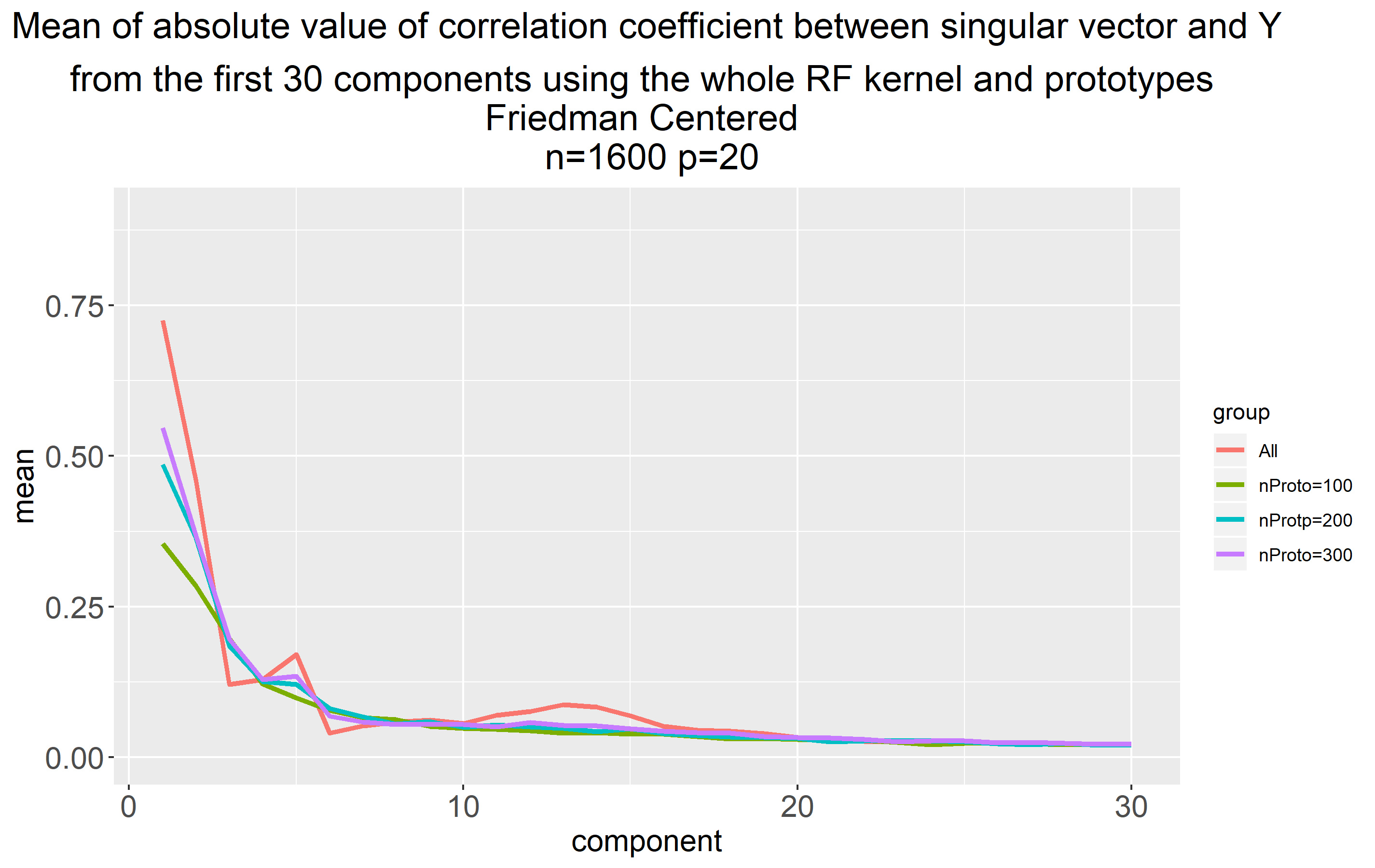}
        \subcaption{}
     \end{minipage}
     \hfill
     \begin{minipage}[b]{0.45\textwidth}
         \centering
         \includegraphics[width=\textwidth]{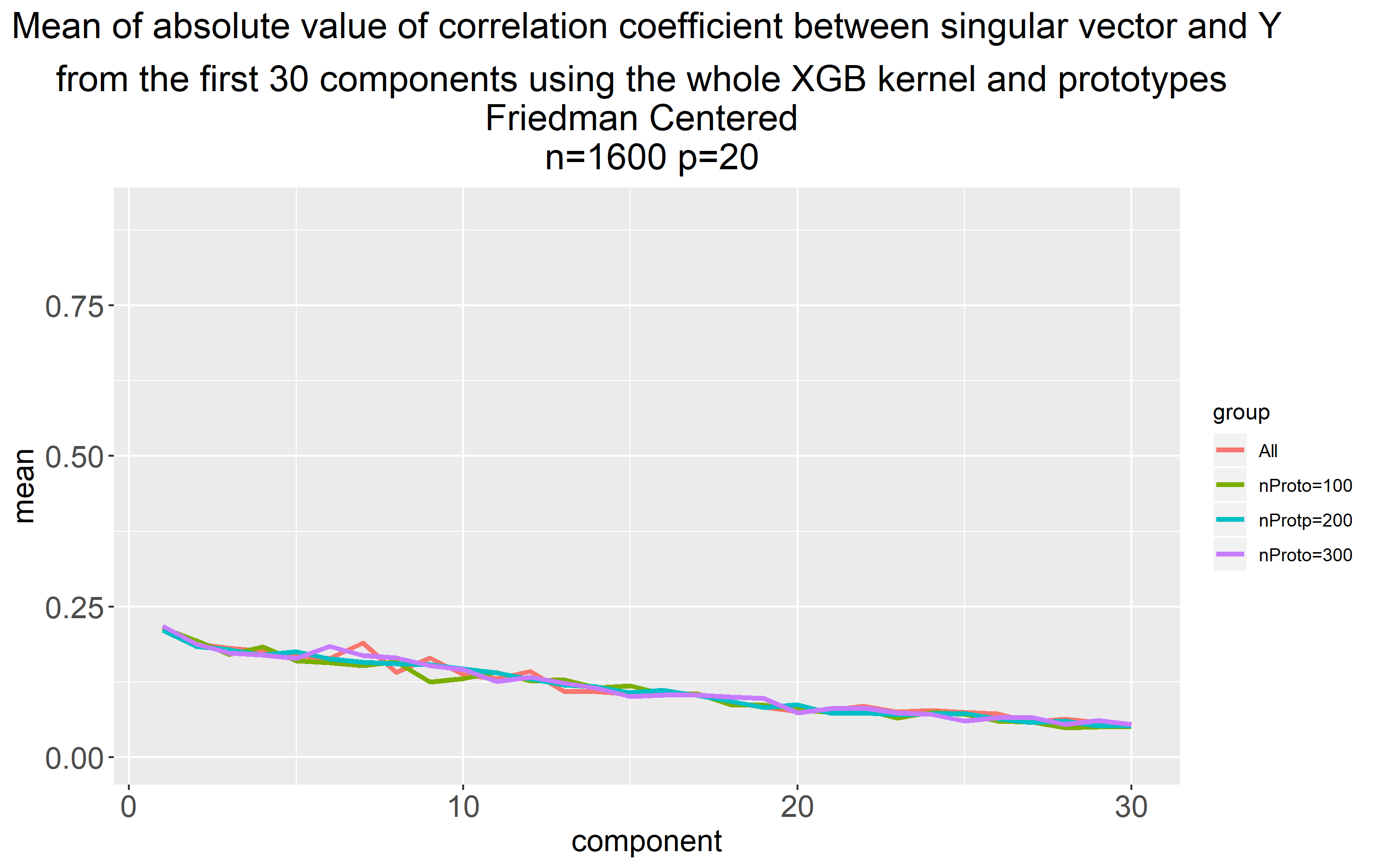}
        \subcaption{}
     \end{minipage}
     \\
     \begin{minipage}[b]{0.45\textwidth}
         \centering
         \includegraphics[width=\textwidth]{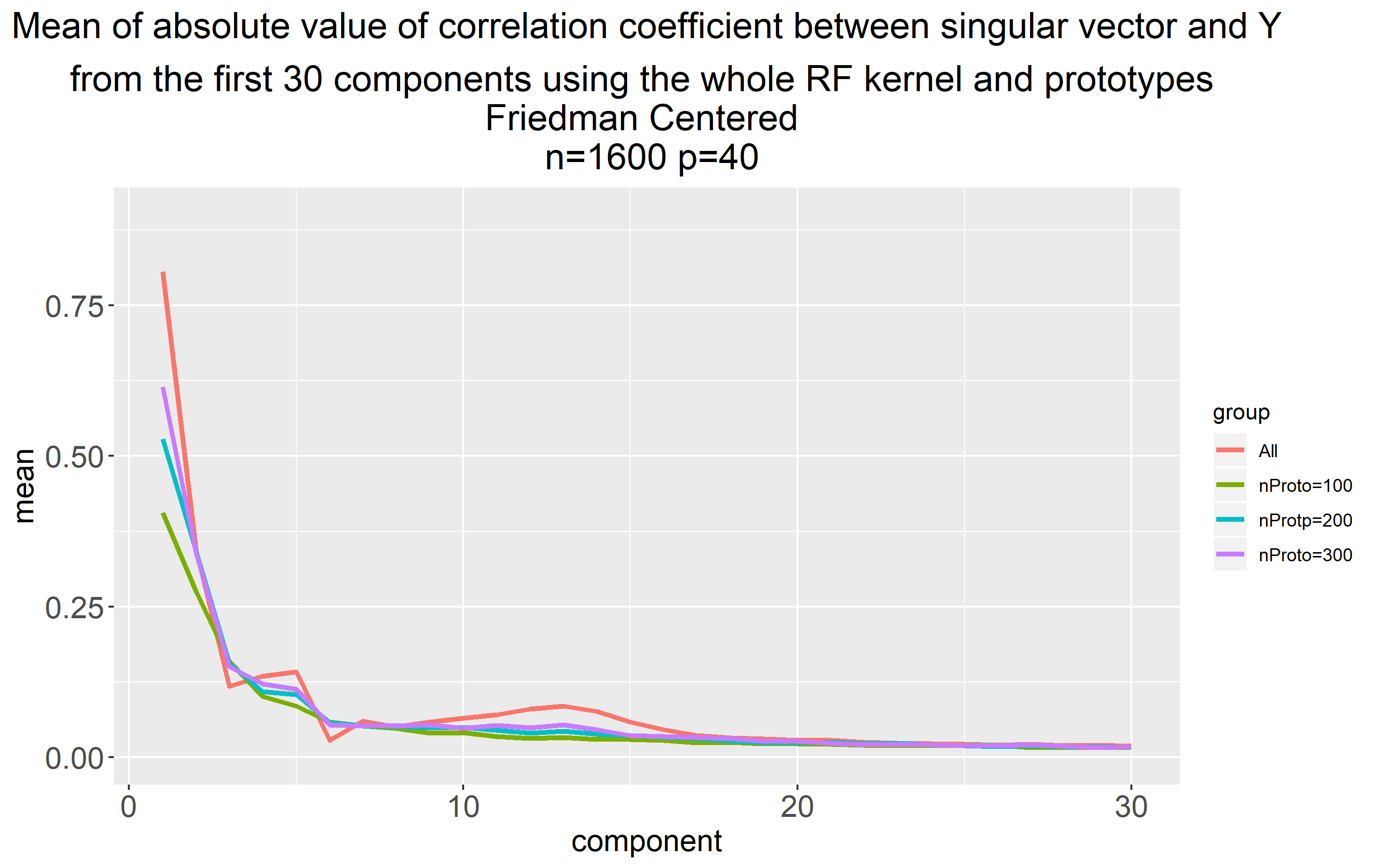}
           \subcaption{}
     \end{minipage}
     \hfill
     \begin{minipage}[b]{0.45\textwidth}
         \centering
         \includegraphics[width=\textwidth]{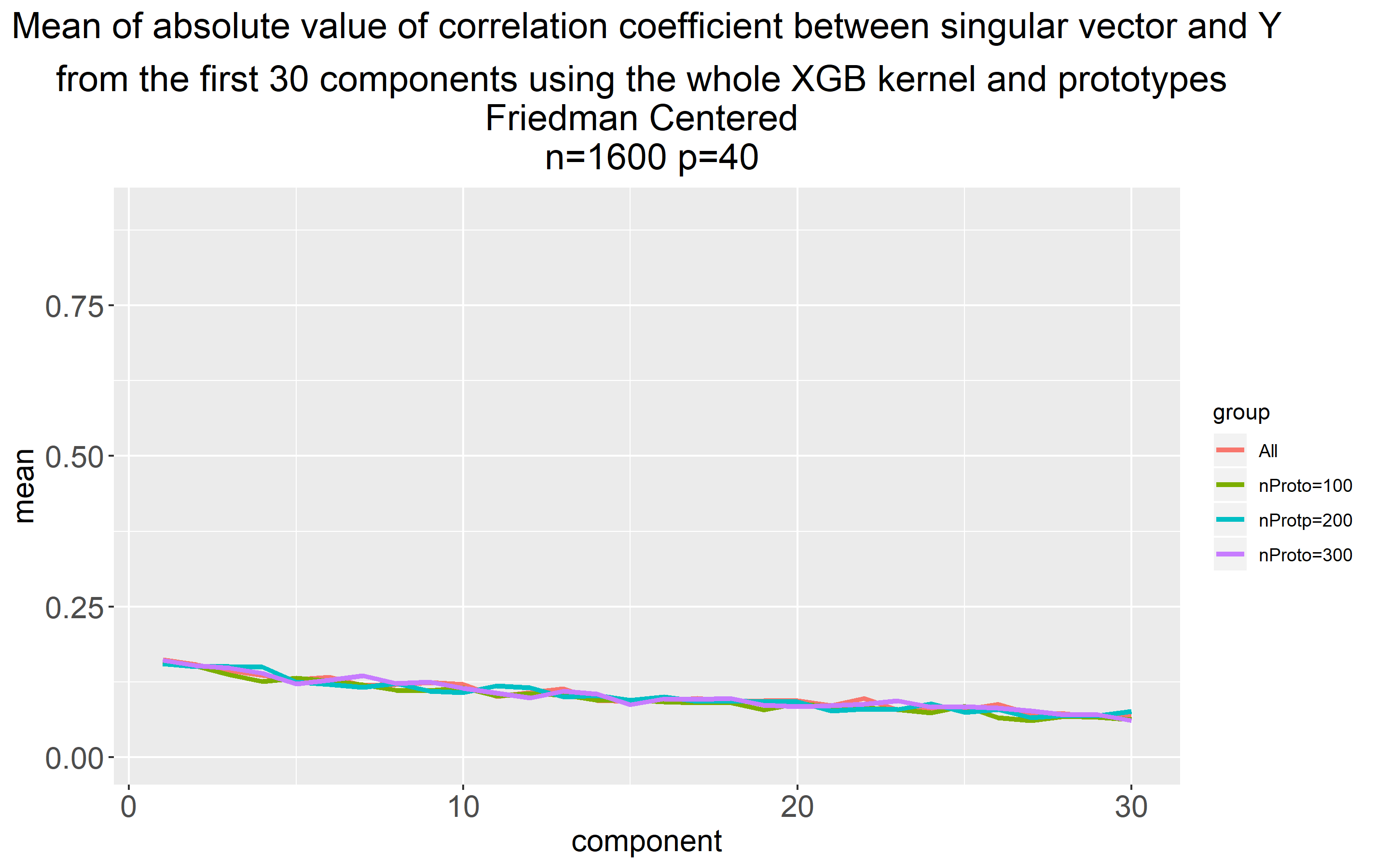}
            \subcaption{}
      \end{minipage}

        \caption{Kernel-target alignment spectra and relevant dimensionality---Friedman}
        \label{fig:rd-friedman}
\end{figure}

\begin{figure}
     \centering
     \begin{minipage}[b]{0.45\textwidth}
         \centering
         \includegraphics[width=\textwidth]{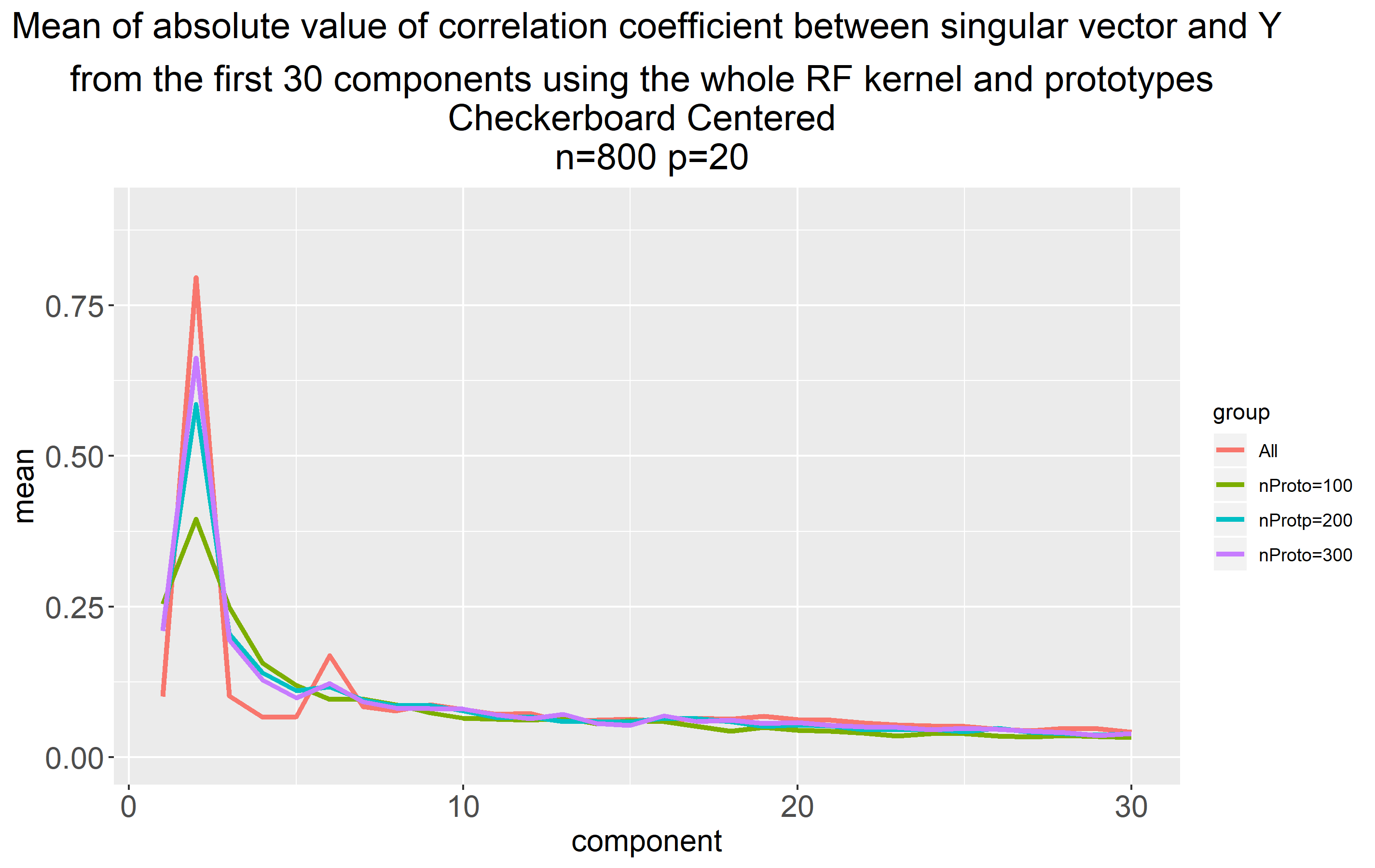}
         \subcaption{}
     \end{minipage}
     \hfill
     \begin{minipage}[b]{0.45\textwidth}
         \centering
         \includegraphics[width=\textwidth]{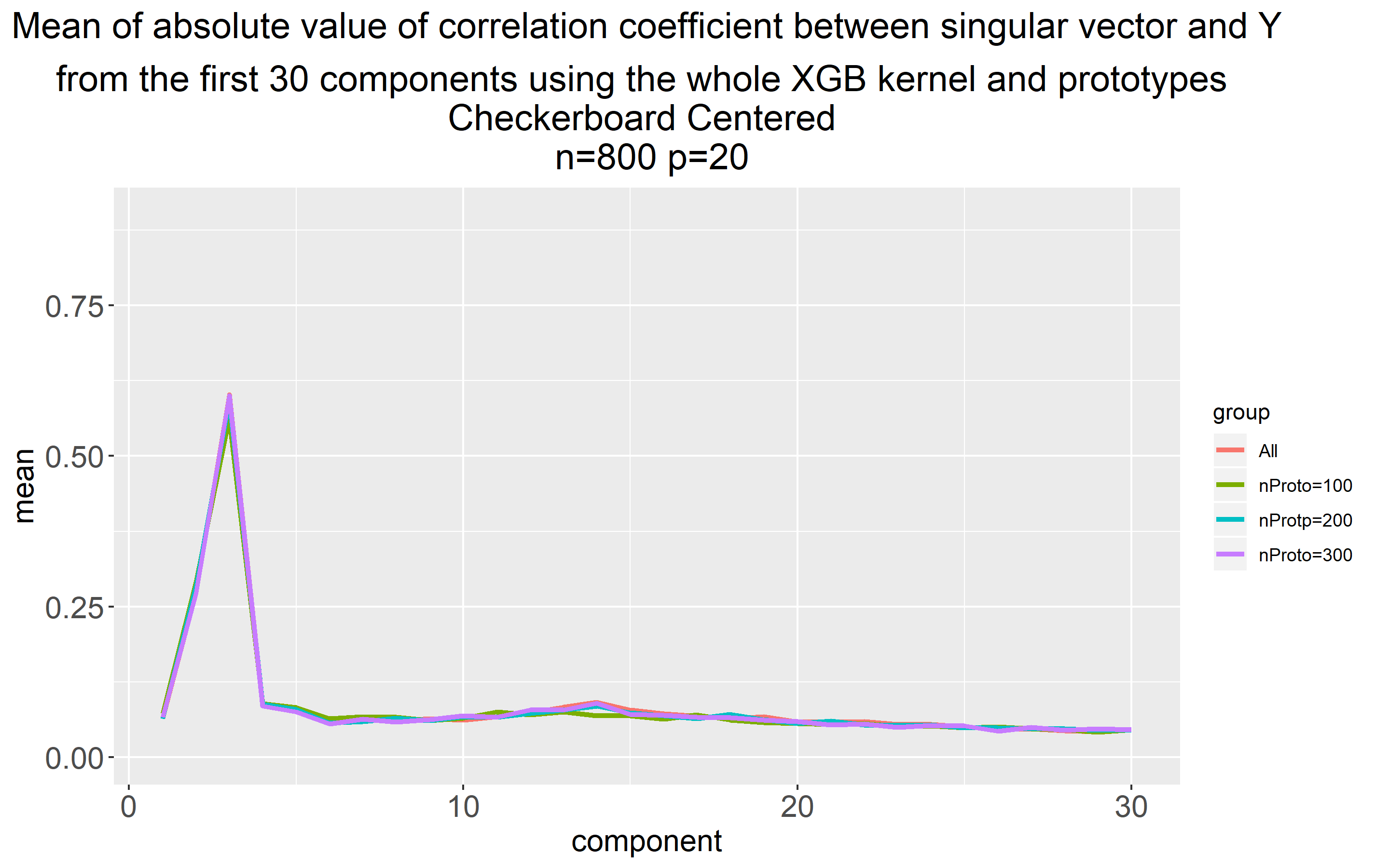}
         \subcaption{}
        \end{minipage}
     \\
     \begin{minipage}[b]{0.45\textwidth}
         %\centering
         \includegraphics[width=\textwidth]{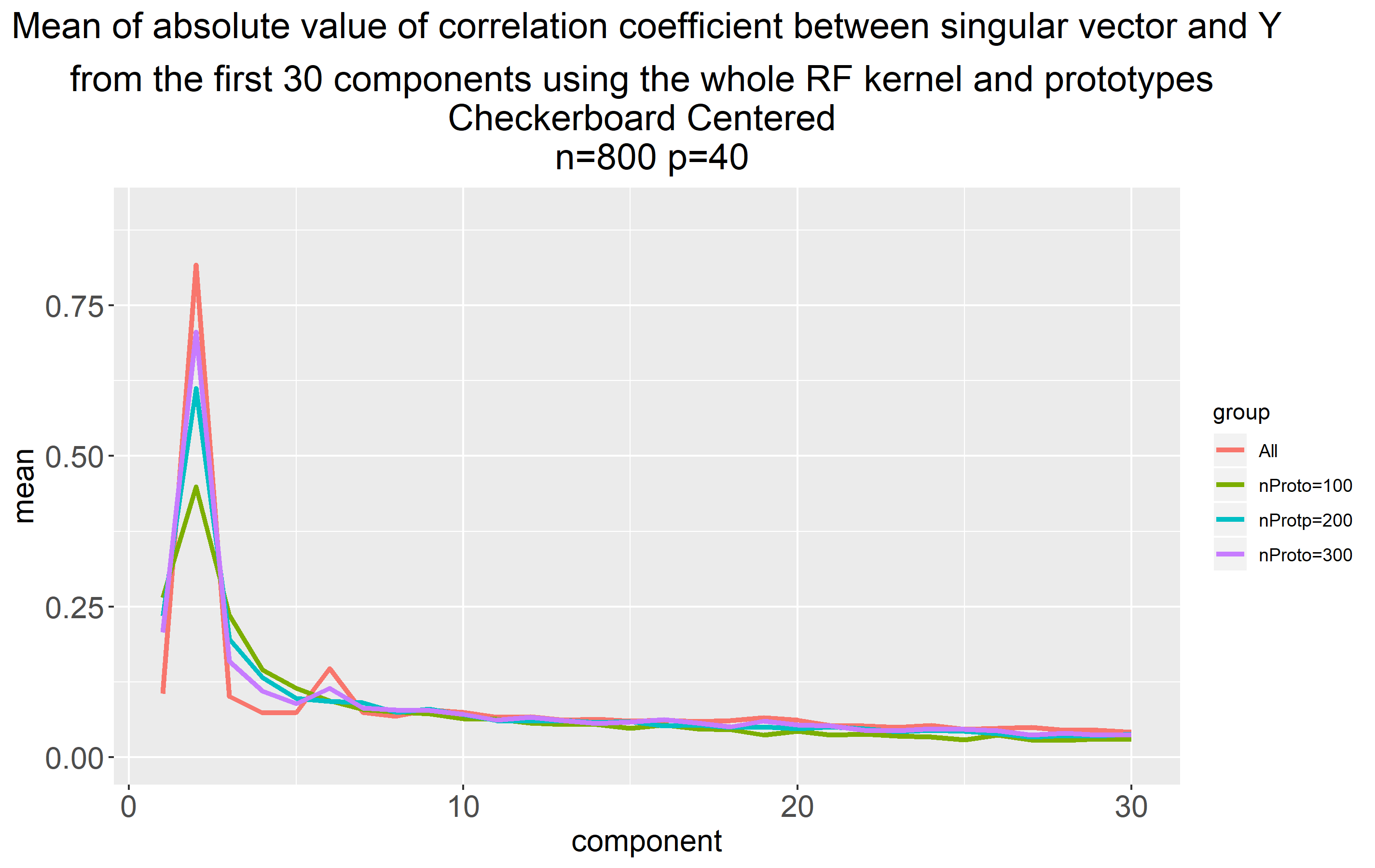}
           \subcaption{}
     \end{minipage}
     \hfill
     \begin{minipage}[b]{0.45\textwidth}
         %\centering
         \includegraphics[width=\textwidth]{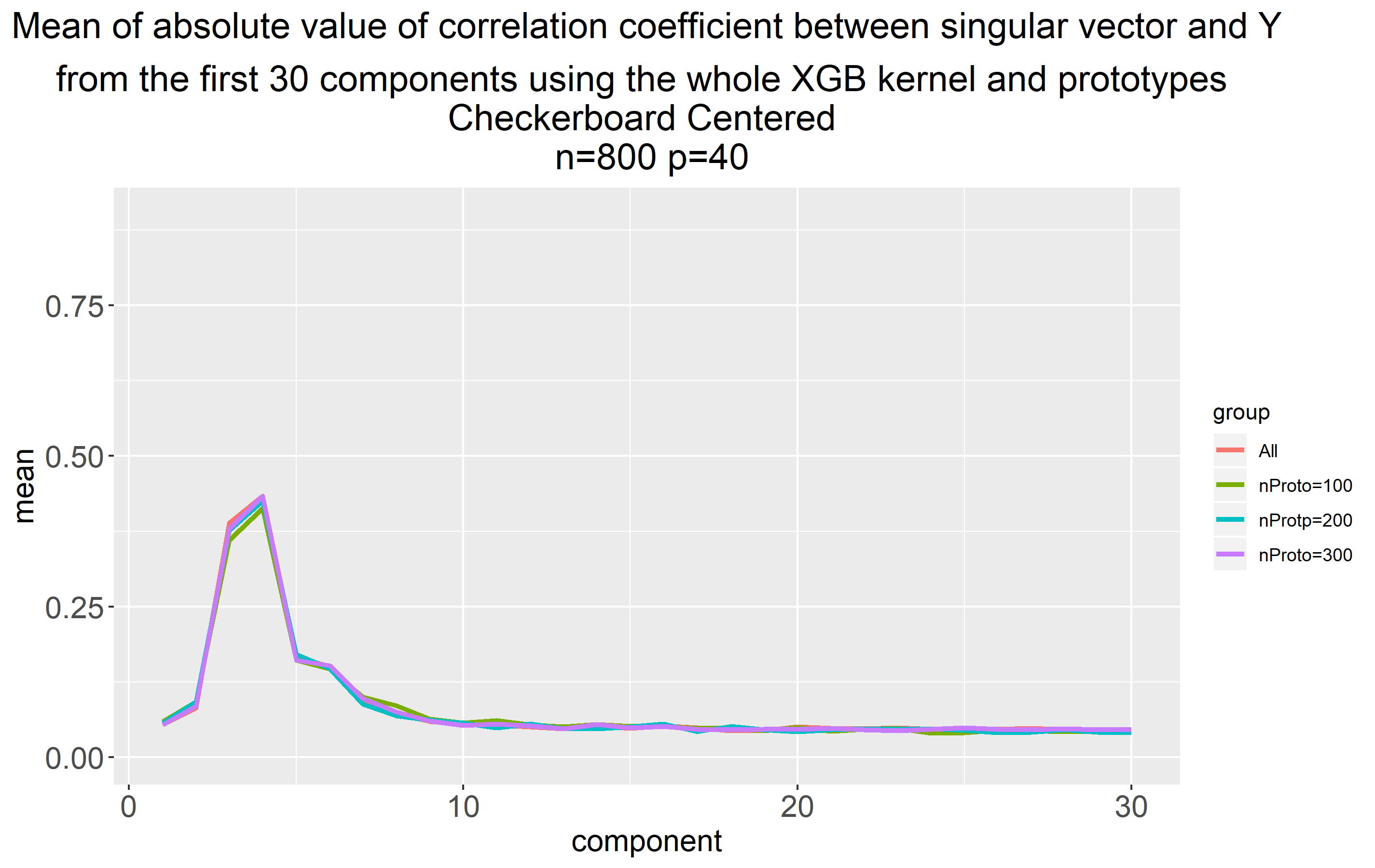}
          \subcaption{}
      \end{minipage}
           \begin{minipage}[b]{0.45\textwidth}
         %\centering
         \includegraphics[width=\textwidth]{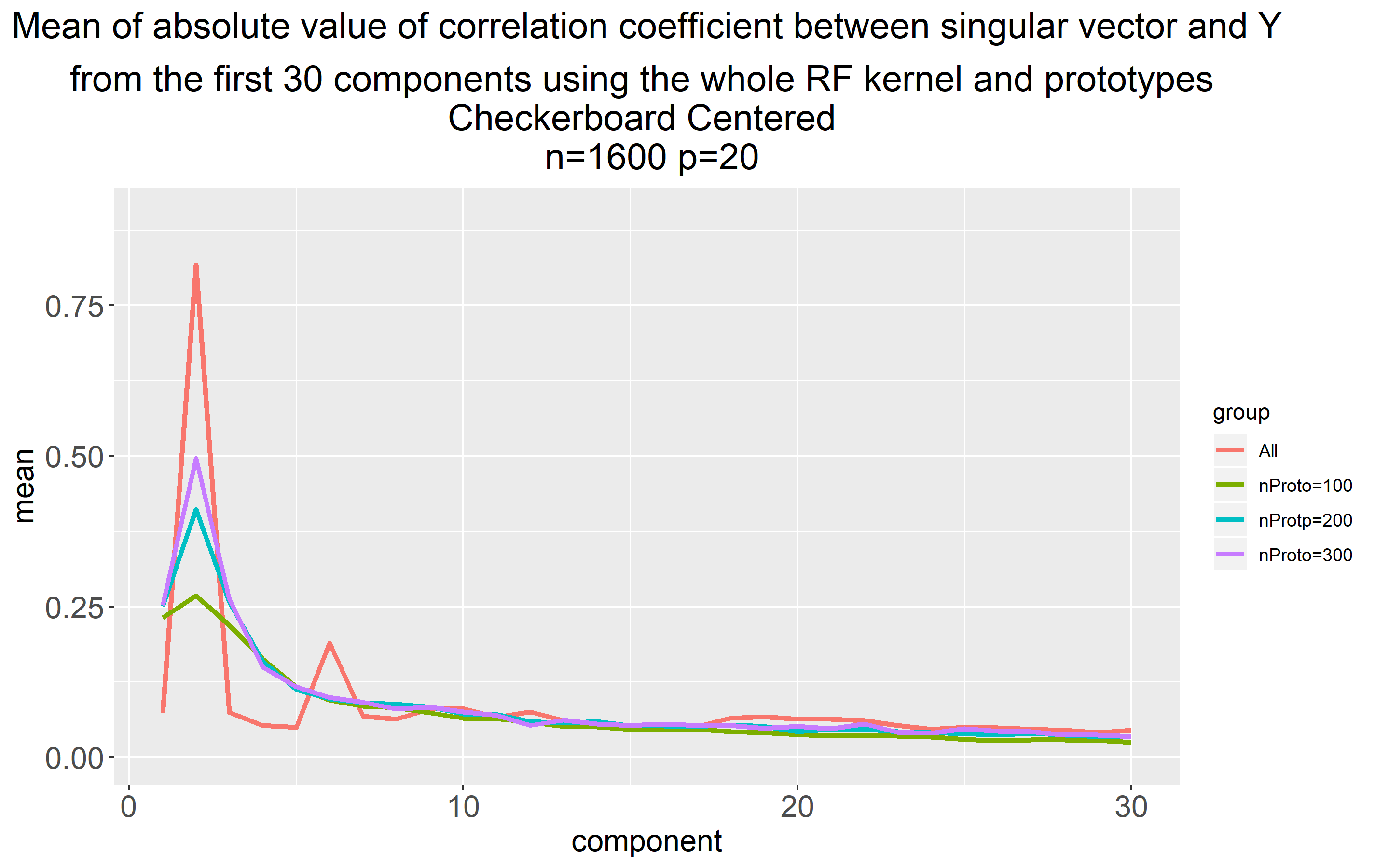}
          \subcaption{}
     \end{minipage}
     \hfill
     \begin{minipage}[b]{0.45\textwidth}
         %\centering
         \includegraphics[width=\textwidth]{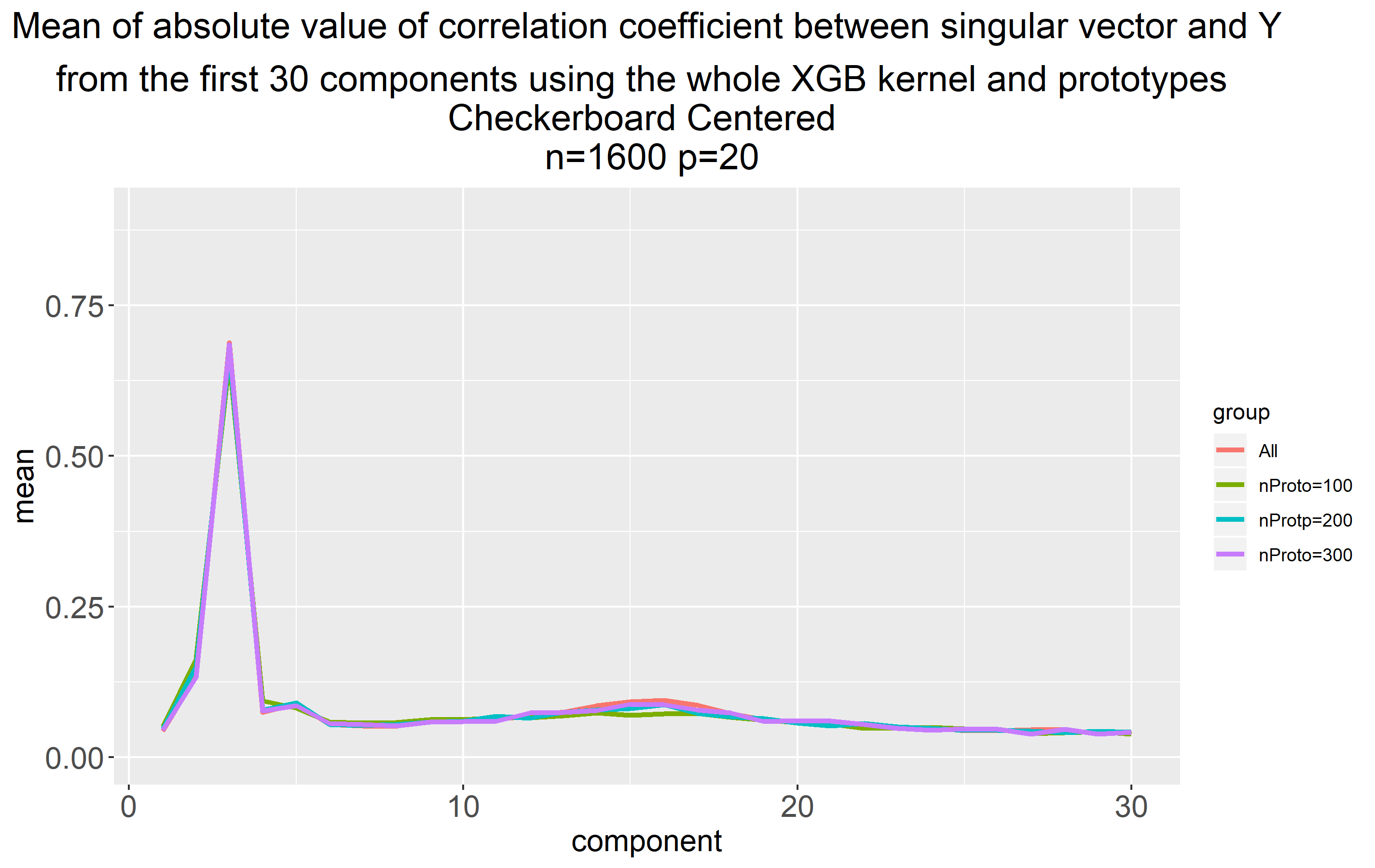}
           \subcaption{}
     \end{minipage}
     \\
     \begin{minipage}[b]{0.45\textwidth}
         %\centering
         \includegraphics[width=\textwidth]{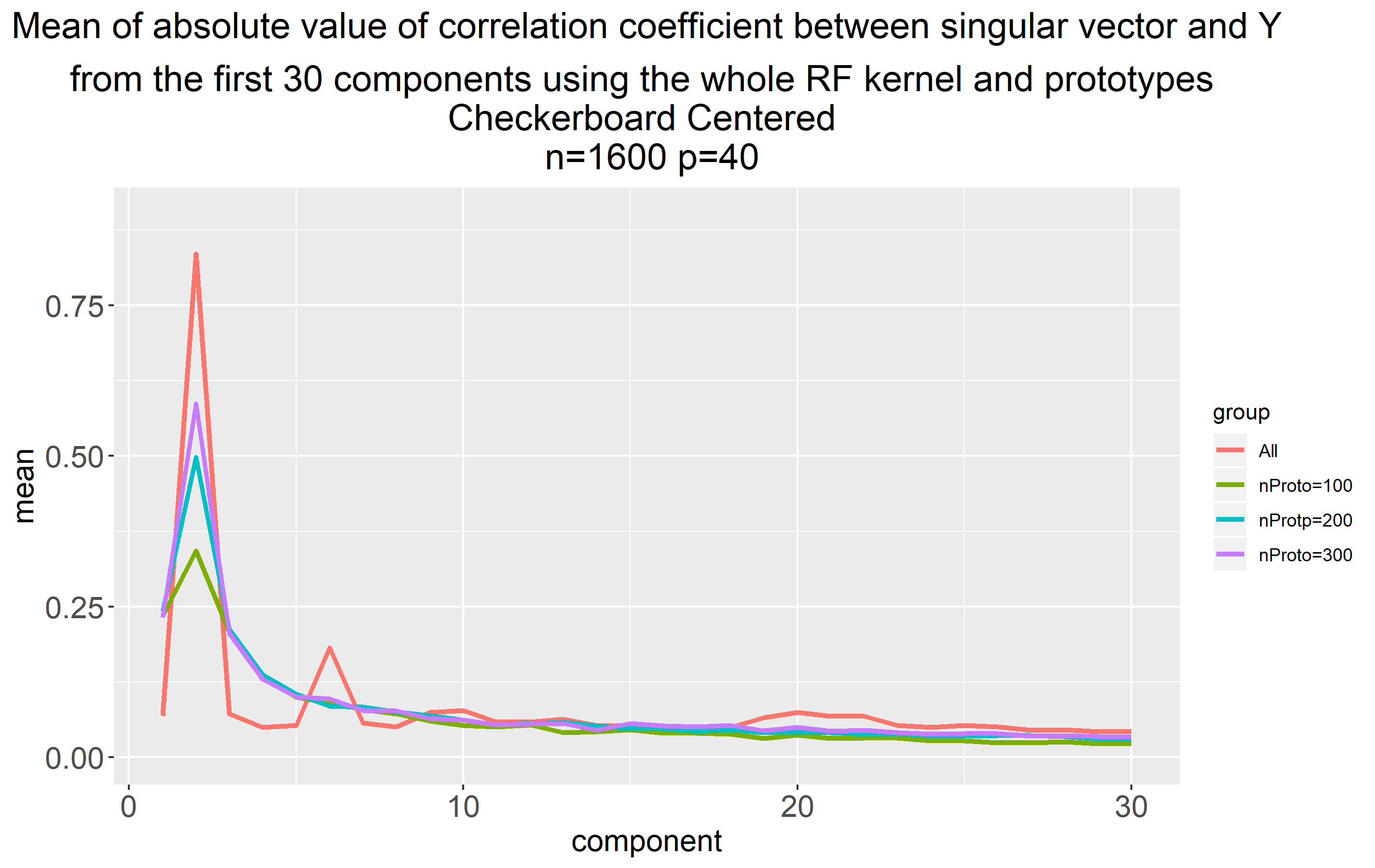}
          \subcaption{}
     \end{minipage}
     \hfill
     \begin{minipage}[b]{0.45\textwidth}
         %\centering
         \includegraphics[width=\textwidth]{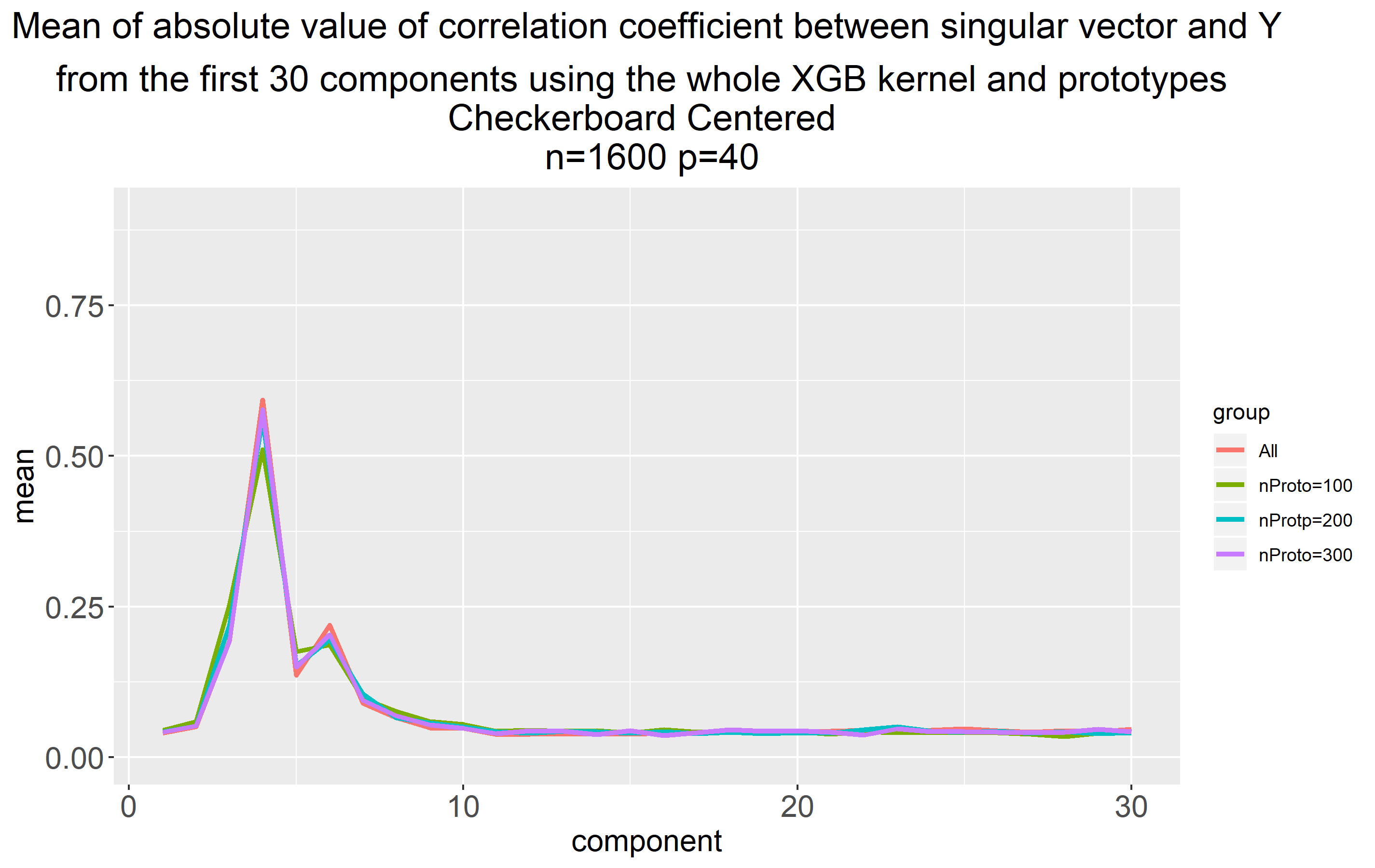}
            \subcaption{}
      \end{minipage}

        \caption{Kernel-target alignment spectra and relevant dimensionality---Checkerboard}
        \label{fig:rd-checkerboard}
\end{figure}

\begin{figure}
     \centering
     \begin{minipage}[b]{0.45\textwidth}
         \centering
         \includegraphics[width=\textwidth]{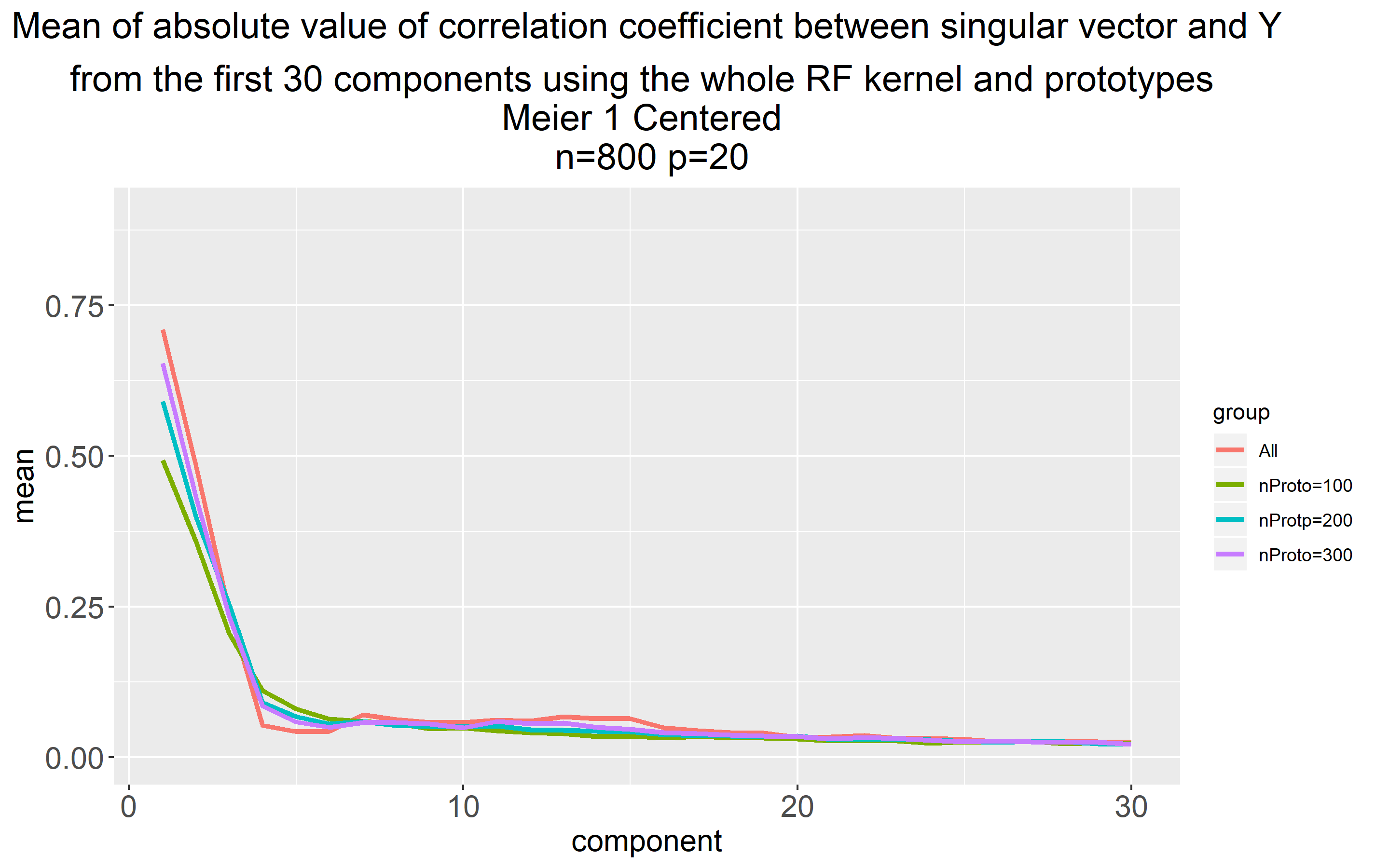}
          \subcaption{}
     \end{minipage}
     \hfill
     \begin{minipage}[b]{0.45\textwidth}
         \centering
         \includegraphics[width=\textwidth]{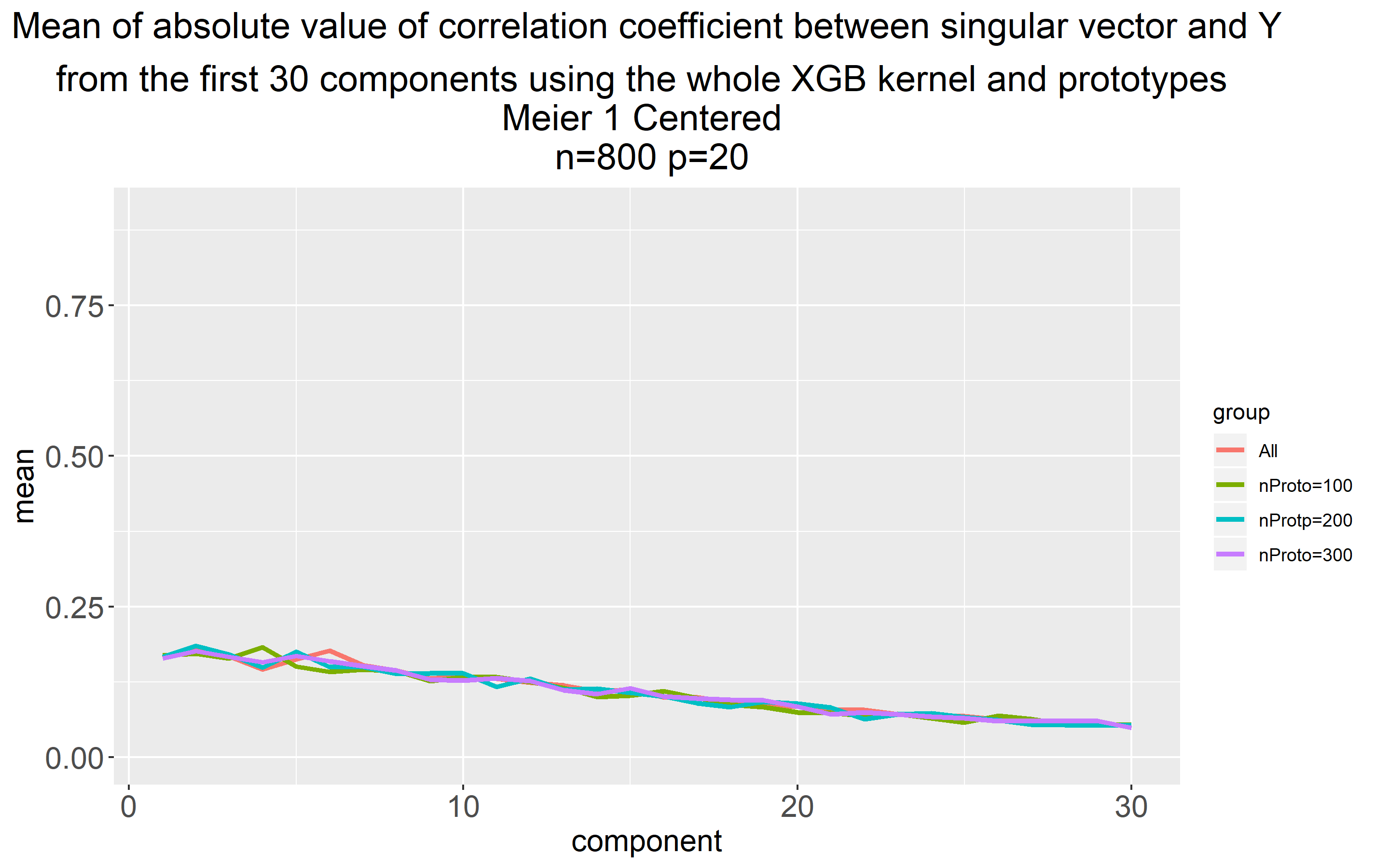}
            \subcaption{}
     \end{minipage}
     \\
     \begin{minipage}[b]{0.45\textwidth}
         \centering
         \includegraphics[width=\textwidth]{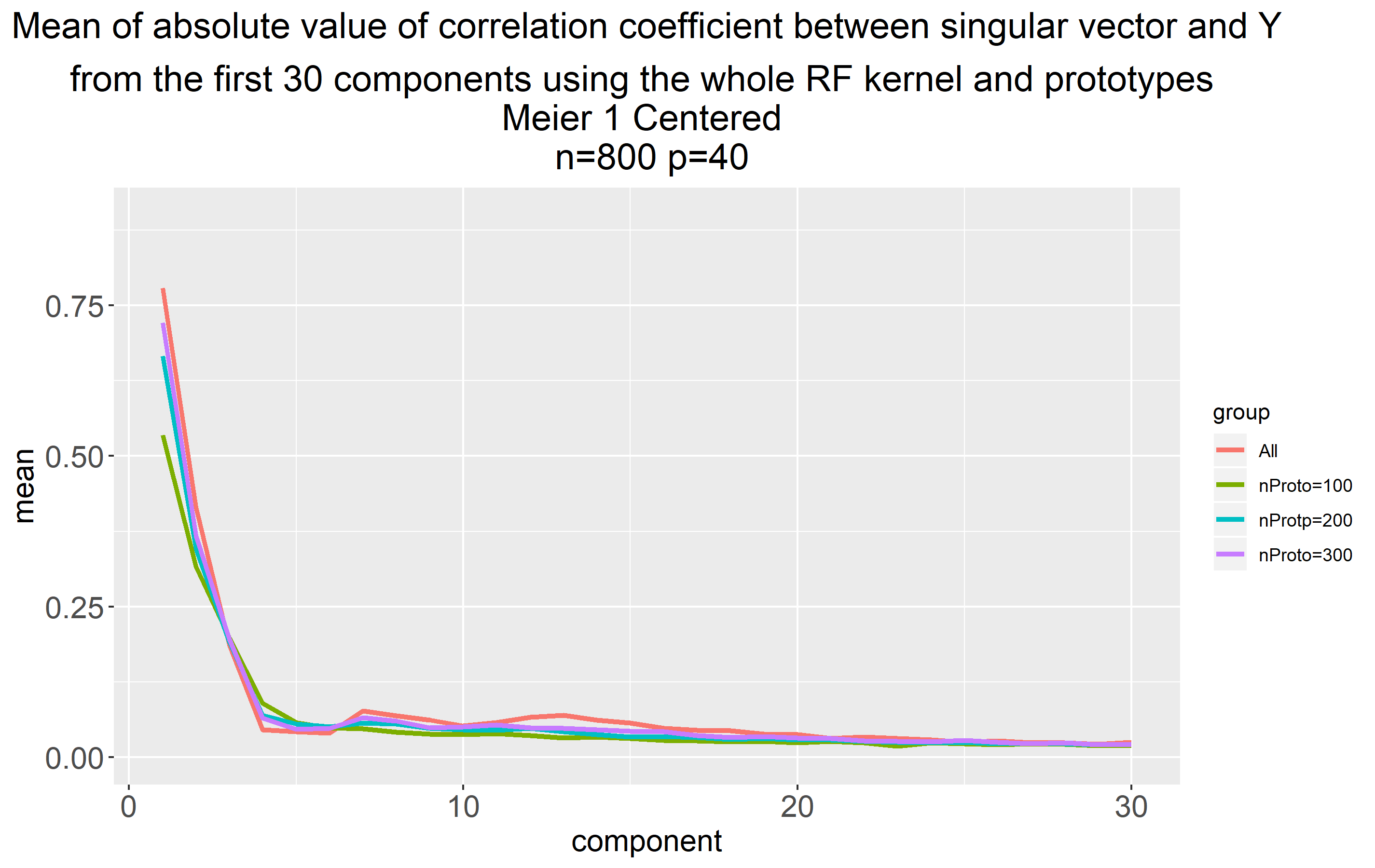}
          \subcaption{}
     \end{minipage}
     \hfill
     \begin{minipage}[b]{0.45\textwidth}
         \centering
         \includegraphics[width=\textwidth]{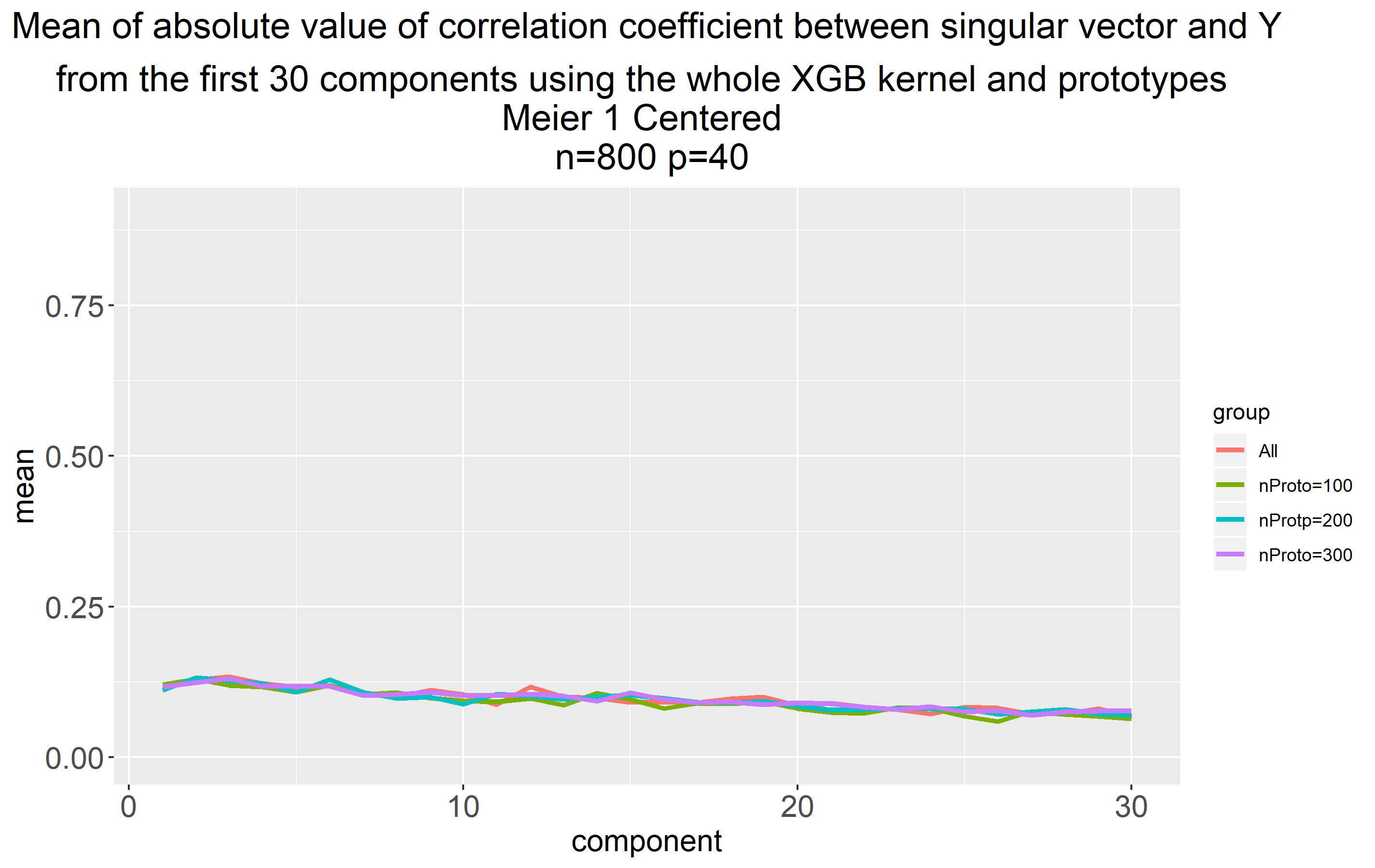}
         \subcaption{}
      \end{minipage}
           \begin{minipage}[b]{0.45\textwidth}
         \centering
         \includegraphics[width=\textwidth]{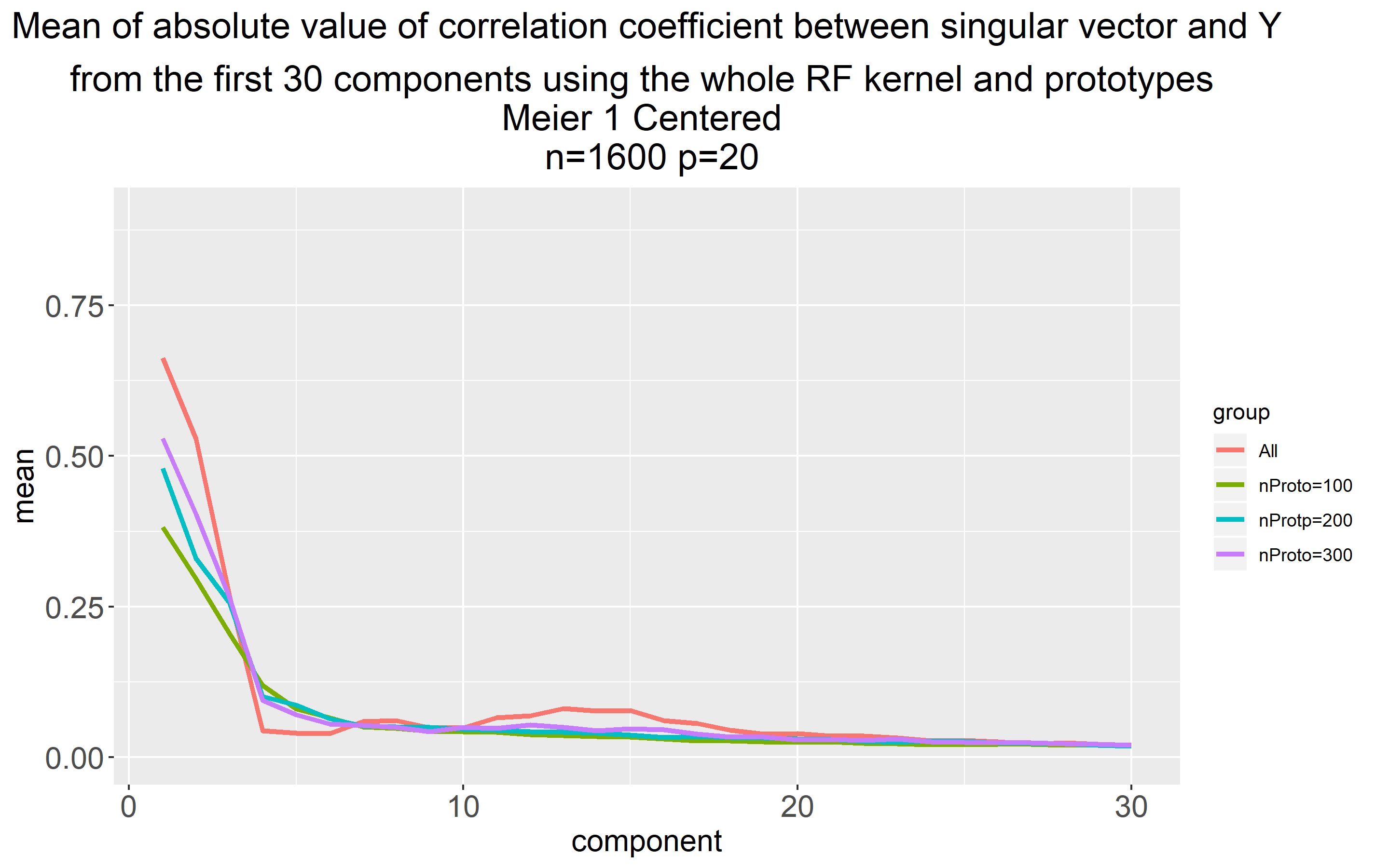}
        \subcaption{}
     \end{minipage}
     \hfill
     \begin{minipage}[b]{0.45\textwidth}
         \centering
         \includegraphics[width=\textwidth]{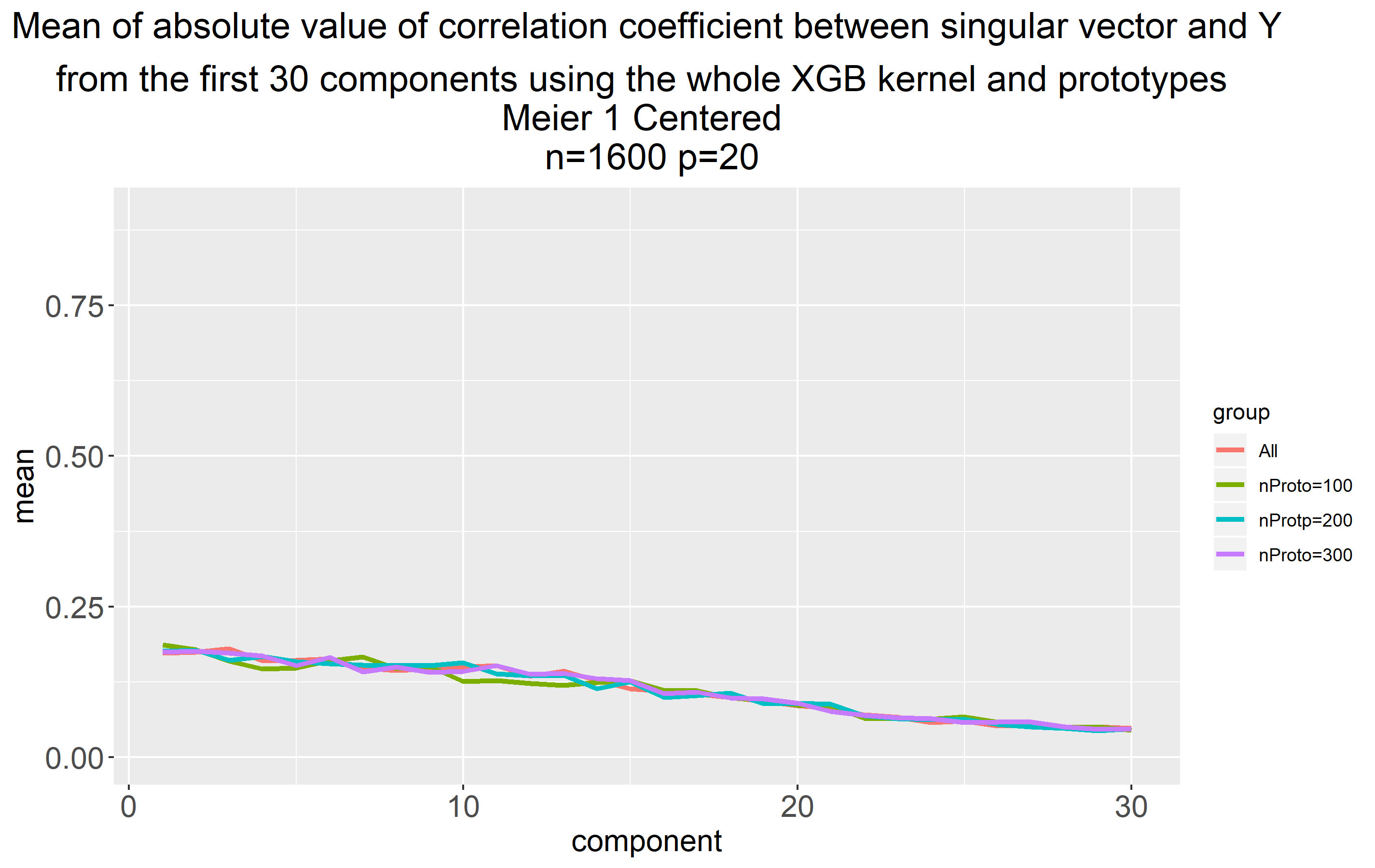}
     \subcaption{}
     \end{minipage}
     \\
     \begin{minipage}[b]{0.45\textwidth}
         \centering
         \includegraphics[width=\textwidth]{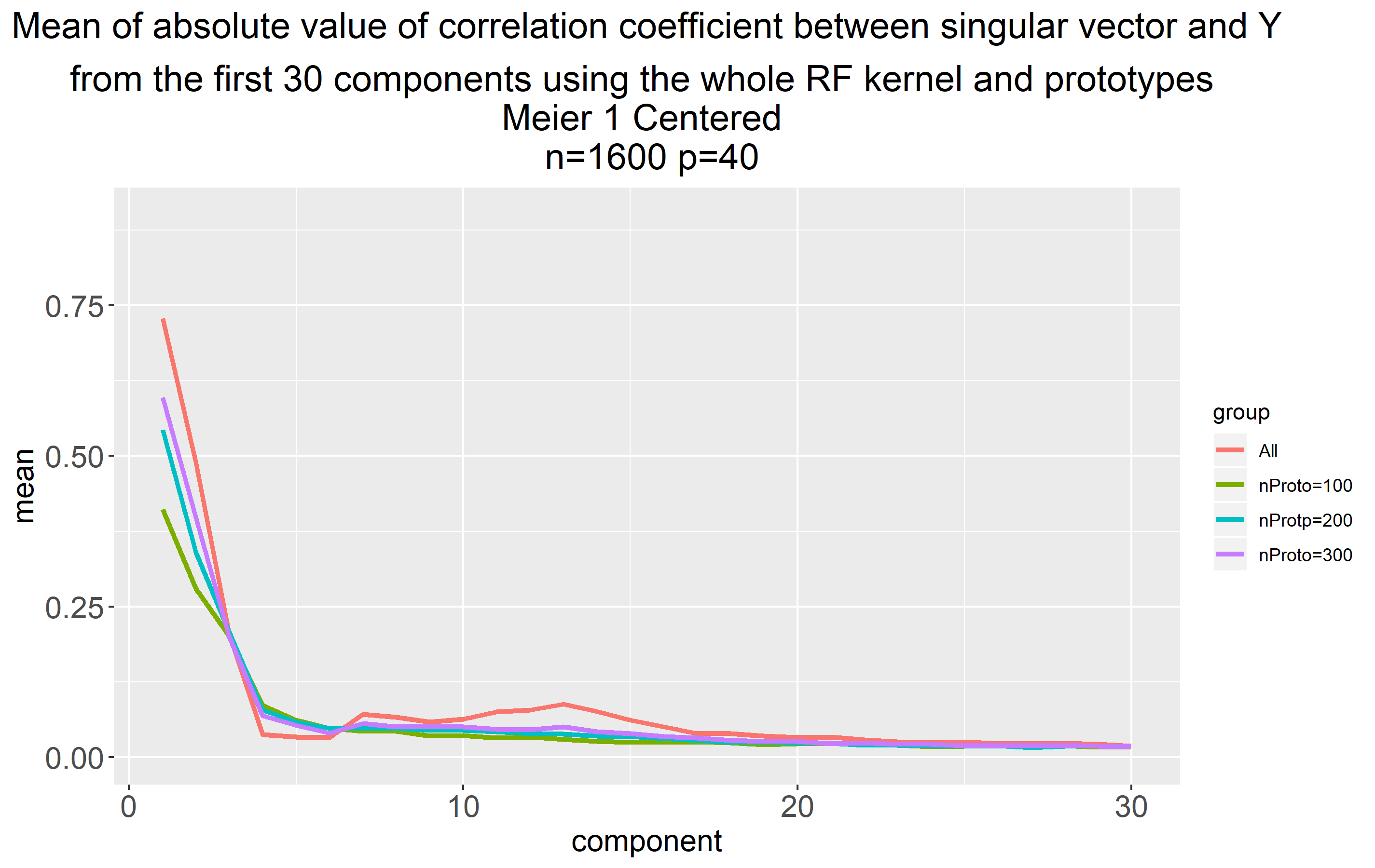}
           \subcaption{}
     \end{minipage}
     \hfill
     \begin{minipage}[b]{0.45\textwidth}
         \centering
         \includegraphics[width=\textwidth]{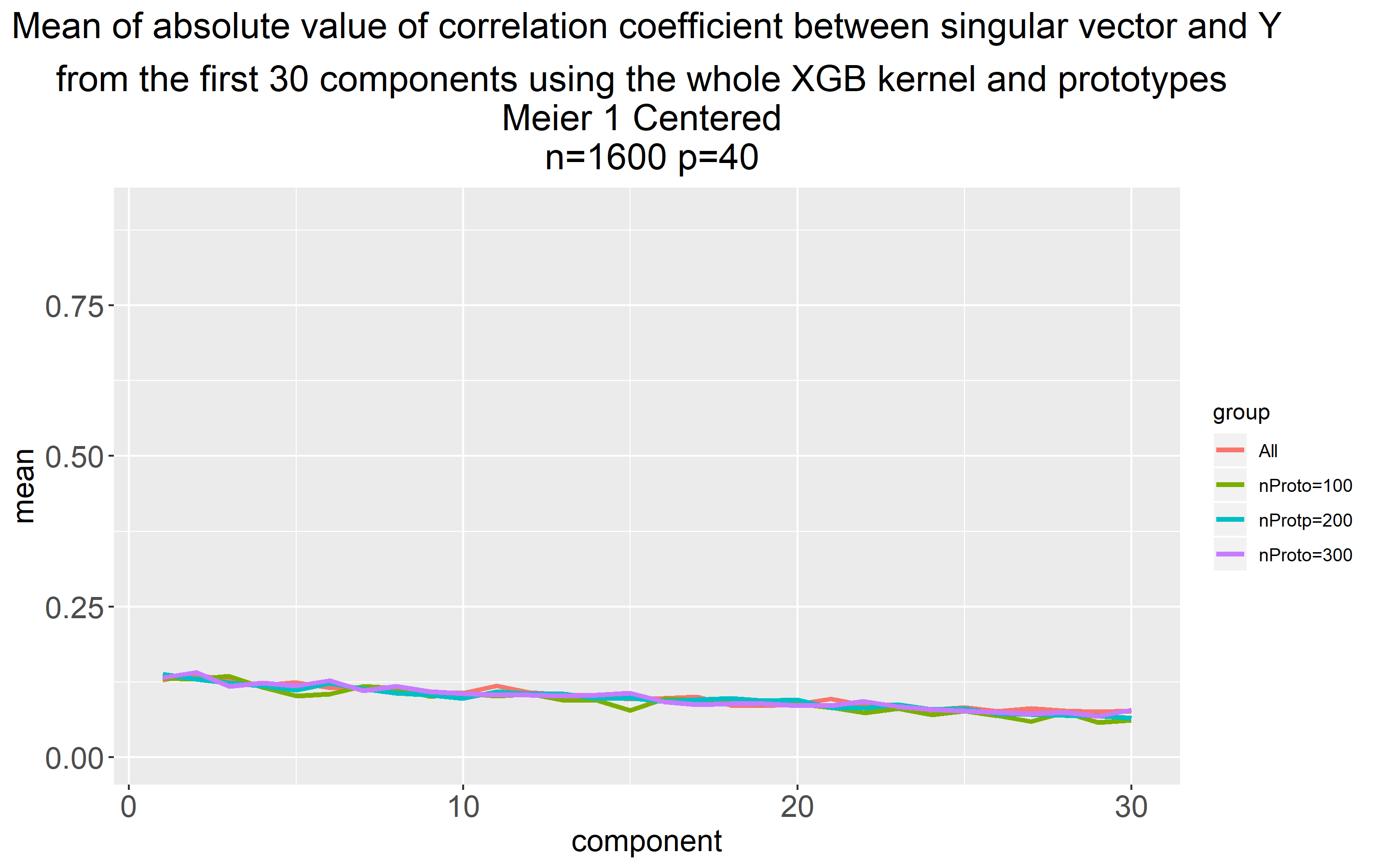}
          \subcaption{}
      \end{minipage}

        \caption{Kernel-target alignment spectra and relevant dimensionality---Meier 1}
        \label{fig:rd-meier1}
\end{figure}

\begin{figure}
     \centering
     \begin{minipage}[b]{0.45\textwidth}
         \centering
         \includegraphics[width=\textwidth]{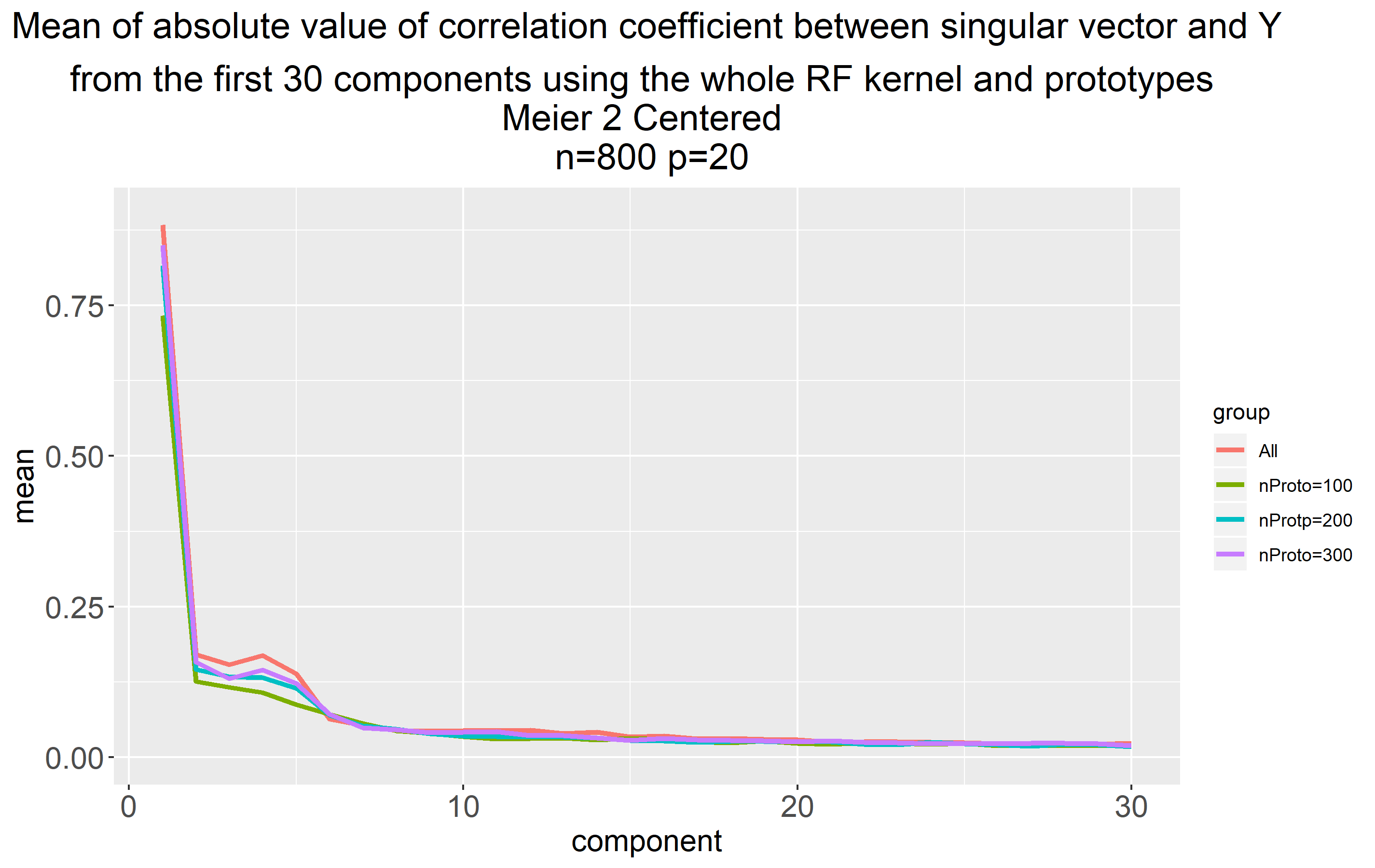}
       \subcaption{}
     \end{minipage}
     \hfill
     \begin{minipage}[b]{0.45\textwidth}
         \centering
         \includegraphics[width=\textwidth]{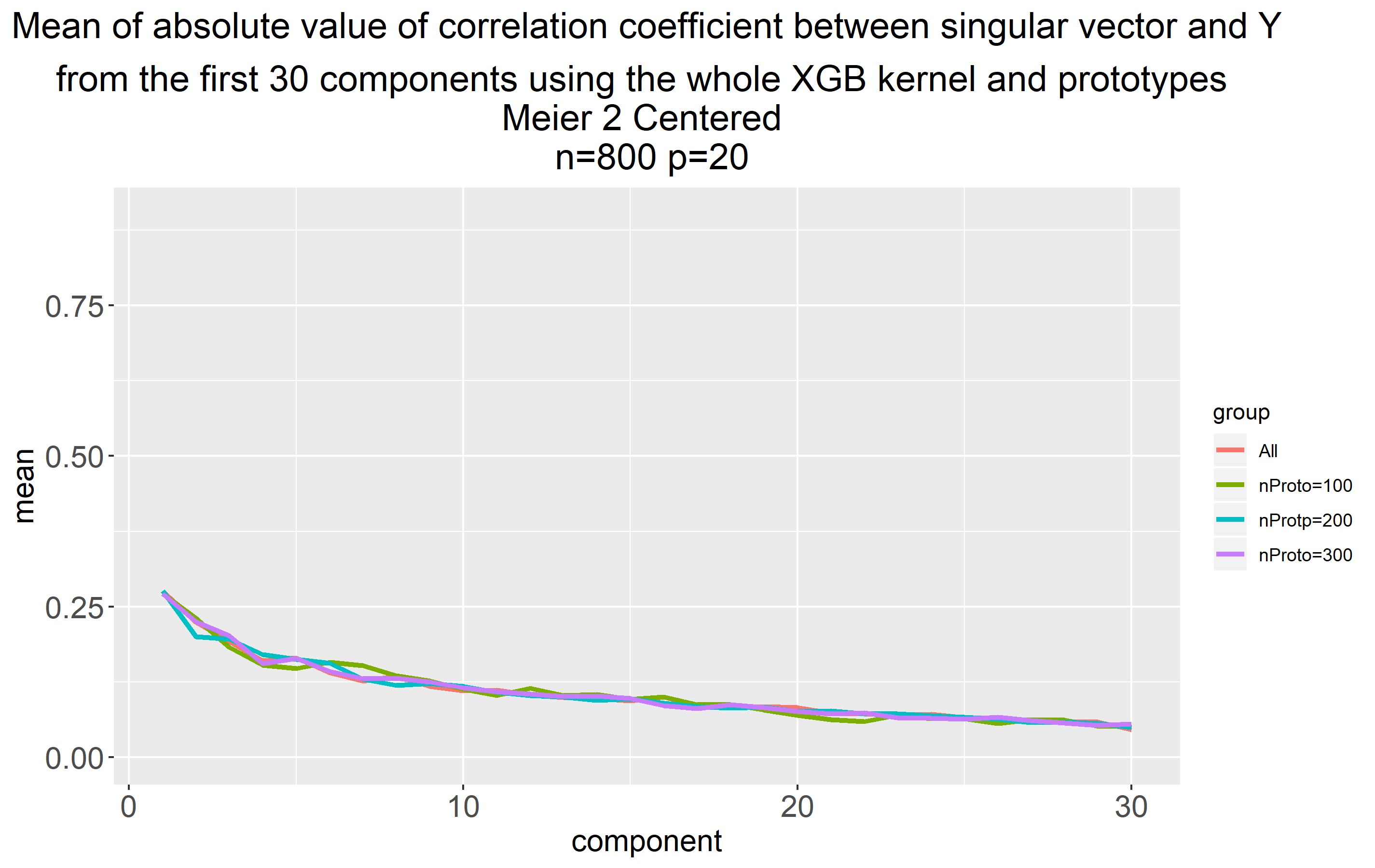}
      \subcaption{}
     \end{minipage}
     \\
     \begin{minipage}[b]{0.45\textwidth}
         \centering
         \includegraphics[width=\textwidth]{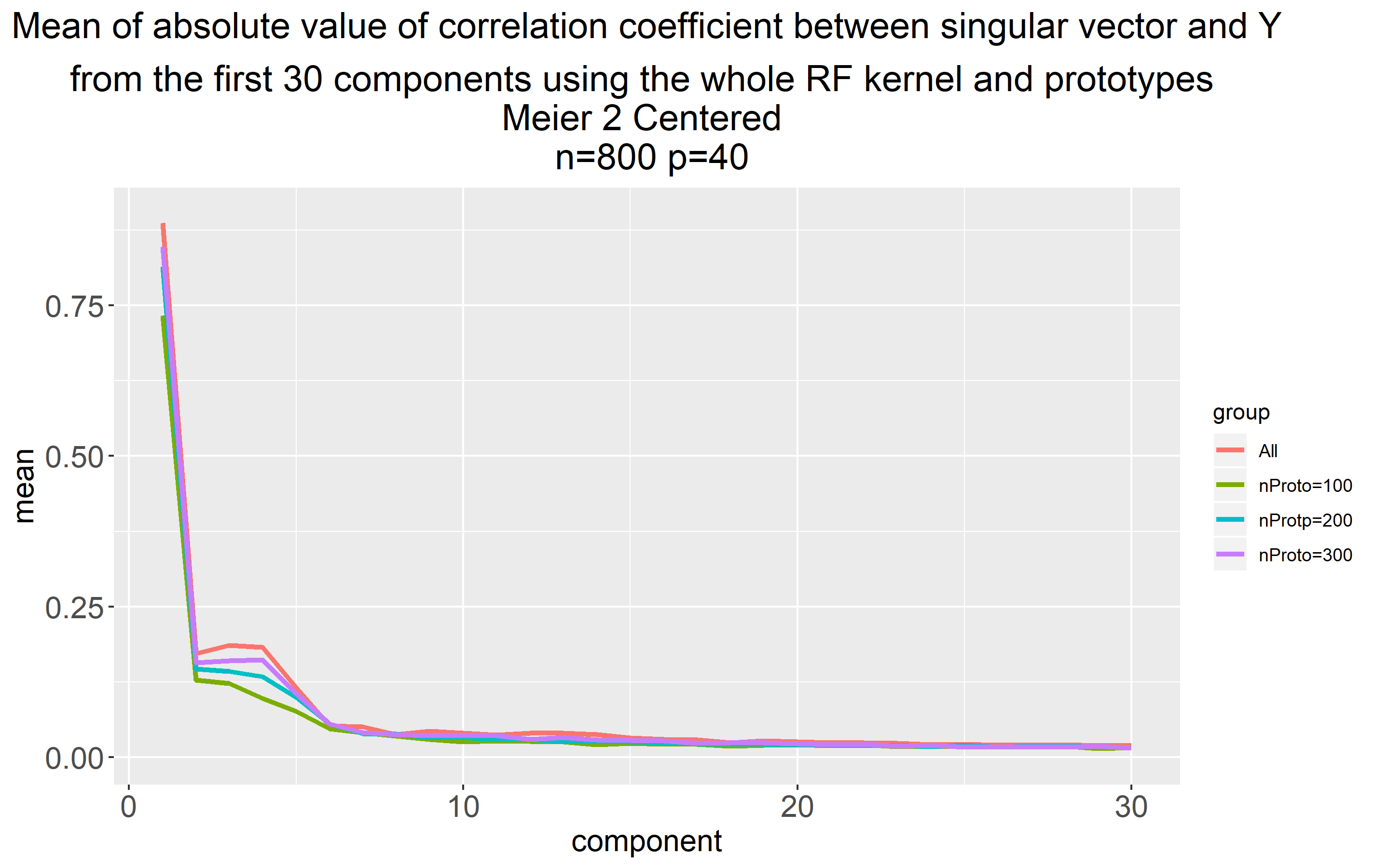}
           \subcaption{}
     \end{minipage}
     \hfill
     \begin{minipage}[b]{0.45\textwidth}
         \centering
         \includegraphics[width=\textwidth]{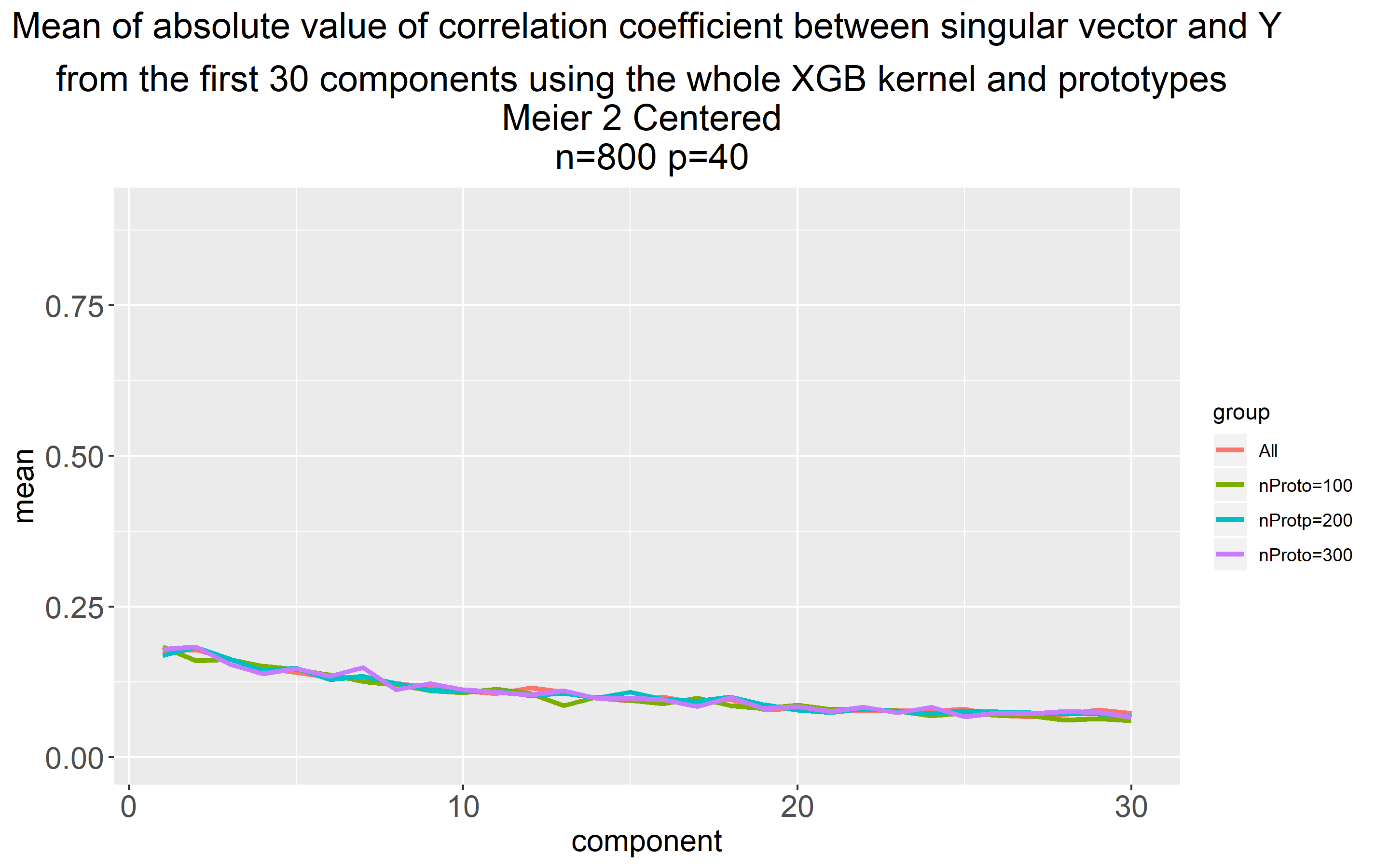}
            \subcaption{}
      \end{minipage}
           \begin{minipage}[b]{0.45\textwidth}
         \centering
         \includegraphics[width=\textwidth]{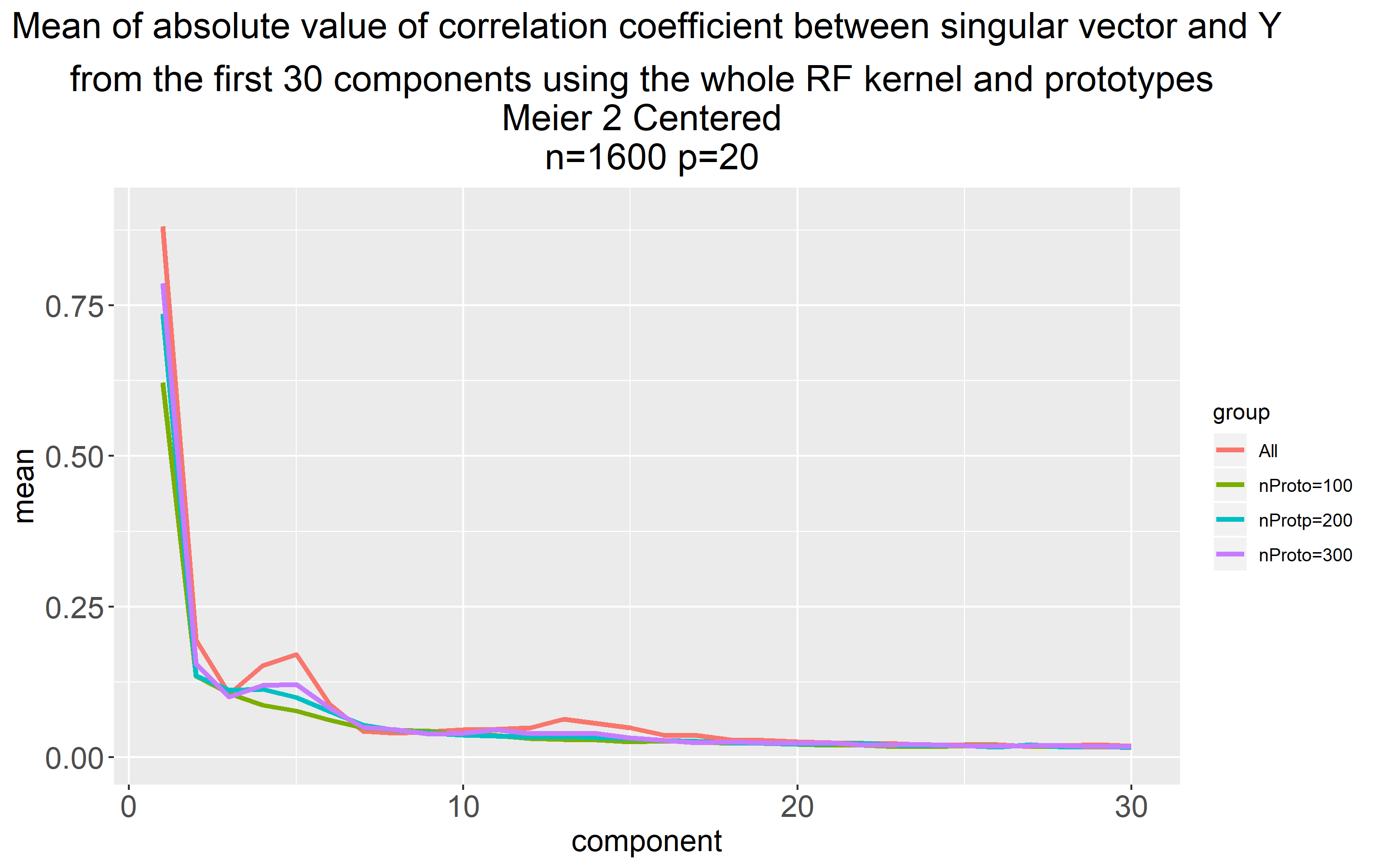}
    \subcaption{}
     \end{minipage}
     \hfill
     \begin{minipage}[b]{0.45\textwidth}
         \centering
         \includegraphics[width=\textwidth]{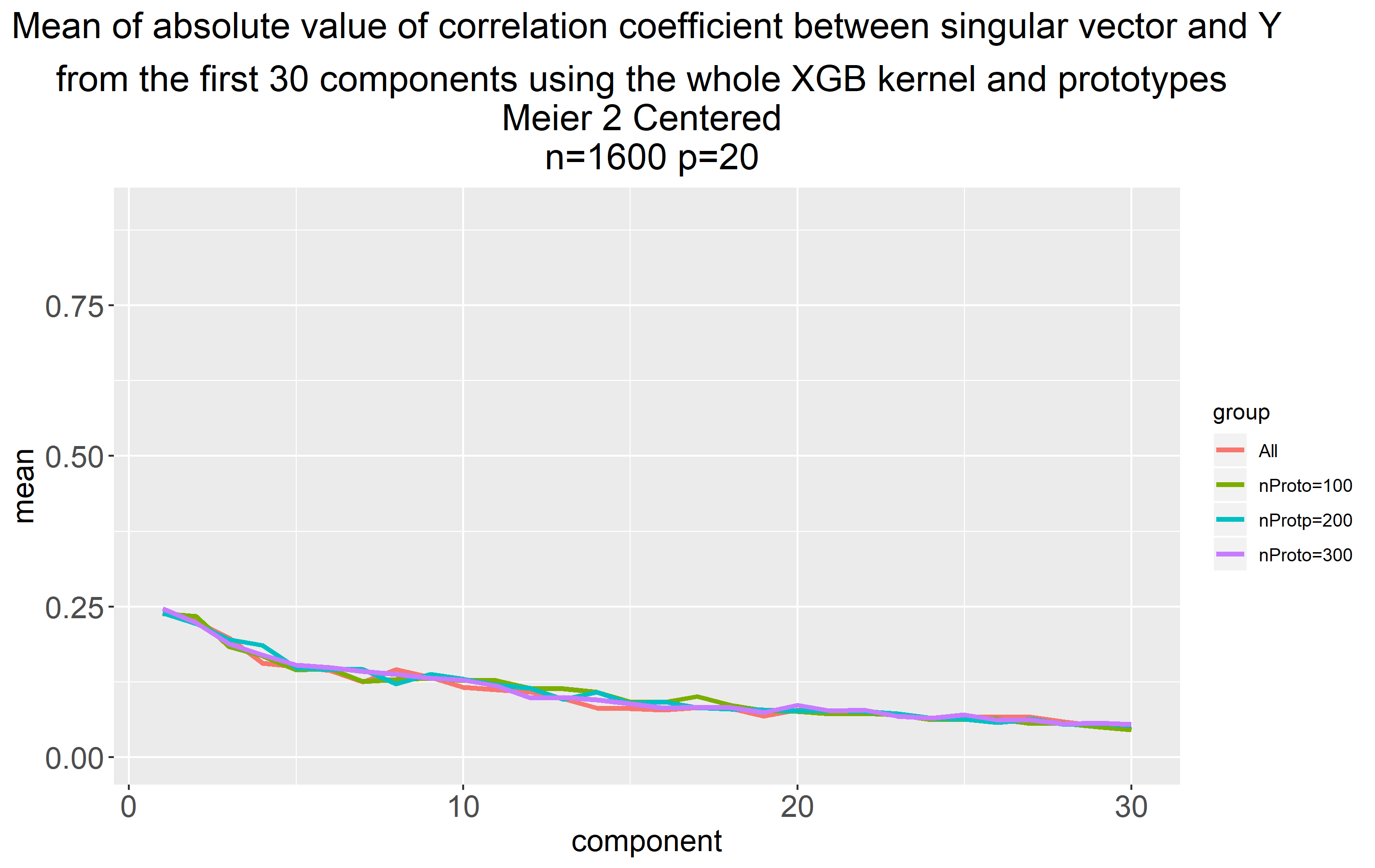}
     \subcaption{}
     \end{minipage}
     \\
     \begin{minipage}[b]{0.45\textwidth}
         \centering
         \includegraphics[width=\textwidth]{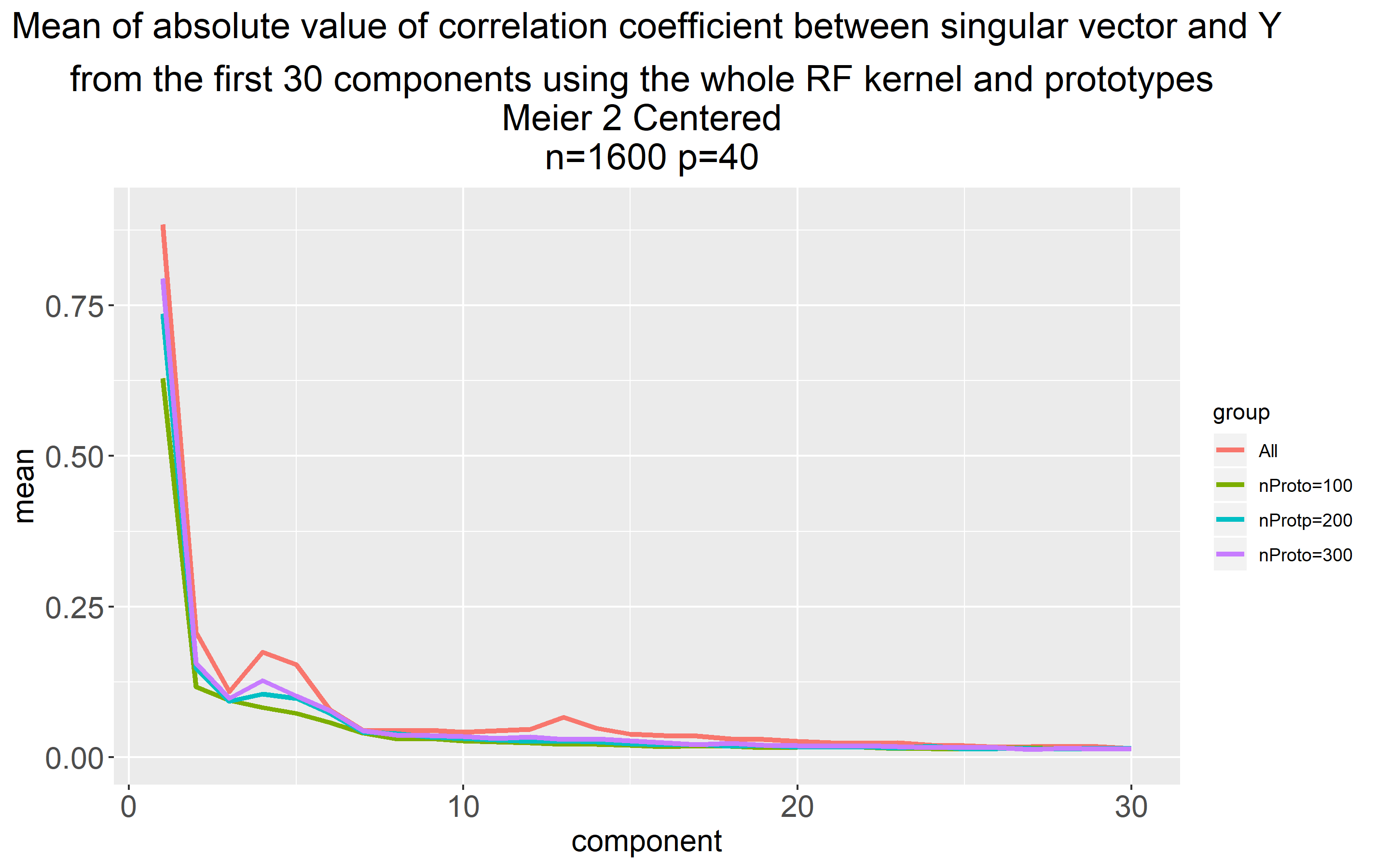}
         \subcaption{}
     \end{minipage}
     \hfill
     \begin{minipage}[b]{0.45\textwidth}
         \centering
         \includegraphics[width=\textwidth]{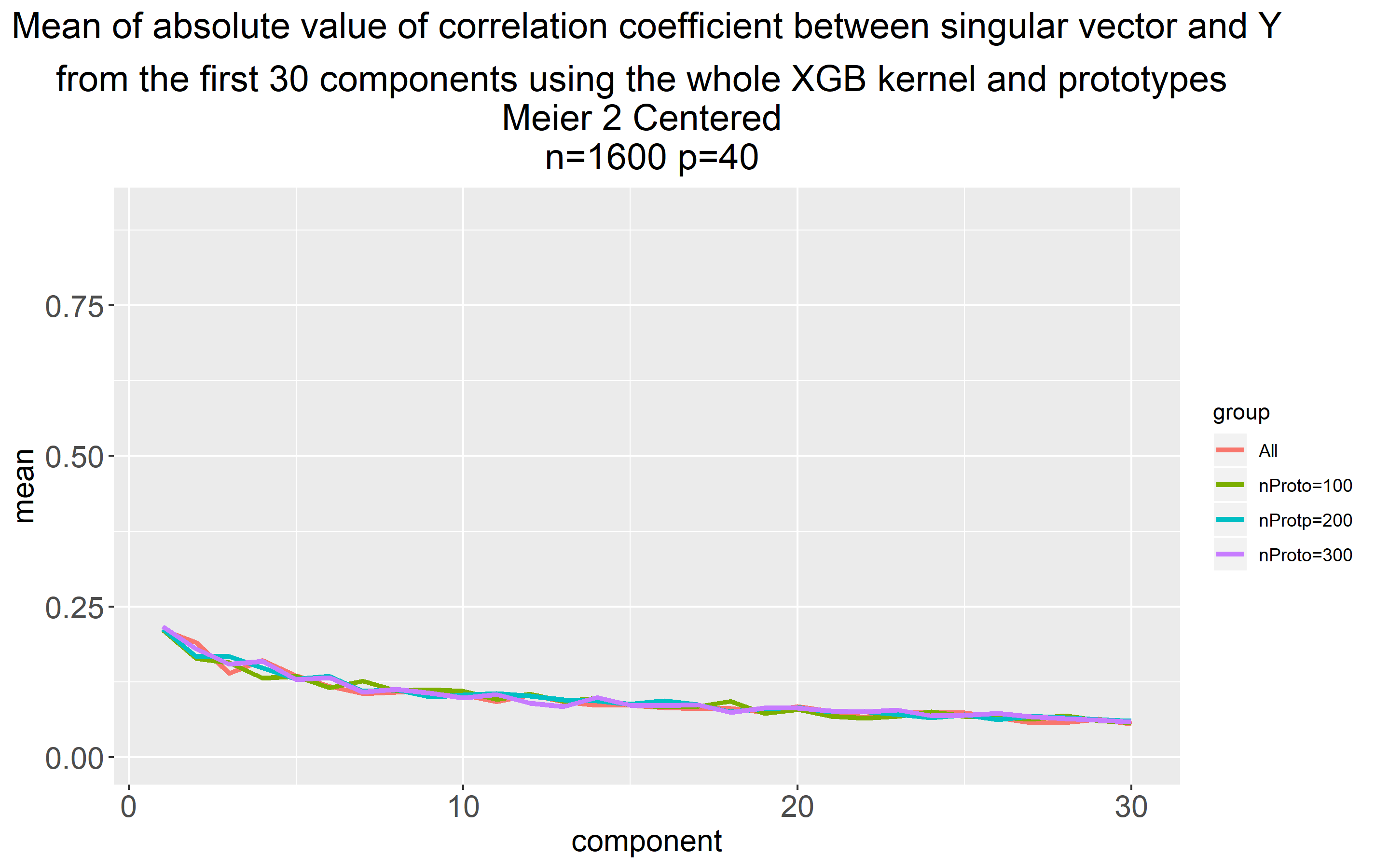}
        \subcaption{}
      \end{minipage}

        \caption{Kernel-target alignment spectra and relevant dimensionality---Meier 2}
        \label{fig:rd-meier2}
\end{figure}

\begin{figure}
     \centering
     \begin{minipage}[b]{0.45\textwidth}
         \centering
         \includegraphics[width=\textwidth]{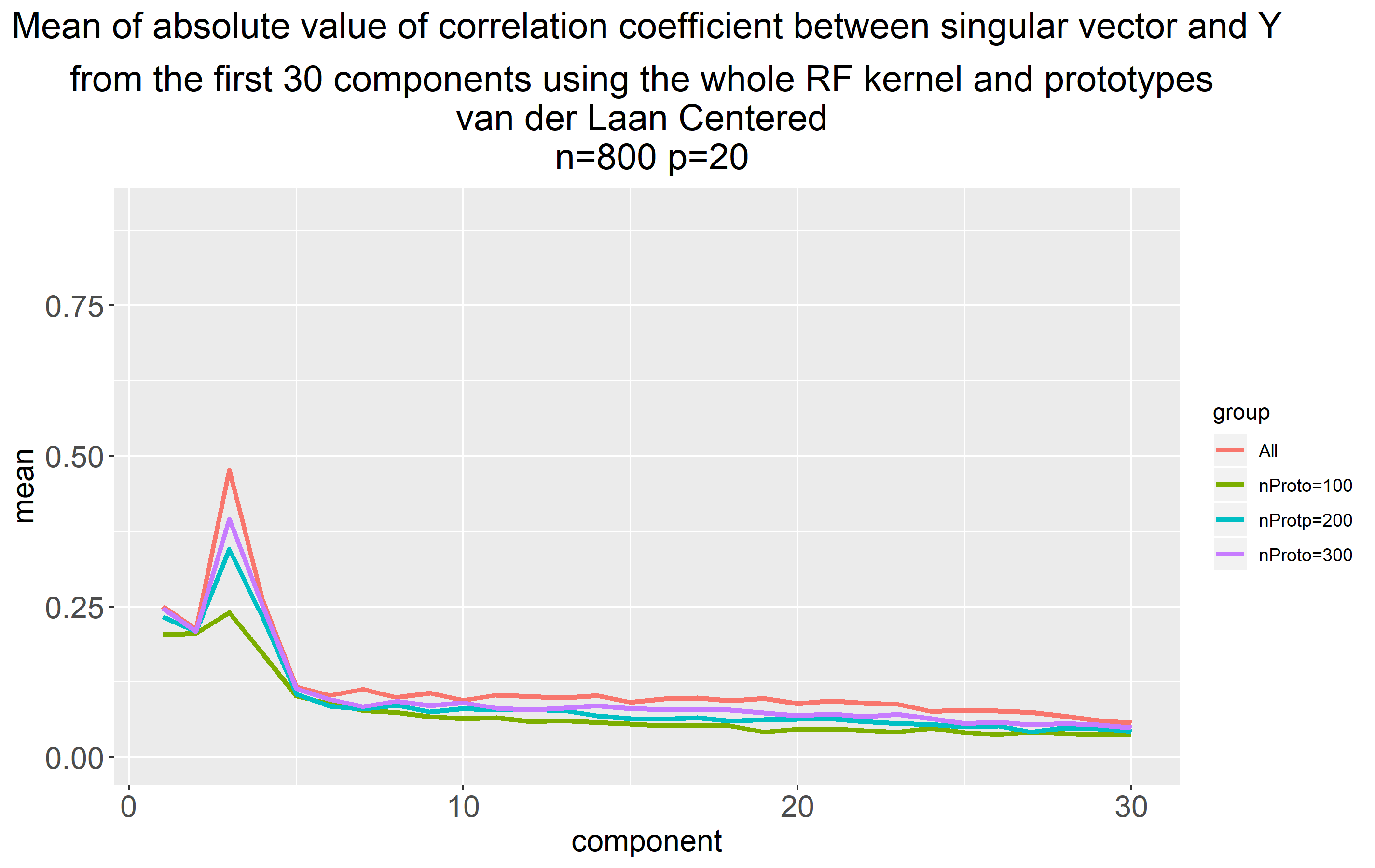}
       \subcaption{}
     \end{minipage}
     \hfill
     \begin{minipage}[b]{0.45\textwidth}
         \centering
         \includegraphics[width=\textwidth]{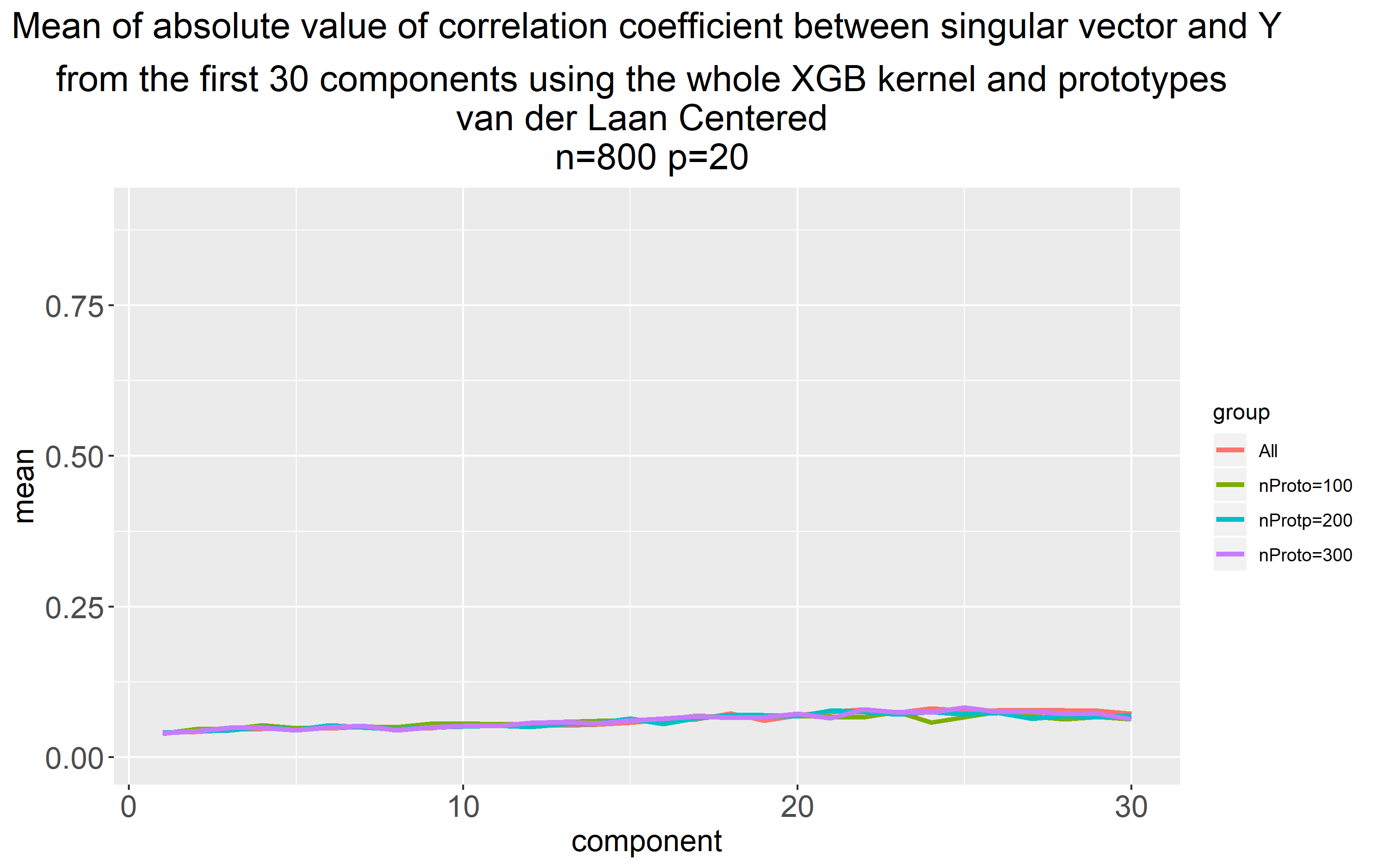}
           \subcaption{}
     \end{minipage}
     \\
     \begin{minipage}[b]{0.45\textwidth}
         \centering
         \includegraphics[width=\textwidth]{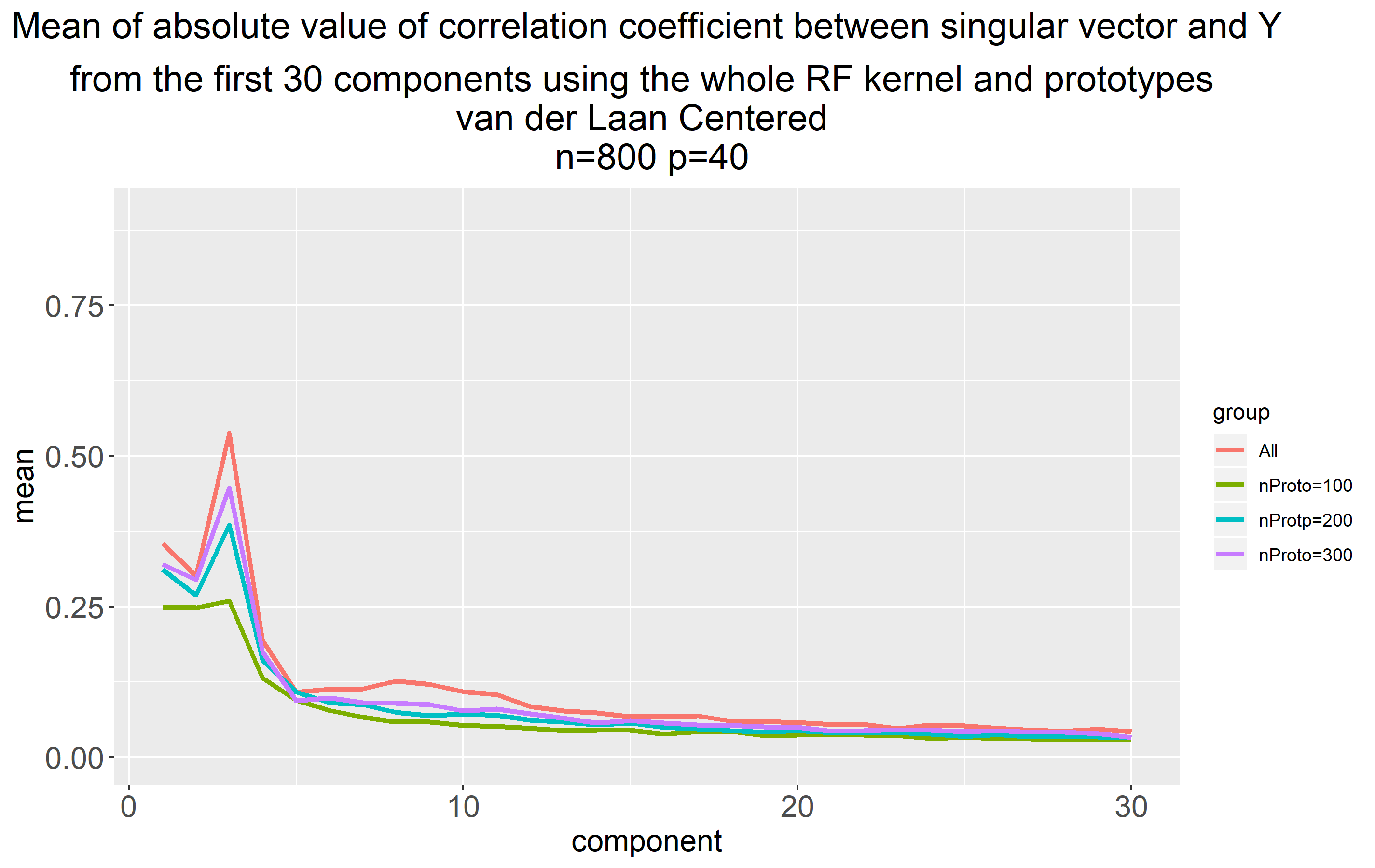}
            \subcaption{}
     \end{minipage}
     \hfill
     \begin{minipage}[b]{0.45\textwidth}
         \centering
         \includegraphics[width=\textwidth]{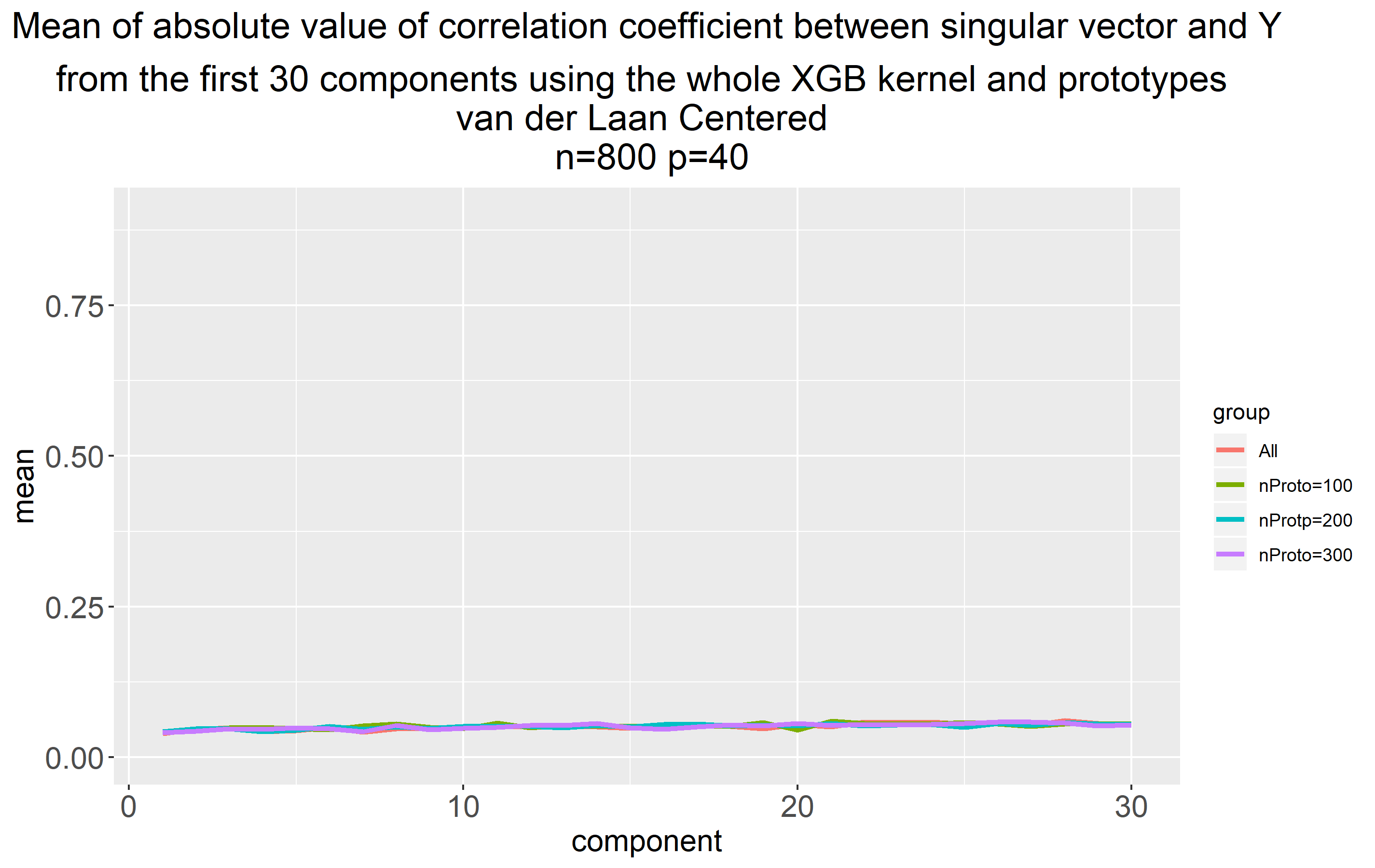}
       \subcaption{}
      \end{minipage}
           \begin{minipage}[b]{0.45\textwidth}
         \centering
         \includegraphics[width=\textwidth]{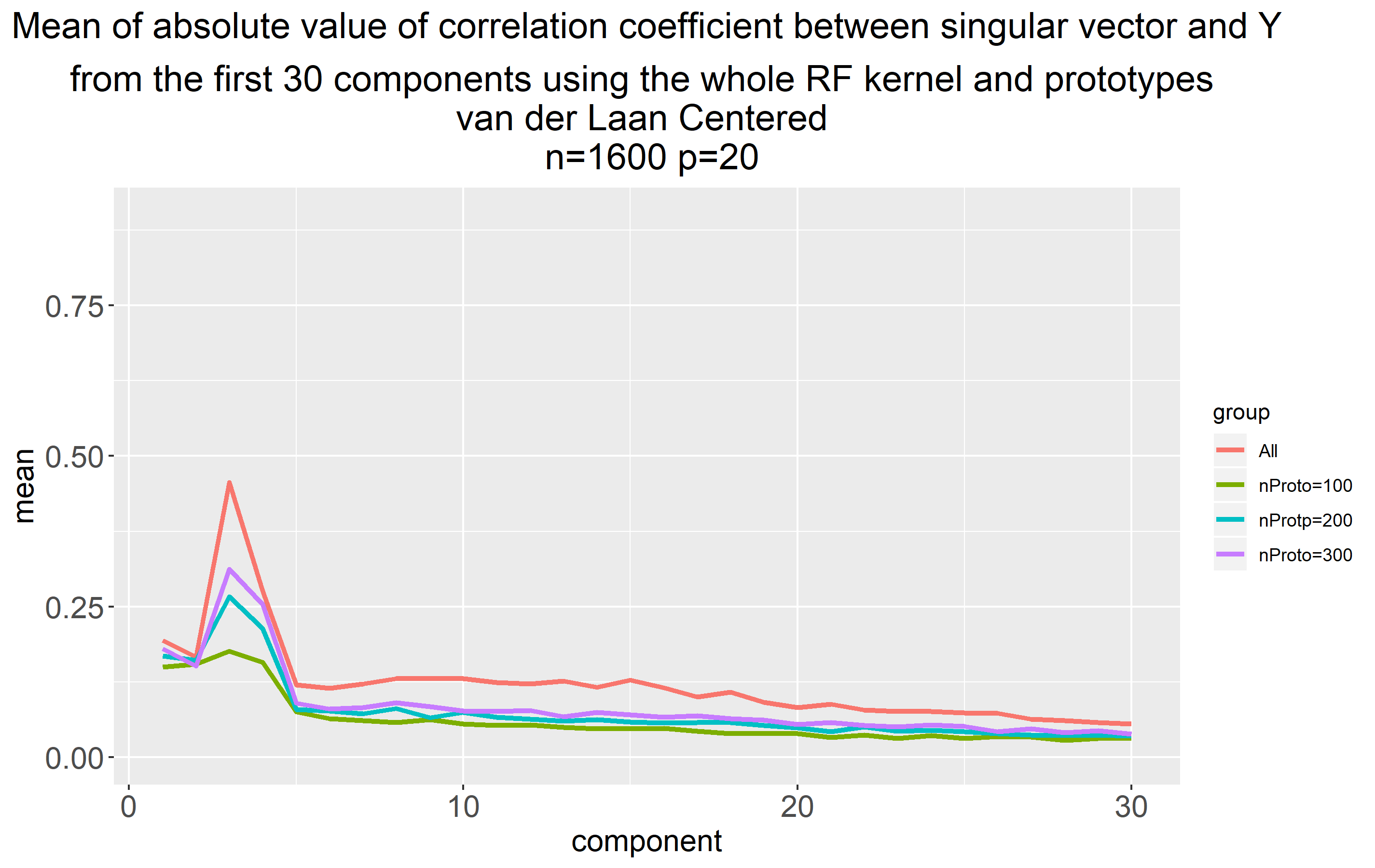}
          \subcaption{}
     \end{minipage}
     \hfill
     \begin{minipage}[b]{0.45\textwidth}
         \centering
         \includegraphics[width=\textwidth]{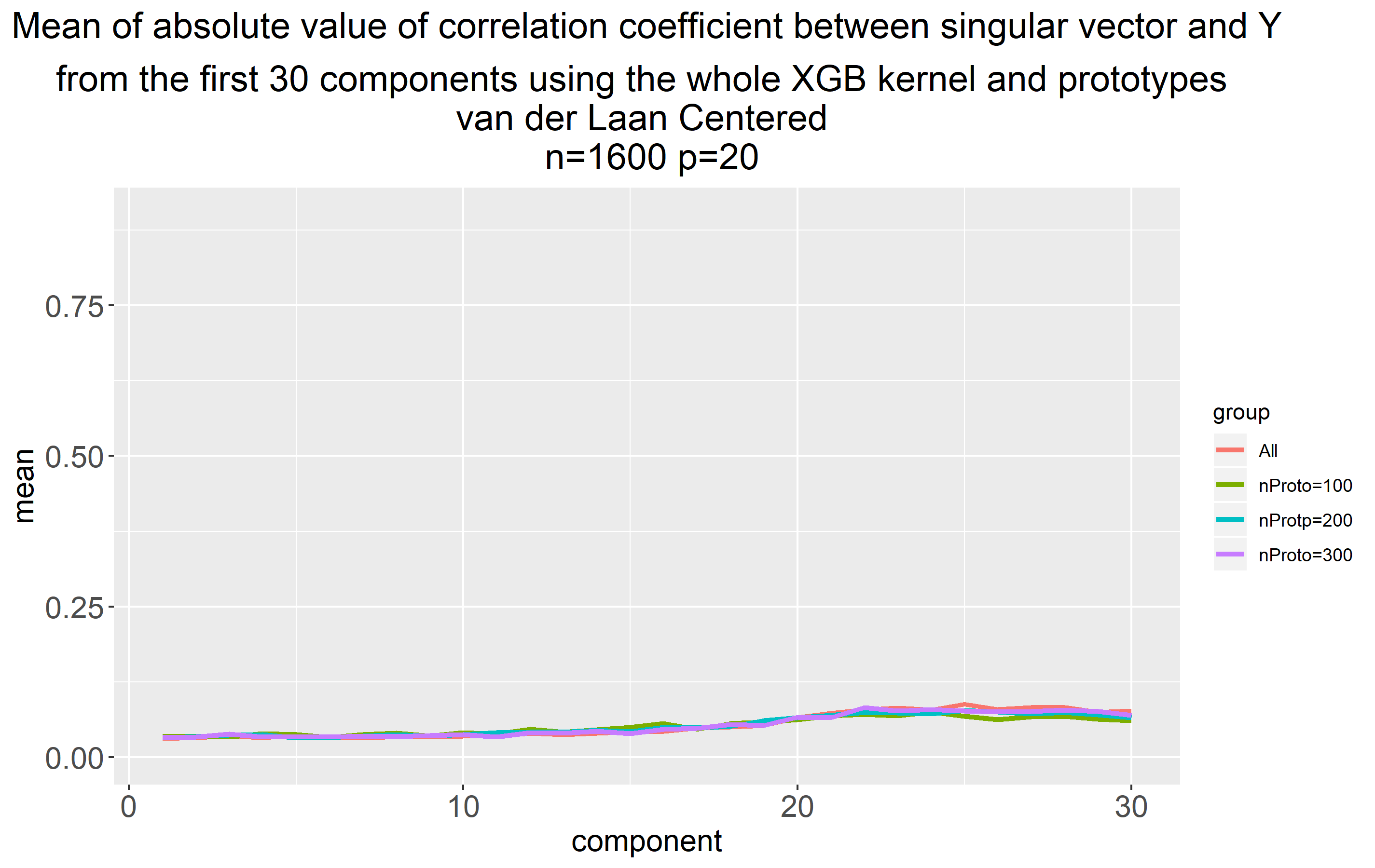}
       \subcaption{}
     \end{minipage}
     \\
     \begin{minipage}[b]{0.45\textwidth}
         \centering
         \includegraphics[width=\textwidth]{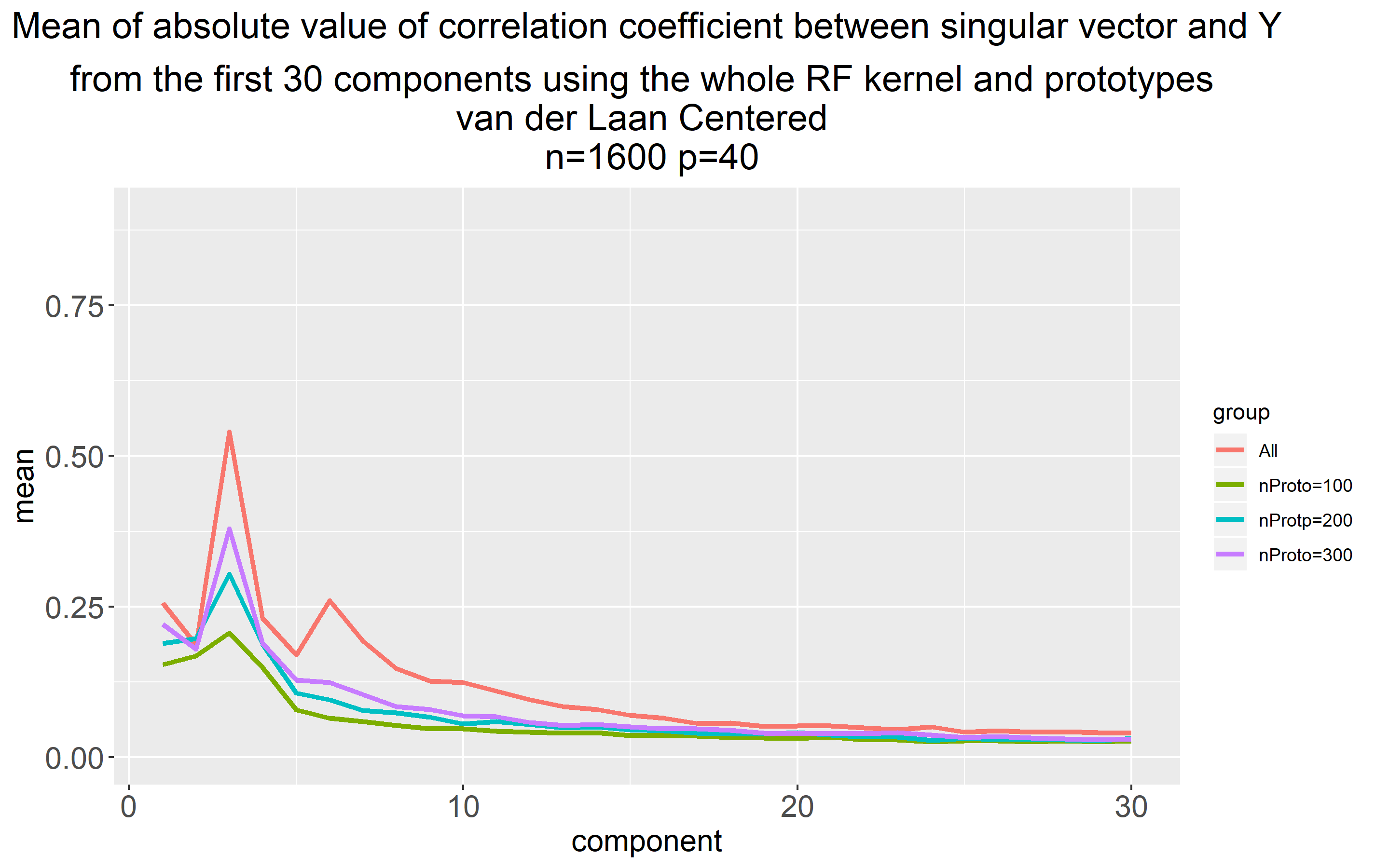}
          \subcaption{}
     \end{minipage}
     \hfill
     \begin{minipage}[b]{0.45\textwidth}
         \centering
         \includegraphics[width=\textwidth]{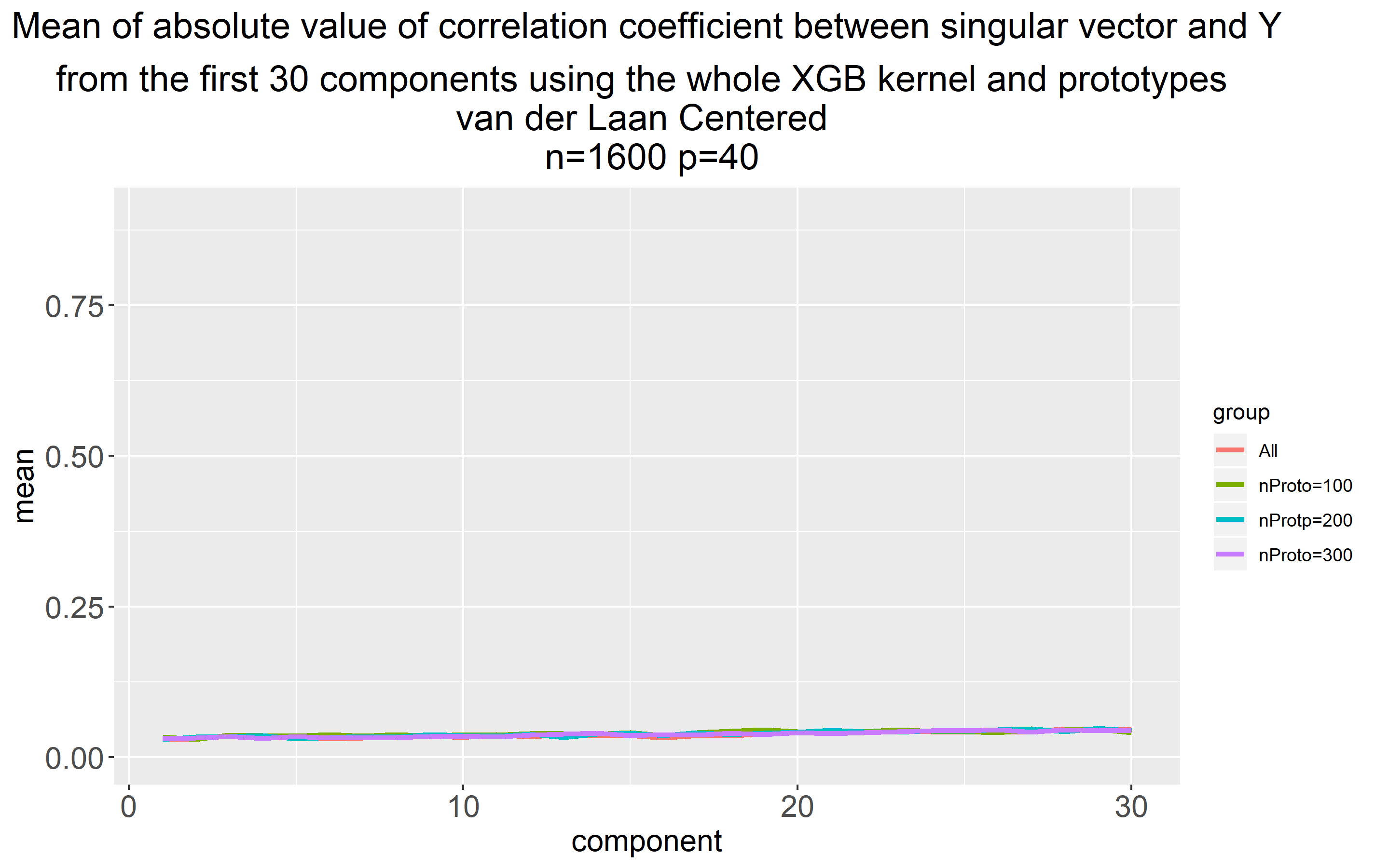}
        \subcaption{}
      \end{minipage}

        \caption{Kernel-target alignment spectra and relevant dimensionality---van der Laan}
        \label{fig:rd-vanderlaan}
\end{figure}

The kernel-target alignment spectra of the Friedman, Checkerboard, Meier 1, Meier 2 and van der Laan simulation settings are shown in the Figs. \ref{fig:rd-friedman}, \ref{fig:rd-checkerboard}, \ref{fig:rd-meier1}, \ref{fig:rd-meier2} and \ref{fig:rd-vanderlaan}, respectively. The left and right panels in the Figures correspond to the results obtained from the RF and XGB, respectively. On the x-axis, the first thirty components ordered according to the decreasing eigenvalues of the kernel matrix (singular values) are shown. These components are shown for the ensemble derived kernel matrix $K_{ensemble}$ and the landmark matrices $L$-s. The landmark matrices $L$ are built from a varying number of landmarks or prototypes (nProto) including 100, 200 and 300 prototypes. The y-axis corresponds to the absolute value of the correlation coefficient between corresponding eigenvectors of $\mbold{K_{ensemble}}$ (or left singular vectors of $L$) and the target $Y$. 

The RF alignment spectrum for Friedman shows strong peaks corresponding to leading eigenvectors (singular vectors) across the simulation settings Figs. \ref{fig:rd-friedman}(a,c,e,g). These peaks show the same pattern for the prototypes and their magnitude is monotonically increasing with increasing number of prototypes (n=100, 200 and 300). 
The XGB alignment spectra for Friedman are flatter than those for the RF Figs. \ref{fig:rd-friedman}(b,d,f,h) and they are overlapping with respect to the increasing number of prototypes.

The RF alignment spectrum for Checkerboard is similar to that of Friedman across all simulation settings. A strong peak is associated with the second leading eigenvector (left singular vector) Figs. \ref{fig:rd-checkerboard}(a,c,e,g). In contrary to Friedman, the XGB alignment spectrum also shows strong components across all simulation settings Figs. \ref{fig:rd-checkerboard}(b,d,f,h). Interestingly, for the Checkerboard, there is little difference in the alignment spectra with respect to the number of prototypes.

The Meier 1 and Meier 2 data sets are similar to the Friedman data set. For the RF alignment spectrum, Meier 1 and Meier 2 show strong peaks, magnitude of which is increases with the increasing number of prototypes (Figs.\ref{fig:rd-meier1}(a,c,e,g) and Figs. \ref{fig:rd-meier2}(a,c,e,g) for Meier 1 and Meier 2, respectively). The XGB kernel alignment spectra are flatter, monotonically decreasing and overlapping with respect to the number of prototypes (Figs.\ref{fig:rd-meier1}(b,d,f,h) and \ref{fig:rd-meier2}(b,d,f,h) for Meier 1 and Meier 2, respectively).

For the van der Laan data set, the RF kernel alignment spectrum shows stronger peaks for leading eigenvectors (singular vectors) as it was for the other simulated data sets (Figs.\ref{fig:rd-vanderlaan}(a,c,e,g)). However, the alignment spectrum for van der Laan is flat across all scenarios, indicating weak performance of XGB for this data set.

\begin{figure}
     \centering
     \begin{minipage}[b]{0.30\textwidth}
         \centering
         \includegraphics[width=\textwidth]{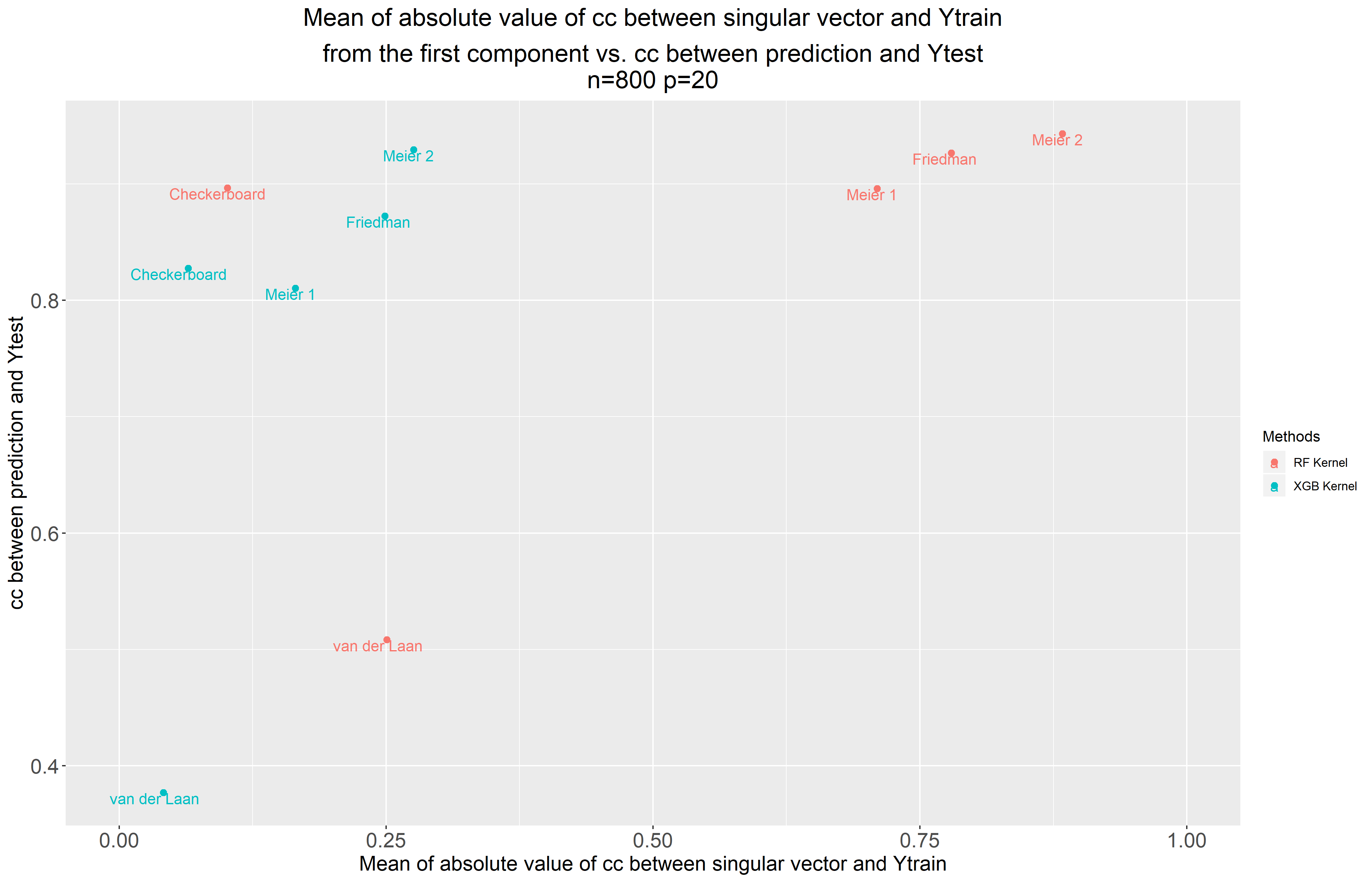}
          \subcaption{}
     \end{minipage}
     \hfill
     \begin{minipage}[b]{0.30\textwidth}
         \centering
         \includegraphics[width=\textwidth]{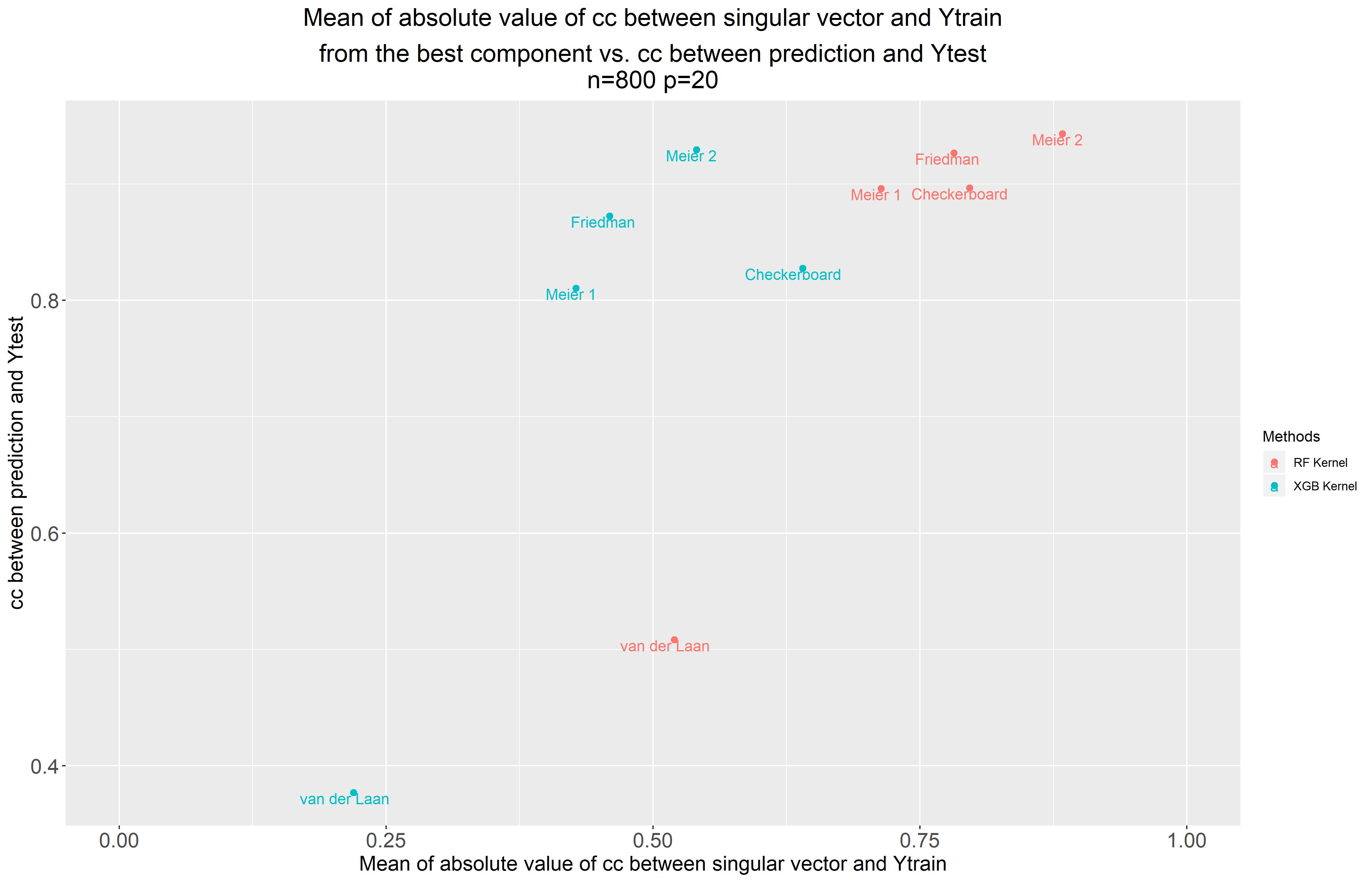}
          \subcaption{}
     \end{minipage}
     \hfill
     \begin{minipage}[b]{0.30\textwidth}
         \centering
         \includegraphics[width=\textwidth]{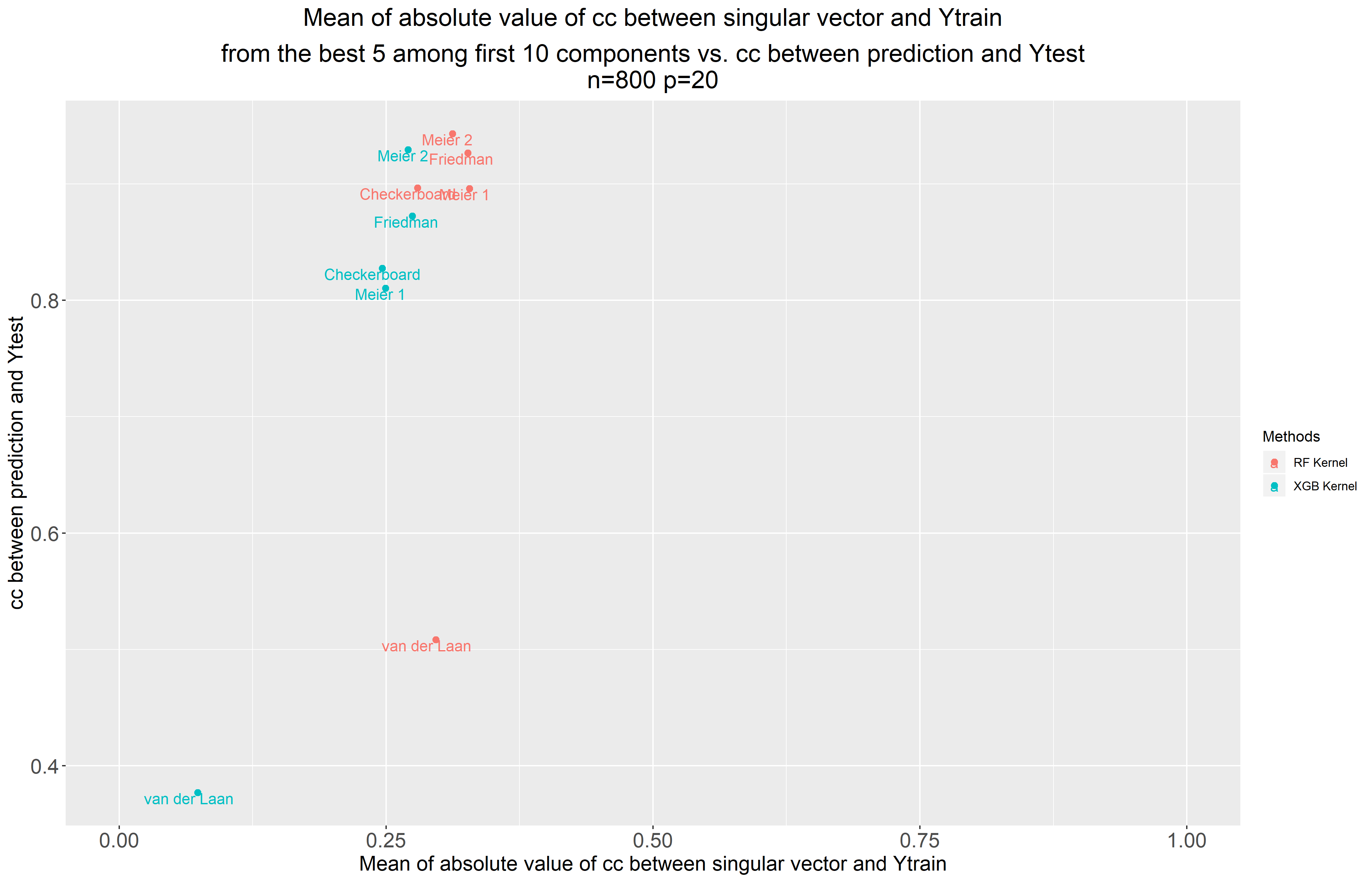}
        \subcaption{}
     \end{minipage}
     \\
     \begin{minipage}[b]{0.3\textwidth}
         \centering
         \includegraphics[width=\textwidth]{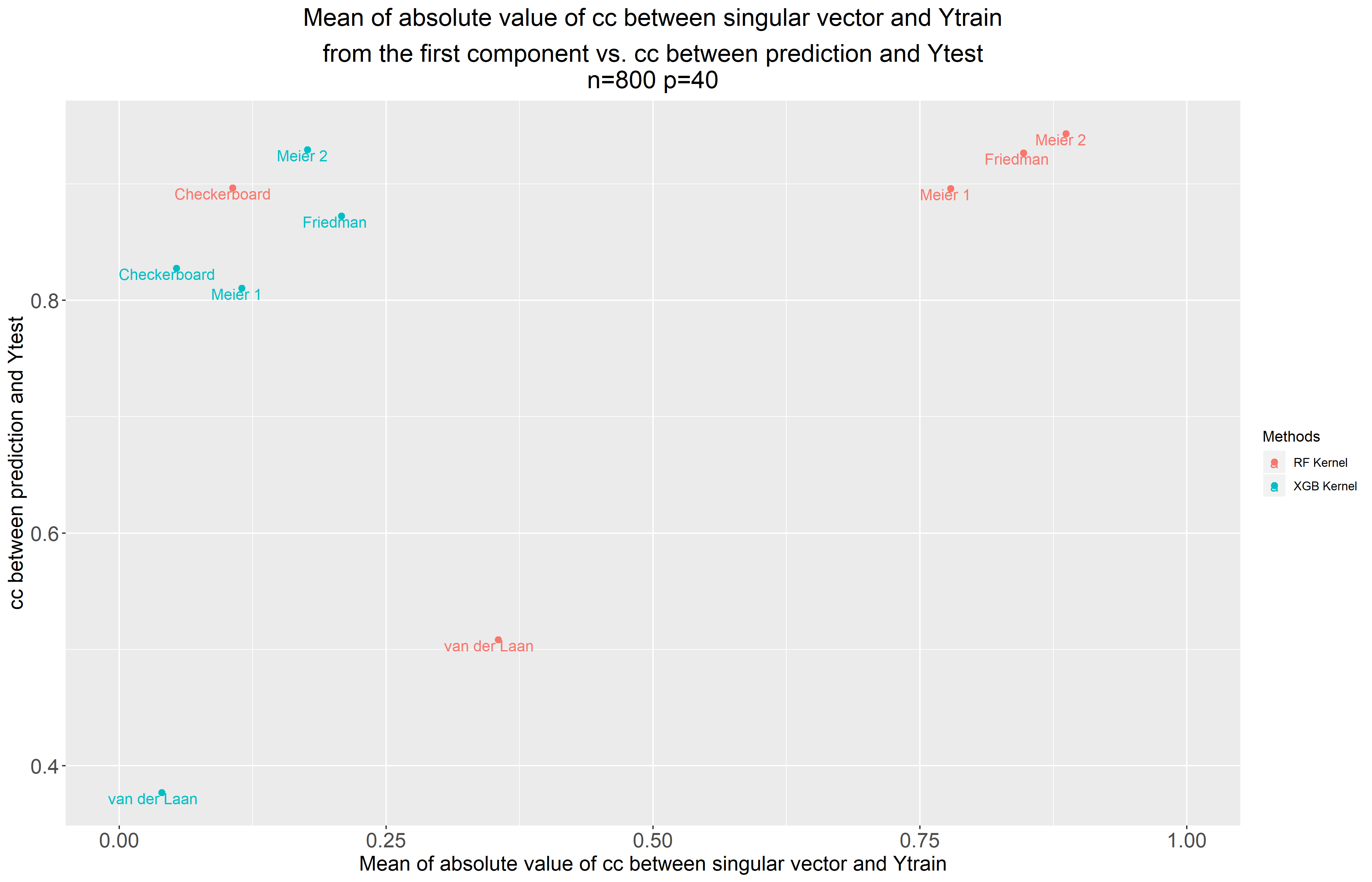}
          \subcaption{}
     \end{minipage}
     \hfill
       \begin{minipage}[b]{0.3\textwidth}
         \centering
         \includegraphics[width=\textwidth]{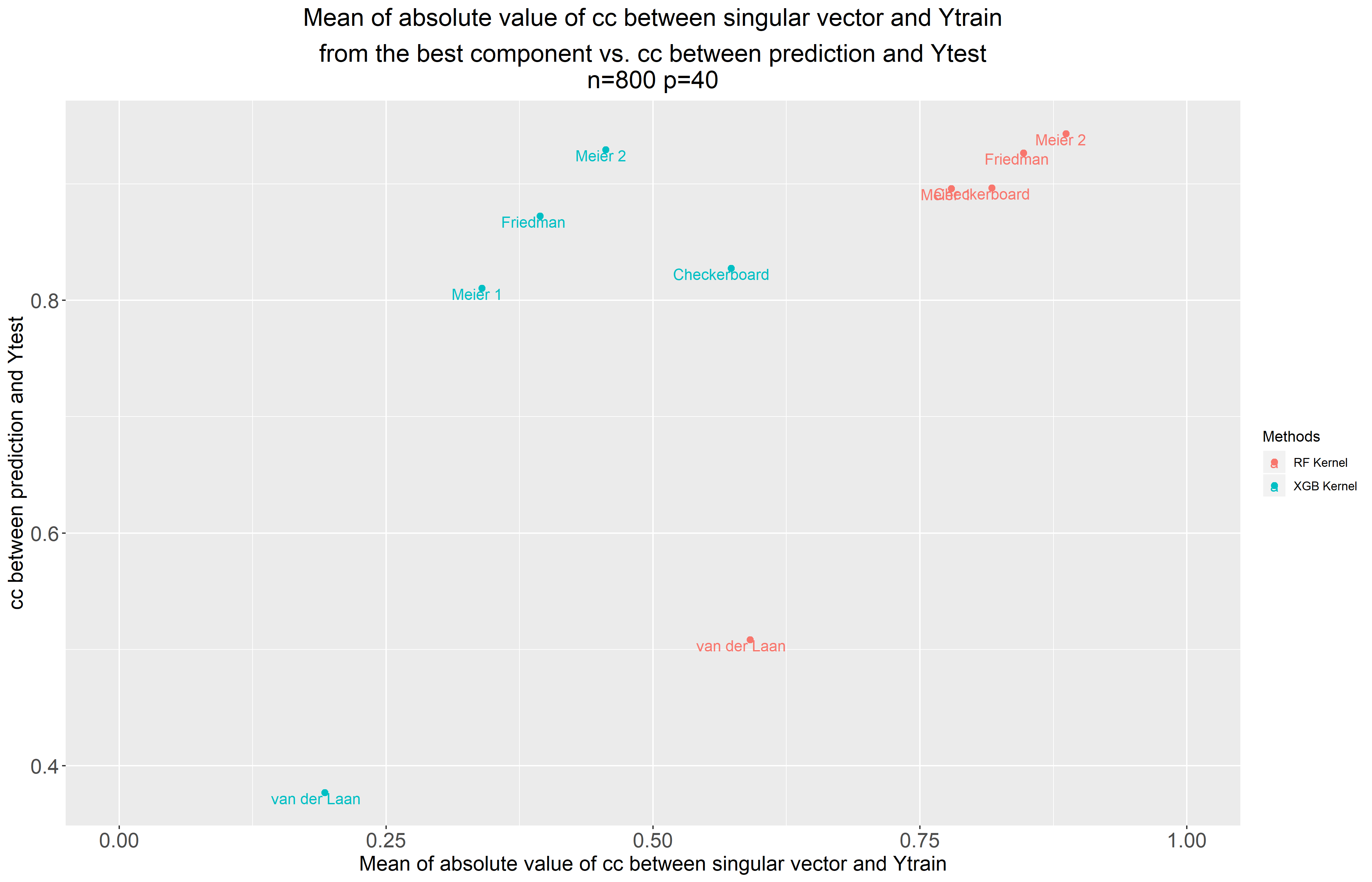}
          \subcaption{}
     \end{minipage}
     \hfill
        \begin{minipage}[b]{0.3\textwidth}
         \centering
         \includegraphics[width=\textwidth]{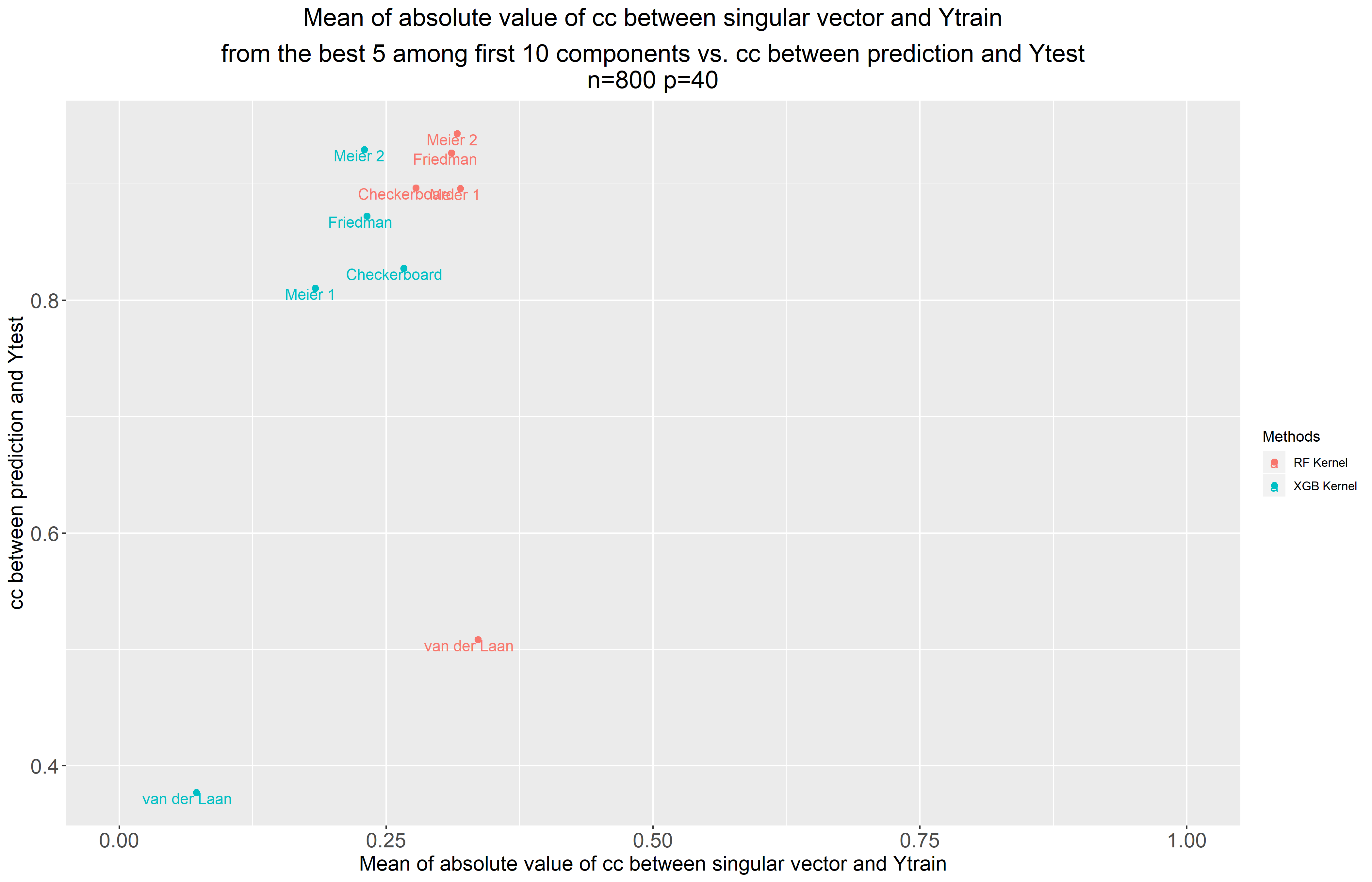}
        \subcaption{}
     \end{minipage}
     \\
    \begin{minipage}[b]{0.3\textwidth}
         \centering
         \includegraphics[width=\textwidth]{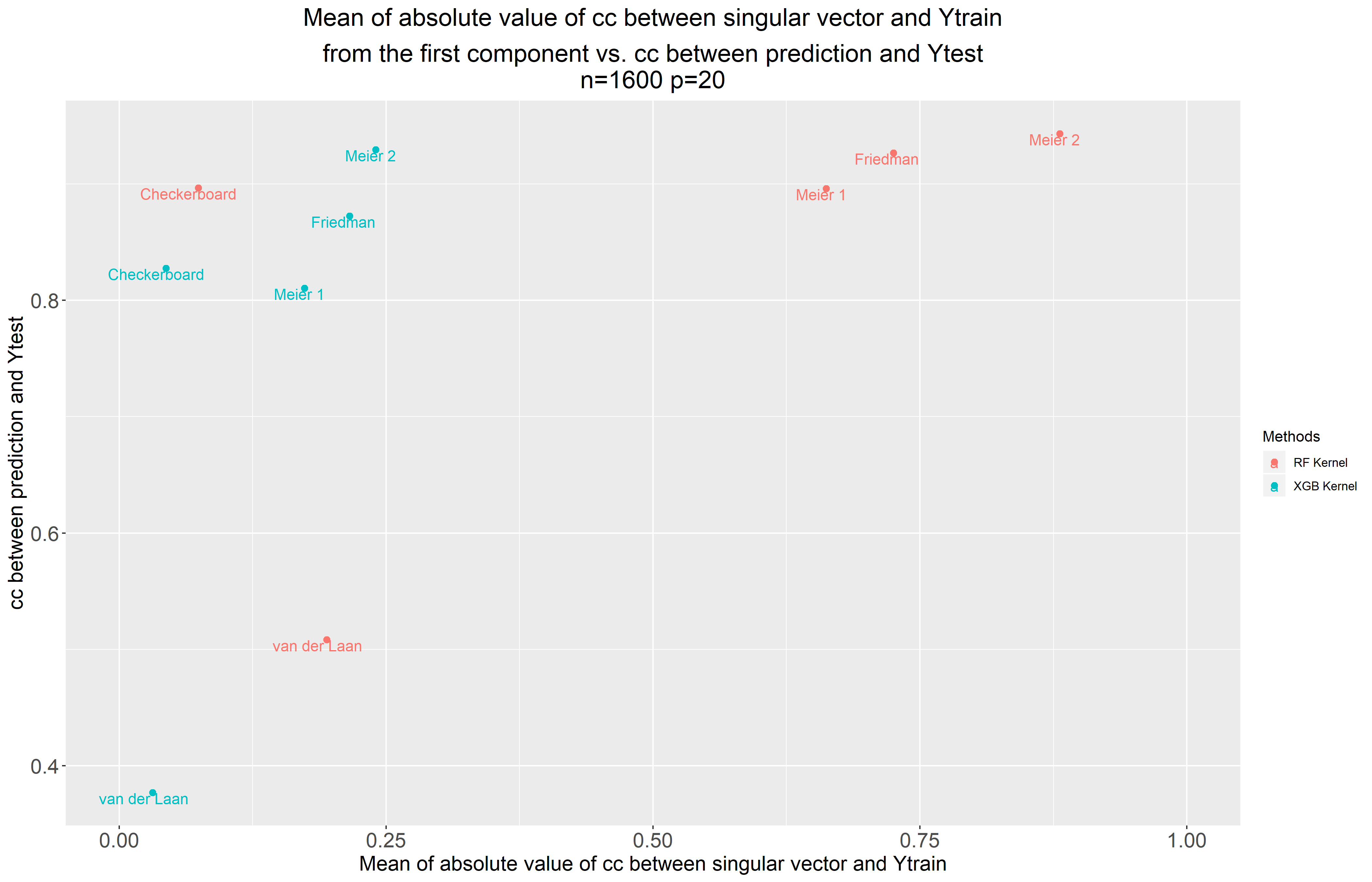}
          \subcaption{}
     \end{minipage}
     \hfill
      \begin{minipage}[b]{0.3\textwidth}
         \centering
         \includegraphics[width=\textwidth]{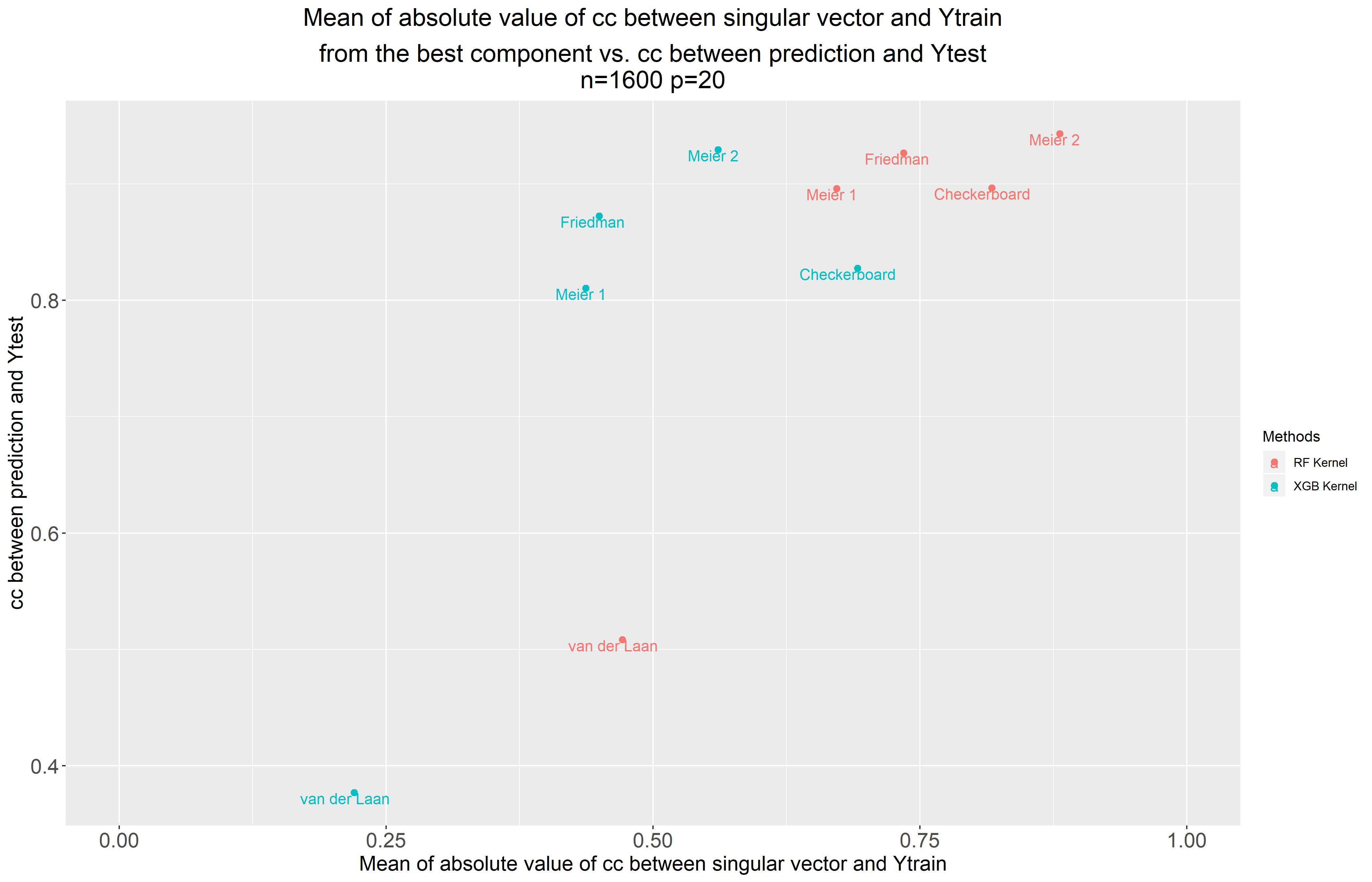}
          \subcaption{}
     \end{minipage}
     \hfill
     \begin{minipage}[b]{0.3\textwidth}
         \centering
         \includegraphics[width=\textwidth]{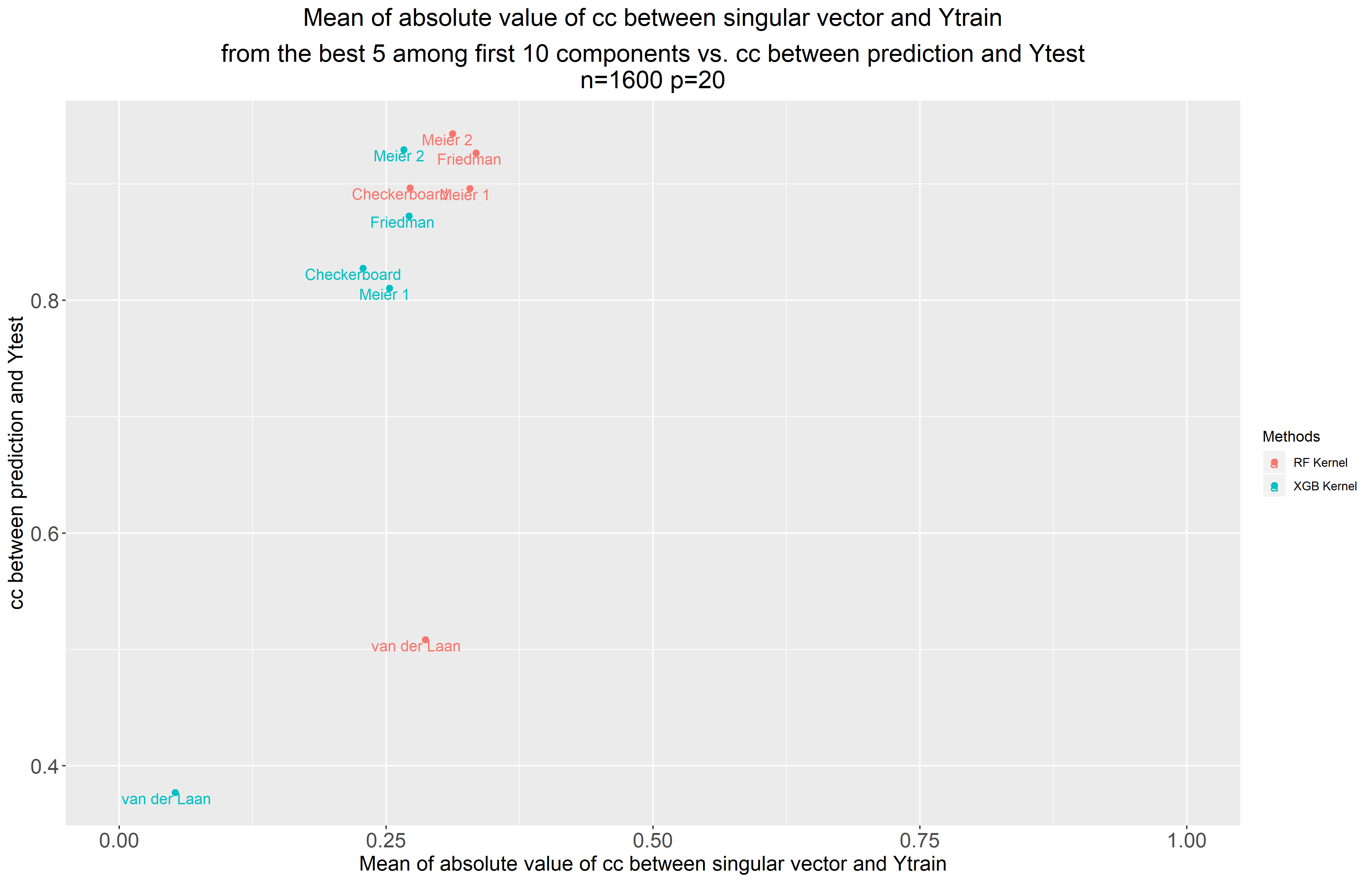}
        \subcaption{}
     \end{minipage}
     \\
     \begin{minipage}[b]{0.3\textwidth}
         \centering
         \includegraphics[width=\textwidth]{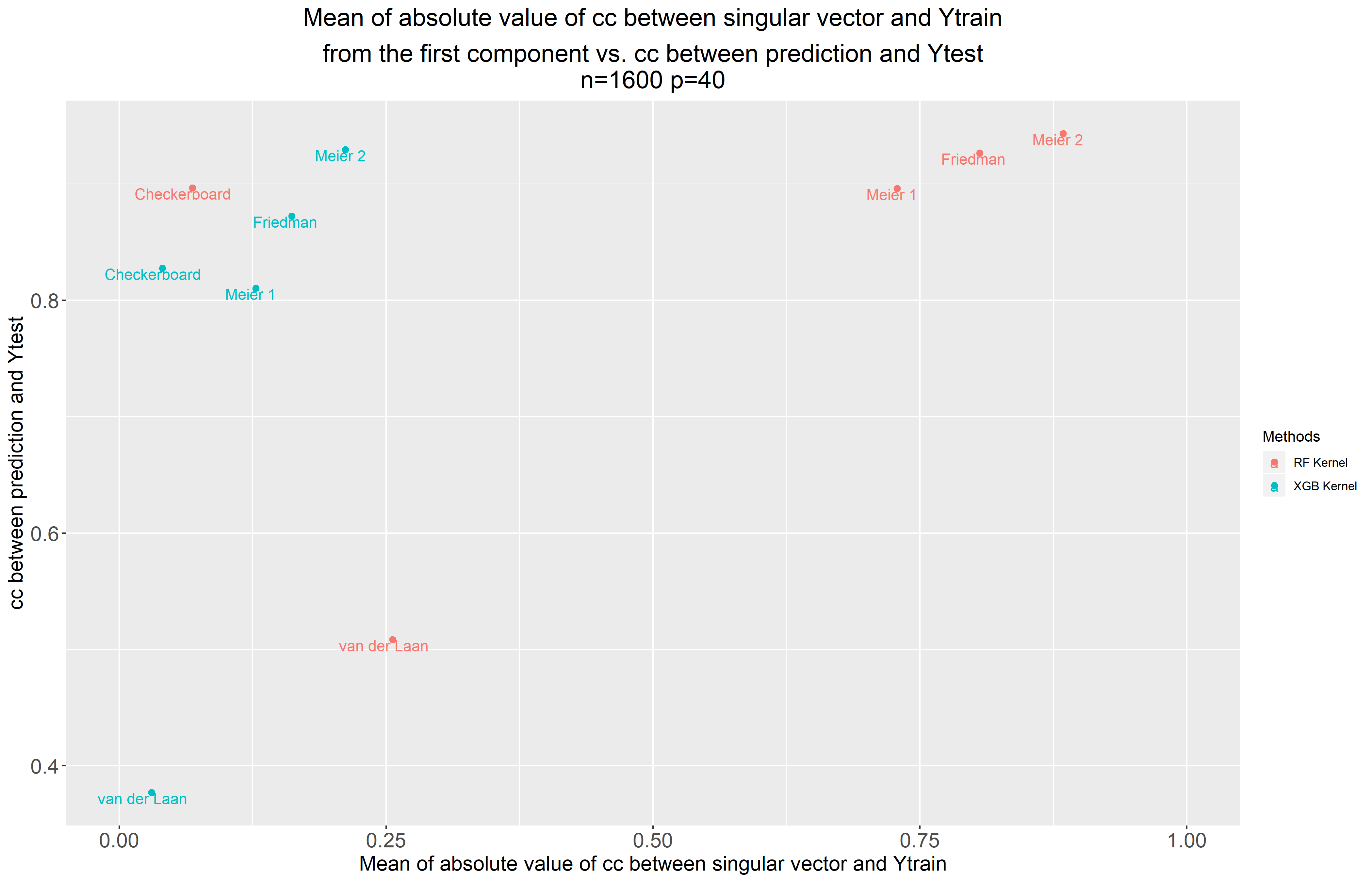}
          \subcaption{}
     \end{minipage}
     \hfill
       \begin{minipage}[b]{0.3\textwidth}
         \centering
         \includegraphics[width=\textwidth]{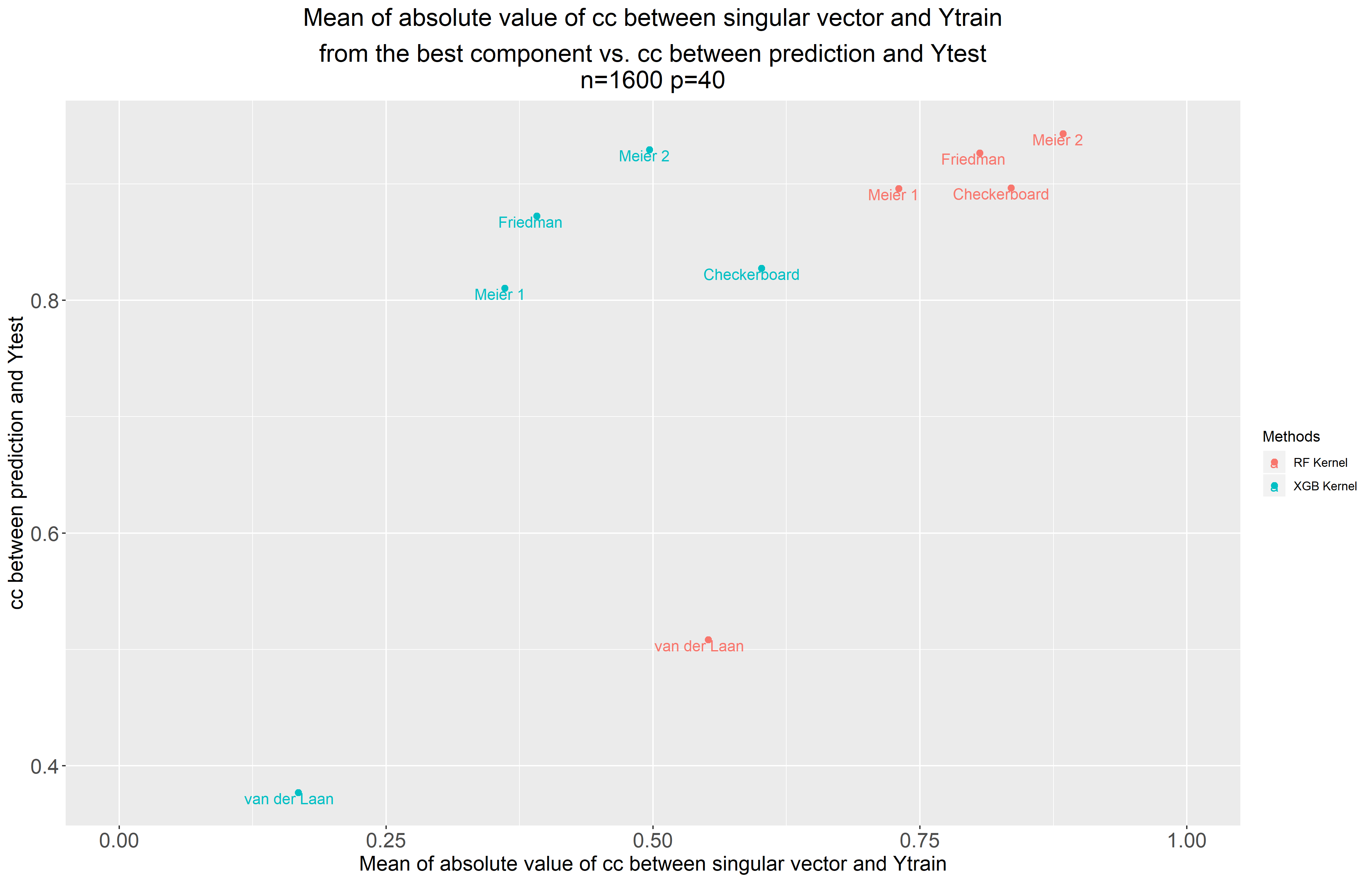}
          \subcaption{}
     \end{minipage}
     \hfill
    \begin{minipage}[b]{0.3\textwidth}
         \centering
         \includegraphics[width=\textwidth]{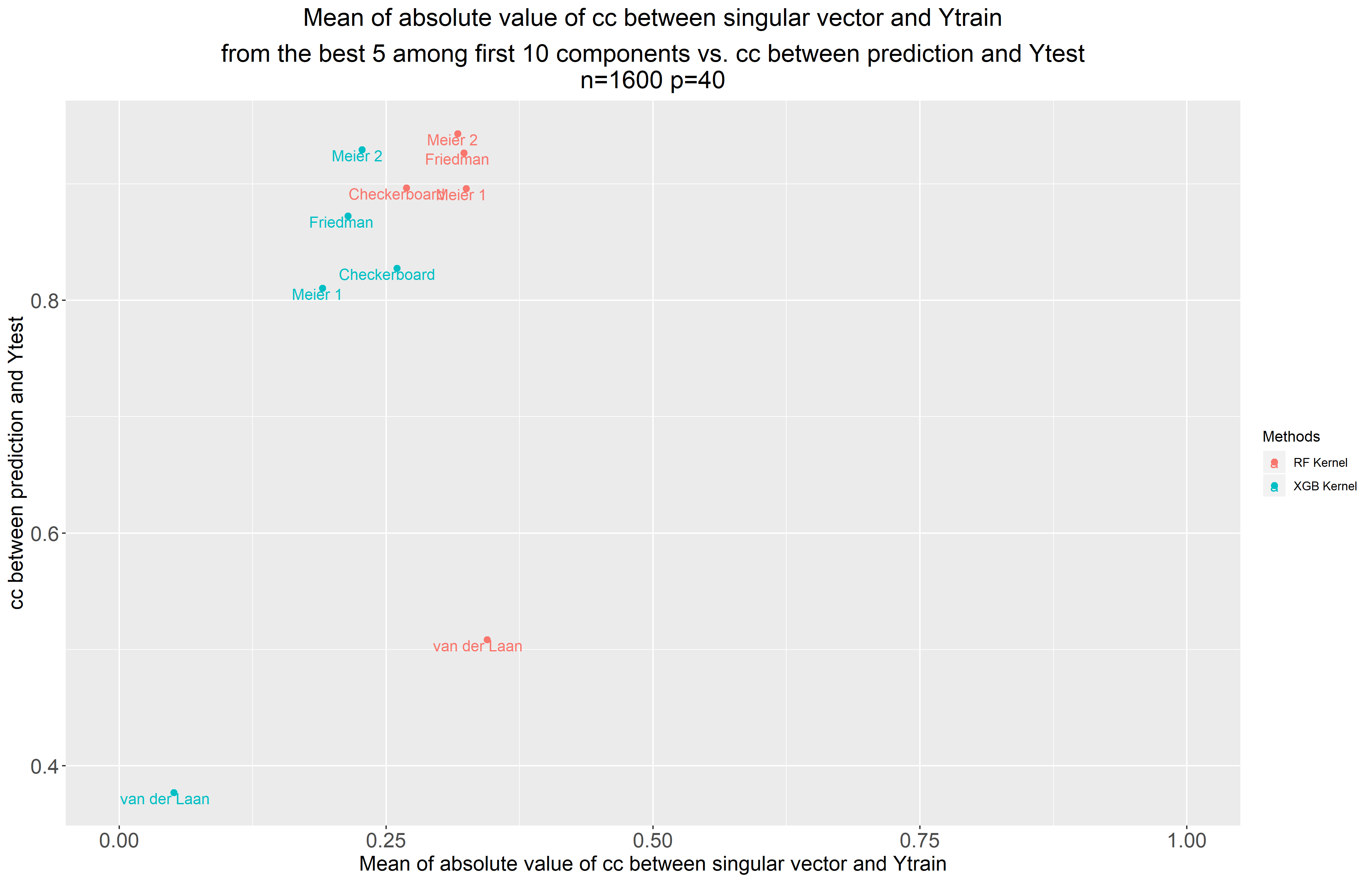}
        \subcaption{}
     \end{minipage}
     \\
    
\caption{Mean of absolute value of cc between the eigenvector and Ytrain from the first  component (i.e. largest eigenvalue) vs. cc between prediction and Ytest (left). Mean of absolute value of cc between eigenvector and Ytrain from the best  component (i.e. largest absolute correlation coefficient) vs. cc between prediction and Ytest (middle). Mean of absolute value of cc between the eigenvector and Ytrain from the best 5 among first 10 components vs. cc between prediction and Ytest (right) }
\label{fig:cc vs cc}
\end{figure}

The kernel-target alignment vs performance of the  kernel and XGB kernel for the different data generating mechanisms and simulation settings is summarized in Figs.\ref{fig:cc vs cc}(a-l). The performance on the test set was evaluated by an absolute value of the Pearson correlation coefficient between the predicted values and the target (Ytest) (the y-axis in the Figs.\ref{fig:cc vs cc}(a-l)). 

Three summaries of the kernel-target alignment on the training set have been evaluated (the x-axis in the Figs.\ref{fig:cc vs cc}(a-l)). On the left side (Figs.\ref{fig:cc vs cc}(a,d,g,j)), the mean of the absolute value of the correlation coefficient between the first eigenvector  corresponding to the largest eigenvalue (singular value) and Ytrain is shown (\cite{cristianini2001},\cite{cristianini2006}). In the middle (Figs.\ref{fig:cc vs cc}(b,e,h,k)), the mean correlation coefficient between the eigenvector with maximum correlation with Ytrain is shown. On the right, the mean of absolute value of the correlation coefficient between the eigenvectors (singular vectors) of $\mbold{K}$ and Ytrain from the best 5 (with highest correlation coefficient) among leading 10 eigenvectors  is displayed. Overall, the higher the kernel-target alignment obtained on the training set, the better the performance of the RF/XGB kernels across the data generating mechanisms and simulation settings. The eigenvectors  corresponding to the largest eigenvalue (singular value) are not necessarily best aligned with the target, which is demonstrated for the Checkerboard data for both (RF and XGB) kernels and the van der Laan data set for the XGB kernel, respectively (Figs.\ref{fig:cc vs cc}(a,b,d,e,g,h,j,k)). 

\section{Application Using Real Life Data Sets}
\label{sec:app}
\subsection{Experimental Setup}
\label{sec:appSetup}
We assessed the kernel-target alignment in tree-based ensemble kernel using real life data sets, the summary of which is given in Table \ref{tab:tableDataSets}.

For the larger data sets (California and Protein) we randomly selected 2000 samples and split them into training and test set, with 1500 and 500 samples, respectively. We repeated the analysis 200 times to evaluate the kernel-target alignment of RF and XGB kernels, respectively. For the other data sets we split the data into training and test set in the ratio 3 to 1, respectively, and repeated the analysis 200 times. Similarly to the simulation, we evaluated the kernel-target alignment on the training sets and related it to the performance on the test sets.

\subsection{Real Life Data Sets Results}
\label{sec:appResults}

\begin{figure}
     \centering
     \begin{minipage}[b]{0.45\textwidth}
         \centering
         \includegraphics[width=\textwidth]{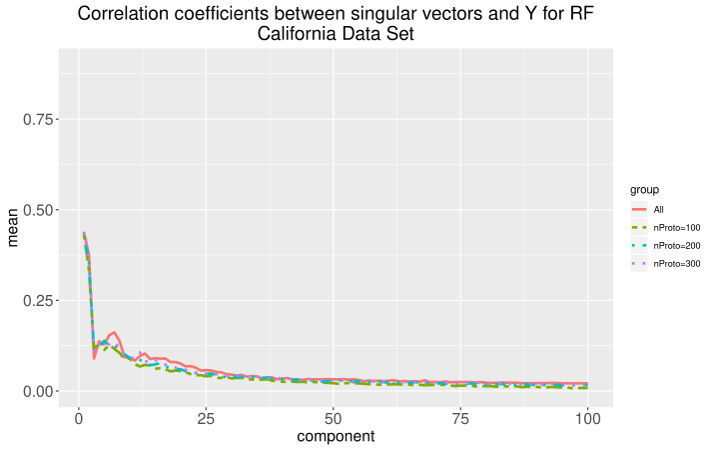}
        \subcaption{}
     \end{minipage}
     \hfill
     \begin{minipage}[b]{0.45\textwidth}
         \centering
         \includegraphics[width=\textwidth]{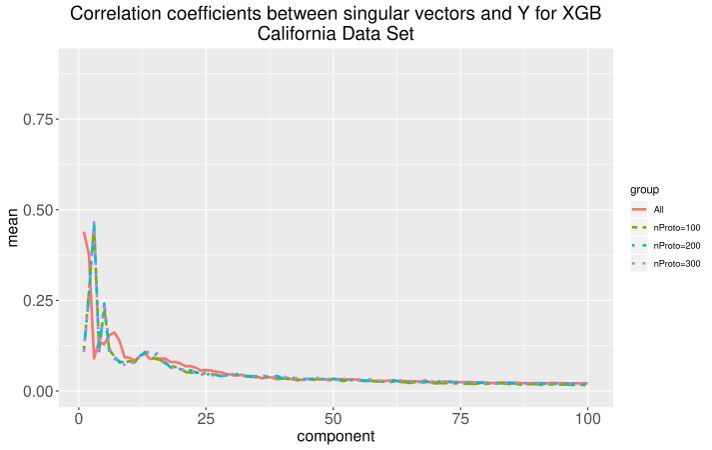}
          \subcaption{}
     \end{minipage}
  \caption{Kernel-target alignment spectra and relevant dimensionality---California housing}
\label{fig:rd-california}
\end{figure}

\begin{figure}
     \centering
     \begin{minipage}[b]{0.45\textwidth}
         \centering
         \includegraphics[width=\textwidth]{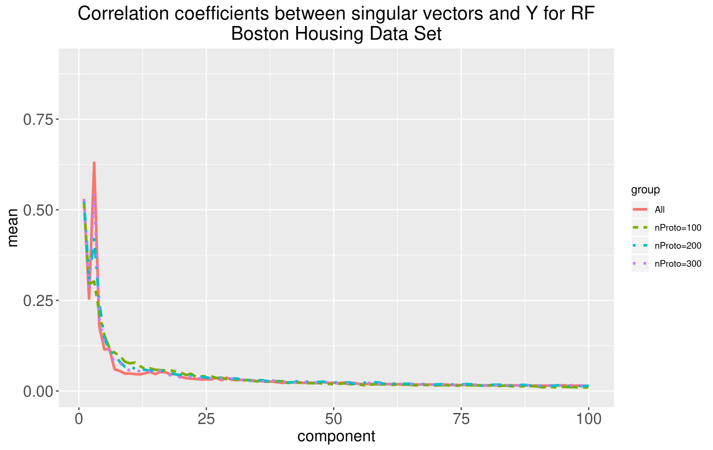}
           \subcaption{}
     \end{minipage}
     \hfill
     \begin{minipage}[b]{0.45\textwidth}
         \centering
         \includegraphics[width=\textwidth]{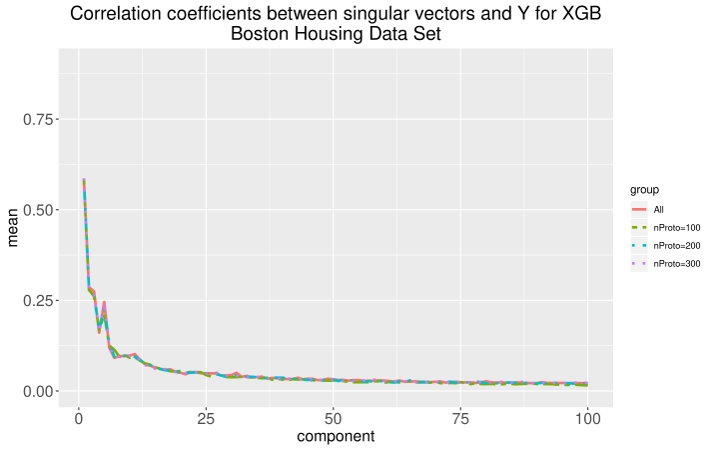}
           \subcaption{}
     \end{minipage}
  \caption{Kernel-target alignment spectra and Relevant Dimensionality---Boston housing}
\label{fig:rd-boston}
\end{figure}

\begin{figure}
     \centering
     \begin{minipage}[b]{0.45\textwidth}
         \centering
         \includegraphics[width=\textwidth]{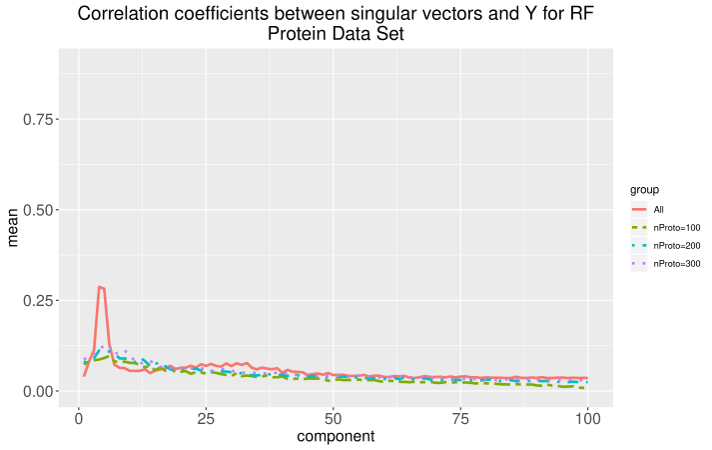}
    \subcaption{}
     \end{minipage}
     \hfill
     \begin{minipage}[b]{0.45\textwidth}
         \centering
         \includegraphics[width=\textwidth]{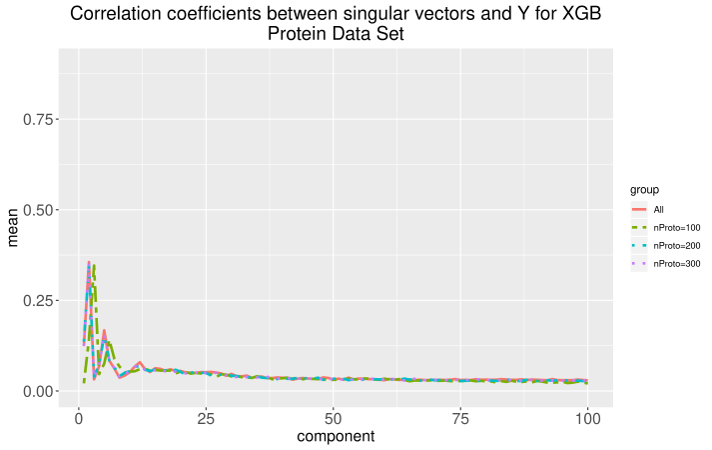}
            \subcaption{}
     \end{minipage}
  \caption{Kernel-target alignment spectra and relevant dimensionality---Protein Tertiary Structure}
\label{fig:rd-protein}
\end{figure}

\begin{figure}
     \centering
     \begin{minipage}[b]{0.45\textwidth}
         \centering
         \includegraphics[width=\textwidth]{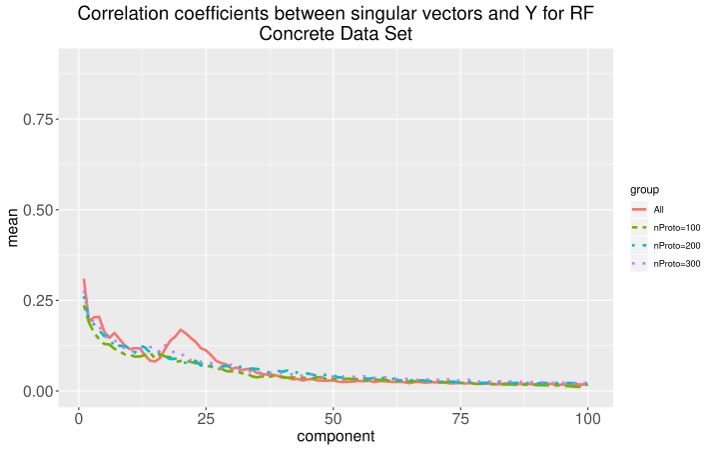}
         \subcaption{}
     \end{minipage}
     \hfill
     \begin{minipage}[b]{0.45\textwidth}
         \centering
         \includegraphics[width=\textwidth]{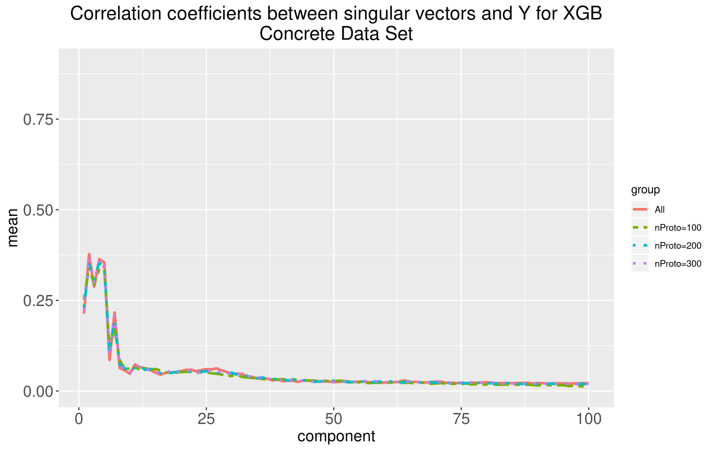}
            \subcaption{}
     \end{minipage}
  \caption{Kernel-target alignment spectra and relevant dimensionality---Concrete Compressive Strength}
\label{fig:rd-concrete}
\end{figure}

\begin{figure}
     \centering
     \begin{minipage}[b]{0.45\textwidth}
         \centering
         \includegraphics[width=\textwidth]{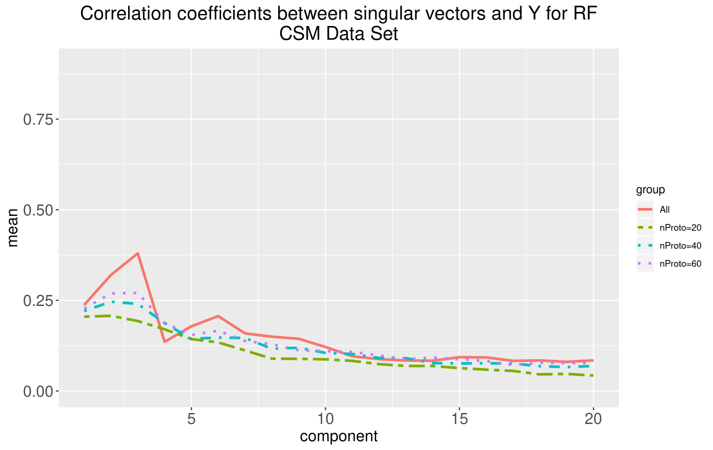}
          \subcaption{}
     \end{minipage}
     \hfill
     \begin{minipage}[b]{0.45\textwidth}
         \centering
         \includegraphics[width=\textwidth]{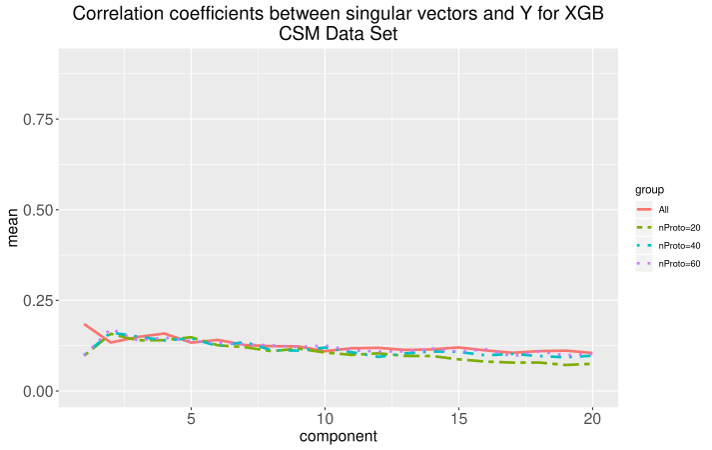}
         \subcaption{}
     \end{minipage}
  \caption{Kernel-target alignment spectra and relevant dimensionality---Conventional and Social Movie}
\label{fig:rd-csm}
\end{figure}

The alignment spectra are given in the Figs. \ref{fig:rd-california}, \ref{fig:rd-boston}, \ref{fig:rd-protein}, \ref{fig:rd-concrete} and \ref{fig:rd-csm} for California, Boston, Protein, Concrete and CSM data sets, respectively. There results from the RF and XGB kernels are provided in sub-figure (a) and (b), respectively. California, Boston and Protein data sets (Figs. \ref{fig:rd-california}(a,b), \ref{fig:rd-boston}(a,b) and \ref{fig:rd-protein}(a,b) data sets are characterized by strong peaks in the alignment spectra for both RF and XGB kernels. The Concrete data set has multiple strong peaks for the XGB kernel (Fig.\ref{fig:rd-concrete}(b)) in contrast to that of the RF kernel. On the other hand, the CSM data set has strong peaks for the RF kernel (Fig.\ref{fig:rd-csm}(a)), whereas the alignment spectrum obtained from the XGB kernel is flat (Fig.\ref{fig:rd-csm}(b)). The performance of the RF and XGB kernels in terms of the correlation of the predictions with the target on the test set are provided in Table \ref{tab:resultsCCs}. In addition, the average of the correlation coefficients of the top five eigenvector components and target obtained from the training sets are provided as a summary measure of the kernel-target alignment (see Table \ref{tab:resultsCCs}).  Using this metric, for three data sets (Boston, Protein and CSM) the RF kernel shows higher alignment with the target than that of the XGB, and in turn, a better performance. On the other hand, for the Concrete data set, the kernel-target alignment for the XGB kernel is higher than that of the RF kernel. As a consequence, the XGB kernel exhibits better performance. For the California data set, the RF and XGB show comparable kernel-target alignment, with RF slightly outperforming the XGB. To note, the overall alignment spectra of the California data set for the RF and the XGB exhibit the same pattern (see Figs.\ref{fig:rd-california}(a),(b)).

For completeness, the results of the prediction performance in terms of the mean squared error (MSE) are given in the Table \ref{tab:resultsMSEs}.

\begin{table}

\centering
   \caption{Summary of the real life datasets}
   \begin{tabular}{lrr}
  \hline
  Dataset & n & p \\
  \hline
  California Housing \cite{pace1997} &  20640 & 9 \\
  Boston Housing \cite{harrison1978},\cite{Dua2019}& 506 & 13 \\
  Protein Tertiary Structure \cite{Dua2019} &  45730 & 9\\

   Concrete Compressive Strength \cite{yeh1998}, \cite{Dua2019} &  1030 & 9 \\
    Conventional and Social Movie (CSM) \cite{ahmed2015}, \cite{Dua2019}&  187 & 12\\
\hline
   \end{tabular}
\label{tab:tableDataSets}   
\end{table}

\begin{table}
%\hl{The first two columns of Table 2 show the the performance of tree ensemble based kernel methods. The next four columns show the kernel-target alignment. Do we want to highlight the method with better kernel-target alignment as well. In addition, the trend is that the better the performance, the better the kernel-target alignment, but not always so, especially for RFkmax and XGBkmax. Do we want to show the results from Rfk/XGBkmax as well? This is related to the question of in addition to a full spectrum, what is the summary statistic or multiple summary statistics we would like to recommend to characterize the spectrum.}
 \begin{adjustwidth}{-.8in}{-.8in}  
 \centering
   \caption{Performance of the RF and XGB kernels on test set and summary measures of the kernel alignment obtained from the training set. RFk and XGBk refer to the mean correlation coefficients between target and the predictions obtained from the test sets for the RF and XGB kernel, respectively. 
   RFk5 and XGBk5 refer to the average correlation coefficient of top 5 components and target ordered according to the eigenvalues of the kernel matrix for RF and XGB kernel (from training set), respectively. 
   The metrics are provided as means and standard deviations.}
   \setlength\tabcolsep{2pt}
  \begin{tabular}{lllll}
  \hline
  Dataset & RFk & XGBk & RFk5 & XGBk5  \\
  \hline
California Housing & $\mbold{0.872 (0.016)}$ & 0.856 (0.016) & 0.234 (0.023) & \mbold{0.235 (0.023)}\\
Boston Housing & $\mbold{0.943 (0.021)}$ & 0.919 (0.022) &\mbold{0.341 (0.020)}& 0.311 (0.022)\\
Protein Tertiary Structure & $\mbold{0.671 (0.029)}$ & 0.575 (0.033) & \mbold{0.161 (0.025)} & 0.120 (0.016) \\
Concrete Compressive Strength&0.965 (0.007) &$\mbold{0.966 (0.008)}$  & 0.215 (0.021)& \mbold{0.319 (0.025)} \\
CSM & $\mbold{0.544 (0.091)}$ & 0.434 (0.104) &\mbold{0.250 (0.027)} & 0.144 (0.032)\\
\hline
   \end{tabular}
\label{tab:resultsCCs}   
    \end{adjustwidth}
\end{table}

\section{Discussion and Future Work}
\label{sec:discuss}
In this paper, we have shown that for regression, the performance of the tree ensemble (RF/XGB) based kernels is associate with the degree of the kernel-target alignment. In a comprehensive simulation study and real life data sets we demonstrated that strong target aligned components of the kernel matrix $\mbold{K}$ are translated into high performance of the tree ensemble based kernels. The strongly target aligned components correspond to (typically a small number of) eigenvectors of the kernel matrix with larger eigenvalues (i.e. they are in the left side of the eigenvalue spectrum). However they do not necessarily exactly follow the magnitude based ordering of the eigenvalues. This suggests that the target aligned components span a low dimensional manifold that is implicitly represented by the tree based kernel. Moreover, the strongly aligned components (peaks) are persistent as shown in the sensitivity analysis with a landmark tree based kernel learning and an increasing number of landmarks.

The kernel-target alignment can be applied to other tree ensemble based kernels, e.g. the recent RF and XGB variants. They include oblique, rotation or mixup forests (\cite{menze2011}, \cite{rodriguez2006} and \cite{rodriguez2020}, respectively). Of interest is also the kernel-target alignment of kernels obtained from Bayesian approaches such as Mondrian forests \cite{balog2016} or Bayesian non parametric partitions and Bayesian additive regression trees (BART), (\cite{fan2020} and \cite{linero2017}).

Furthermore, the concept of the tree ensemble based kernel is inherent to other prediction targets such as the binomial and time to event targets that represent classification and survival, respectively. For example, the proximity matrix i.e. ensemble kernel for the survival forest is readily available \cite{ishwaran2019}. For the estimation of the kernel-target alignment for a survival target, an additional challenge of incomplete information about the target due to censoring needs to be addressed and is an interesting topic for future research. 

We used kernel ridge regression as a kernel learning algorithm in our contribution. There have been also recent advancements in the development of the response guided principal components \cite{lang2020}, \cite{Tay2021}. These have focused on principal component regression and incorporation of sparsity constraints through the LASSO penalty. As our results support the notion of relevant dimensionality expressed by the target aligned components for tree ensemble based kernels, we plan to explore sparse, response (target) guided nonlinear principal component regression for the tree ensemble kernel learning in the future. 

\pagebreak

\section{Appendix}
\label{sec:append}

\subsection{RF and the RF Kernel}\label{sec2sub4}
 Random Forest (RF) is defined  as an ensemble of tree predictors grown on bootstrapped samples of a training set \cite{breiman2000}. When considering an ensemble of tree predictors $\{h(.,\Theta_m,D_n), m=1,2,\ldots,M\}$, with $\{h(.,\Theta_m,D_n)\}$ representing a single tree. The $\Theta_1, \Theta_2,\ldots\Theta_M$ are iid random variables that encode the randomization necessary for the tree construction \cite{scornet2016}, \cite{ishwaran2019}.

The RF predictor is obtained as:
\begin{eqnarray}
h_{\textrm{RF}}(\mbold{X},\Theta_1,\ldots,\Theta_m,D_n)&=& \frac{1}{M}\sum_{m=1}^M h(\mbold{X},\Theta_m,D_n)
 \end{eqnarray}
RF kernel ensuing from the RF is defined as a probability that  $\mbold{X_i}$ and $\mbold{X_j}$ are in the same terminal node $R_k(\Theta_m$) \cite{breiman2000}, \cite{scornet2016}.
\begin{eqnarray}
k_{RF}(\mbold{X_i},\mbold{X_j})=\frac{1}{M}
\sum_{m=1}^M \sum_{k=1}^T I(\mbold{X_i},\mbold{X_j} \in R_k(\Theta_m))
\label{EqKernelRF}
\end{eqnarray}
%\uline{where $I(\cdot)$ denotes the indicator function.}
where $I(\cdot)$ denotes the indicator function.

\subsection{Gradient Boosted Trees (GBT) and the GBT Kernel}\label{sec2sub5}
The GBT are (similarly to RF) ensemble of tree predictors. In contrast to the RF, the GBT ensemble predictor is obtained as a sum of weighted individual tree predictors $h_m(\mbold{X},D_n)$ through iterative optimization of an objective (cost) function \cite{friedman2001},\cite{Chen2016} :

\begin{eqnarray}
h_{\textrm{GBT}}(\mbold{X},D_n)&=& \sum_{m=1}^M h_m(\mbold{X},D_n)
\end{eqnarray}

The objective function of GBT comprises of a loss function and for the extreme gradient boosting a regularization term is added to control the model complexity. In our work we used the extreme gradient boosting (XGB) %(\uline{xgb \sout{xgboost}}) 
implementation of the GBTs \cite{Chen2016}. 

The objective function of the XGB algorithm $L_{\textrm{XGB}}$ is given follows:

\begin{eqnarray}
L_{\textrm{XGB}} &=& \sum_{i=1}^{n} l(y_i, h_{\textrm{XGB}}(X_i))+\Omega(h_m) \\
&=& \sum_{i=1}^{n} l(y_i, h_{\textrm{XGB}}(X_i))+\gamma T+ 0.5 \lambda ||w||^2
\end{eqnarray}
where \\
$l(.)$ is the loss function. The loss function used in xgboost is squared error and logistic loss for regression and classification, respectively  \\
$\Omega$ denotes the regularization penalty \\
$T$, $\gamma$ is the number of tree terminal nodes and a corresponding regularization parameter, respectively\\
$\lambda$ is a regularization parameter controlling for the L2 norm of the individual tree weights $||w||^2$

As for the RF kernel, the XGB kernel is defined as a probability that  $\mbold{X_i}$ and $\mbold{X_j}$ are in the same terminal node $R_k$($h_m$) \cite{Chen2018}.

\begin{eqnarray}
k_{xgb}(\mbold{X_i},\mbold{X_j})=\frac{1}{M}
\sum_{m=1}^M \sum_{k=1}^T I(\mbold{X_i},\mbold{X_j} \in R_k(h_m))
\label{EqKernelxgb}
\end{eqnarray}

%-----------------------meier 1----------------
%-----------------------meier 2-------------

%---------------------van der laan---------------------------

%-------------------------------------cc vs. cc first component only----------------------

%--------------------------------Real Data---------------------------------
%-----------------------------rd-California housing--------------------------------
%-----------------------------rd-Boston housing--------------------------------

%-----------------------------rd-Protein--------------------------------

%-----------------------------rd-concrete--------------------------------

%-----------------------------rd-csm--------------------------------

\begin{table}
    \begin{adjustwidth}{-.5in}{-.5in}  
\centering
   \caption{Performance (measured by the MSE) for the RF/XGB and the RF/XGB kernels}
   \begin{tabular}{lllll}
  \hline
  Dataset & RF & XGB & RFk & XGBk \\
  \hline
California housing & 3.64 $\times10^9$ & 5.46 $\times10^9$ & $\mbold{3.24 \times10^9}$ & 3.64$\times10^9$ \\
&($3.97\times10^8$)&($6.06\times10^8$)&
$\mbold{( 3.89\times10^8)}$&($3.86\times10^8$)\\
Boston housing &12.512 (4.570) & 19.172 (5.697) & $\mbold{9.806 (3.61)}$ & 13.349 (3.544)\\
Protein Tertiary Structure & 21.248 (1.560)&35.121(2.771)& $\mbold{20.504(1.651)}$&27.385(2.253)\\
Concrete Compressive Strength &31.083(4.878)&38.049(7.849)&19.851(4.12)&$\mbold{18.647(3.907)}$\\
CSM & 0.727 (0.158)& 1.122(0.266)& $\mbold{0.718(0.152)}$ & 0.906(0.183)\\
\hline
   \end{tabular}
\label{tab:resultsMSEs}   
\end{adjustwidth}
\end{table}

%\section{Acknowledgements}

%\bibliographystyle{unsrt}  
%\bibliographystyle{natbib}

%\bibliographystyle{model5-names}
%\biboptions{authoryear}

%\printbibliography

\bibliography{rfkernel}

\begin{thebibliography}{10}
\expandafter\ifx\csname url\endcsname\relax
  \def\url#1{\texttt{#1}}\fi
\expandafter\ifx\csname urlprefix\endcsname\relax\def\urlprefix{URL }\fi
\expandafter\ifx\csname href\endcsname\relax
  \def\href#1#2{#2} \def\path#1{#1}\fi

\bibitem{feng2018}
J.~Feng, Y.~Yu, Z.-H. Zhou,
  \href{https://proceedings.neurips.cc/paper/2018/file/39027dfad5138c9ca0c474d71db915c3-Paper.pdf}{Multi-layered
  gradient boosting decision trees}, in: S.~Bengio, H.~Wallach, H.~Larochelle,
  K.~Grauman, N.~Cesa-Bianchi, R.~Garnett (Eds.), Advances in Neural
  Information Processing Systems, Vol.~31, Curran Associates, Inc., 2018.
\newline\urlprefix\url{https://proceedings.neurips.cc/paper/2018/file/39027dfad5138c9ca0c474d71db915c3-Paper.pdf}

\bibitem{scornet2016}
E.~Scornet, Random forests and kernel methods, IEEE Transactions on Information
  Theory 62~(3) (2016) 1485 -- 1500.

\bibitem{Chen2018}
G.~Chen, D.~Shah, Explaining the Success of Nearest Neighbor Methods in
  Prediction, MIT Press, 2018, Ch. Adaptive neighbors and far away neighbors.

\bibitem{davies2014}
A.~Davies, Z.~Ghahramani, The random forest kernel and other kernels for big
  data from random partitions., arXiv preprint arXiv:1402.4293.

\bibitem{feng2021}
D.~Feng, R.~Baumgartner, (decision and regression) tree ensemble based kernels
  for regression and classification., arXiv preprint arXiv:2012.10737.

\bibitem{cristianini2001}
N.~Cristianini, J.~Elisseev, J.~Shawe~Taylor, et~al., On kernel target
  alignment, in: Proceedings of the Neural Information Processing Systems,
  NeurIPS, 2001.

\bibitem{cristianini2006}
N.~Cristianini, J.~Elisseev, J.~Shawe~Taylor, et~al., Innovations in Machinne
  Learning, Springer, 2006, Ch. On Kernel Target Alignment.

\bibitem{braun2009}
M.~Braun, J.~Buhmann, K.~Muller, On relevant dimensions in kernel feature
  spaces, Journal of Machine Learning Research 9 (2008) 1875--1908.

\bibitem{montavon2011}
G.~Montavon, M.~Braun, K.~Muller, Kernel analysis of deep networks, Journal of
  Machine Learning Research 12 (2011) 2563--2581.

\bibitem{pekalska2001}
E.~Pekalska, P.~Paclik, R.~Duin, A generalized kernel approach to
  dissimilarity-based classification, Journal of Machine Learning Research 2
  (2001) 175--211.

\bibitem{bien2011}
J.~Bien, R.~Tibshirani, Prototype selection for interpretable classification,
  Annals of Applied Statistics 5~(4) (2011) 2403--2424.

\bibitem{kar2011}
P.~Kar, P.~Jain, Similarity-based learning via data driven embeddings,
  Proceedings of the Advances in Neural Information Processing Systems, (2011)
  1998–2006.

\bibitem{balcan2008}
M.~Balcan, A.~Blum, S.~N, A theory of learning with similarity functions,
  Machine Learning 72 (2008) 89--112.

\bibitem{breiman2000}
L.~Breiman, Some infinity theory for predictor ensembles, Tech. rep., Technical
  Report 579, Statistics Dept. UCB (2000).

\bibitem{ishwaran2019}
H.~Ishwaran, M.~Lu, Standard errors and confidence intervals for variable
  importance in random forest regression, classification, and survival,
  Statistics in Medicine (2019) 558--582.

\bibitem{herbich2001}
R.~Herbich, Learning kernel classifiers, MIT Press, 2001.

\bibitem{schoelkopf2001}
B.~Schoelkopf, A.~Smola, Learning with kernels, MIT Press, 2001.

\bibitem{friedmanHastieTibshirani2009}
J.~Friedman, T.~Hastie, R.~Tibshirani, The Elements of Statistical Learning,
  Springer, 2009.

\bibitem{fan2020}
X.~Fan, B.~Li, L.~Luo, et~al., Bayesian nonparametric space partitions: A
  survey., arXiv preprint arXiv:2002.11394.

\bibitem{Rcran}
{R Core Team}, \href{https://www.R-project.org/}{R: A Language and Environment
  for Statistical Computing}, R Foundation for Statistical Computing, Vienna,
  Austria (2017).
\newline\urlprefix\url{https://www.R-project.org/}

\bibitem{ranger}
M.~N. Wright, A.~Ziegler, {ranger}: A fast implementation of random forests for
  high dimensional data in {C++} and {R}, Journal of Statistical Software
  77~(1) (2017) 1--17.
\newblock \href {http://dx.doi.org/10.18637/jss.v077.i01}
  {\path{doi:10.18637/jss.v077.i01}}.

\bibitem{xgboost}
T.~Chen, T.~He, M.~Benesty, V.~Khotilovich, Y.~Tang, H.~Cho, K.~Chen,
  R.~Mitchell, I.~Cano, T.~Zhou, M.~Li, J.~Xie, M.~Lin, Y.~Geng, Y.~Li,
  \href{https://CRAN.R-project.org/package=xgboost}{xgboost: Extreme Gradient
  Boosting}, r package version 1.2.0.1 (2020).
\newline\urlprefix\url{https://CRAN.R-project.org/package=xgboost}

\bibitem{friedman}
J.~H. Friedman, Multivariate adaptive regression splines, The annals of
  statistics (1991) 1--67.

\bibitem{meier}
L.~Meier, S.~Van~de Geer, P.~B{\"u}hlmann, et~al., High-dimensional additive
  modeling, The Annals of Statistics 37~(6B) (2009) 3779--3821.

\bibitem{van2007super}
M.~J. Van~der Laan, E.~C. Polley, A.~E. Hubbard, Super learner, Statistical
  applications in genetics and molecular biology 6~(1).

\bibitem{zhu2015}
R.~Zhu, D.~Zeng, M.~R. Kosorok, Reinforcement learning trees, Journal of the
  American Statistical Association 110~(512) (2015) 1770--1784.

\bibitem{pace1997}
K.~Pace, R.~Barry, Sparse spatial autoregressions, Statistics \& Probability
  Letters 33~(3) (1997) 291–297.

\bibitem{harrison1978}
D.~Harrison~Jr, D.~Rubinfeld, Hedonic housing prices and the demand for clean
  air., Journal of environmental economics and management 5 (1978) 81--102.

\bibitem{Dua2019}
D.~Dua, C.~Graff, \href{http://archive.ics.uci.edu/ml}{{UCI} machine learning
  repository} (2017).
\newline\urlprefix\url{http://archive.ics.uci.edu/ml}

\bibitem{yeh1998}
J.~Yeh, Modeling of strength of high-performance concrete using artificial
  neural networks., Cement and Concrete research 28 (1998) 1797--1808.

\bibitem{ahmed2015}
D.~Ahmed, A.~Jahnagir, M.~A, et~al., Using crowd source based features from
  social media and conventional features to predict the movies popularity., in:
  IEEE International Conference on Smart City/SocialCom/SustainCom (SmartCity),
  IEEE, 2015, pp. 273--278.

\bibitem{menze2011}
D.~Menze, B.~Kelm, U.~Koethe, et~al., On oblique random forests., in:
  Proceedings ECML/PKDD, Lecture Notes in Computer Science, 6911, Springer,
  2011, pp. 453--469.

\bibitem{rodriguez2006}
J.~Rodriguez, L.~Kuncheva, C.~Alonso, Rotation forest: A new classifier
  ensemble method, IEEE Trans Pattern Anal Mach Intell 28~(10) (2006) 1619--30.

\bibitem{rodriguez2020}
J.~Rodriguez, M.~Juez-Gil, A.~Arnaiz-González, et~al., An experimental
  evaluation of mixup regression forests, Expert Systems and Applications 151
  (2020) 113376.

\bibitem{balog2016}
M.~Balog, B.~Lakshminarayanan, Z.~Ghahramani, D.~M. Roy, Y.~W. Teh, The
  mondrian kernel, in: Proceedings of the 32nd Conference on Uncertainty in
  Artificial Intelligence, 2016, p. 32–41.

\bibitem{linero2017}
A.~R. Linero, A review of tree-based bayesian methods, Communications for
  Statistical Applications and Methods 24~(6) (2017) 543--559.

\bibitem{lang2020}
W.~Lang, H.~Zou, A simple method to improve principal components regression.,
  Stat (2020) e288.

\bibitem{Tay2021}
J.~Tay, J.~Friedman, R.~Tibshirani, Principal component-guided sparse
  regression., The Canadian Journal of Statistics.

\bibitem{friedman2001}
J.~Friedman, Greedy function approximation: A gradient boosting machine.,
  Annals of Statistics 29~(5) (2001) 1189--1232.

\bibitem{Chen2016}
T.~Chen, G.~C, A scalable tree boosting system., in: Proc. 22nd ACM SIGKDD Int.
  Conf. Knowl. Discovery Data Mining, 2016, pp. 785--794.

\end{thebibliography}
\end{document}